\documentclass[11pt]{article}
\usepackage[margin=1in]{geometry}
\usepackage{mathptmx}
\usepackage{graphicx,tikz}
\usetikzlibrary{arrows.meta}
\usepackage{courier}
\usepackage{microtype}
\usepackage[T1]{fontenc}
\usepackage{amsmath,amssymb}
\usepackage{graphicx}
\usepackage{booktabs}
\usepackage{array}
\usepackage{placeins}            

\makeatletter\def\fps@figure{!htbp}\def\fps@table{!htbp}\makeatother 
\usepackage{multirow}
\usepackage{xcolor}
\usepackage{url}
\usepackage[hidelinks]{hyperref}
\usepackage{cite}
\input{numbers.tex}
\input{numbers_subject.tex}
\newcommand{\pp}{\,pp}

\usepackage[export]{adjustbox}
\newcommand{\fitcol}[1]{\adjustbox{max width=\columnwidth}{#1}}
\newcommand{\fitwide}[1]{\adjustbox{max width=\textwidth}{#1}}
\newcommand{\figcol}{0.62\textwidth}
\newcommand{\signalfig}{\includegraphics[width=0.66\textwidth]{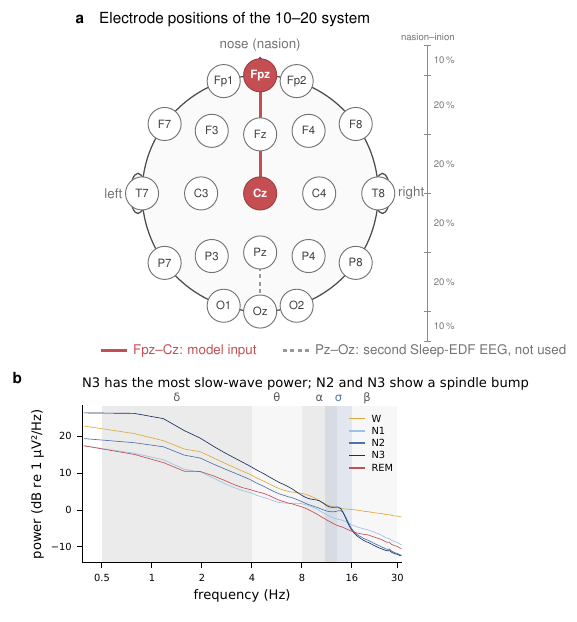}}
\newcommand{\stagefeatw}{10.6cm}\newcommand{\stagerkw}{2.9cm}
\makeatletter\newcommand{\appfloatsetup}{\def\fps@figure{!ht}\def\fps@table{!ht}}\makeatother
\newenvironment{IEEEkeywords}{\par\smallskip\noindent\textbf{Keywords:}\ }{\par}
\newcommand{\appendices}{\appendix}
\usepackage{caption}
\begin{document}
\title{Is 4 Bits Enough for EEG Sleep Staging?\\ A Subject-Level Benchmark of MXFP4, NVFP4 and FP8}
\author{Emese Batizi \qquad Hossein Valavi\\[2pt]
\normalsize Princeton University, Princeton, NJ, USA\\
\normalsize \texttt{\{eb1870, hvalavi\}@princeton.edu}}
\date{}
\maketitle
\begin{abstract}
Staging sleep on a wearable device keeps the raw electroencephalogram (EEG) private and saves the energy needed to transmit it. It also requires small, energy-efficient neural networks. We benchmark 13 post-training quantization (PTQ) configurations, from FP16 to the block-scaled 4-bit formats MXFP4 and NVFP4, for five-stage sleep staging from single-channel Fpz--Cz EEG. We use Sleep-EDF-78 (\val{data.subjects} subjects), subject-wise cross-validation, three training seeds, value-exact emulation, and per-subject paired statistics. All 8- and 16-bit formats stay within \val{eightbit.maxabs} percentage points (pp) of FP32 Macro-F1. With 4-bit weights and activations, an epoch-wise CNN loses \val{w4a4.cnn.range}\pp{}, and a temporal-context network loses at most \val{w4a4.ctx.max}\pp{}. A same-encoder ablation shows that temporal context explains little of this gap. At equal block size, NVFP4 outperforms MXFP4. The first layer, which quantizes the raw EEG, is the most sensitive layer in both networks. Keeping the first and last layers in FP8 brings both tested 4-bit formats within \val{mix.maxabs}\pp{} of FP32. This recipe adds at most \val{mix.ovhmax}\% weight storage, needs no retraining, and is equivalent to quantization-aware training (QAT) within an exploratory 1-pp margin. The 4-bit losses fall on the same subjects across architectures (Spearman $\rho=\val{subj.mx.xmodel}$--$\val{abl.rho.max}$), and they reduce N3 recall the most. On the nested Sleep-EDF-20 subset, NVFP4 with FP8 first and last layers again stays within 0.25\pp{} of FP32. Under MXFP4, however, the ranking of the architectures reverses, and FP8 first and last layers do not fully recover two of the three tested models. In our emulation, NVFP4 with FP8 first and last layers thus loses little accuracy without QAT in both cohorts.
\end{abstract}


\section{Introduction}
Sleep staging assigns each 30-second epoch of a polysomnogram to one of five stages: wake (W), N1, N2, N3, or rapid-eye-movement (REM) sleep~\cite{iber2007aasm,rechtschaffen1968manual}. It is the basis of most sleep diagnostics. Manual scoring is slow, and agreement between scorers is only moderate. In the American Academy of Sleep Medicine (AASM) inter-scorer reliability program, overall epoch agreement was 82.6\%~\cite{rosenberg2013aasm}. Among expert scorers, Cohen's $\kappa$ was 0.68 under the Rechtschaffen and Kales (R\&K) rules and 0.76 under the AASM rules~\cite{dankerhopfe2009interrater}. Deep networks that read a single electroencephalography (EEG) channel now reach comparable agreement~\cite{supratak2017deepsleepnet,supratak2020tinysleepnet,phan2019seqsleepnet,perslev2019utime,eldele2021attnsleep,phan2022sleeptransformer}. This makes sleep monitoring at home and with wearable devices feasible~\cite{dezambotti2019wearable}. Dry-electrode EEG headbands, for example, already stage sleep with an accuracy comparable to that of polysomnography scoring~\cite{arnal2020dreem}.

There are three main reasons to run the staging model on the device instead of streaming the signal to a phone or a server. \emph{Privacy:} EEG is health data, and neural recordings raise privacy concerns of their own~\cite{yuste2017four,ienca2017towards}. A device that stages sleep locally needs to send only a hypnogram. For an 8-hour night (960 epochs), the hypnogram takes about 1\,kB, whereas the raw signal takes about 5.8\,MB (100\,Hz, 16\,bit). The raw EEG never has to leave the device. \emph{Energy:} In wearable EEG systems, the radio can dominate the power budget. Processing the EEG on the chip instead of transmitting it has reduced system power 14-fold~\cite{verma2010micropower}. \emph{Latency:} Closed-loop interventions must act on the signal in real time and cannot depend on a network connection. One example is auditory stimulation timed to the slow oscillations of slow-wave (N3) sleep, which has been shown to enhance memory retention~\cite{ngo2013auditory}. In all three cases, the model must fit within the memory and energy budget of a battery-powered device.

Low-precision arithmetic is one of the most effective ways to meet such a budget. INT8 post-training quantization (PTQ) is mature~\cite{jacob2018quantization,krishnamoorthi2018quantizing,nagel2021white}, and 8-bit floating point (FP8) is standard on recent accelerators~\cite{micikevicius2022fp8}. The Open Compute Project (OCP) Microscaling (MX) specification~\cite{ocp2023mx,rouhani2023microscaling} and NVIDIA's NVFP4~\cite{nvidia2025nvfp4} go further. They define 4-bit floating-point formats in which small blocks of values share one scale factor. These block-scaled formats were developed and tested mainly on large language and vision models. Small biosignal networks differ in three ways. Their input is a raw physiological signal with a high dynamic range. They have only $10^4$--$10^5$ parameters. Finally, the clinical question is whether every patient is staged reliably, not only whether the average is acceptable. Compression is known to affect some inputs much more than others~\cite{hooker2019forget}, and mean accuracy can stay unchanged while many individual predictions change~\cite{dutta2024accuracy}. We are not aware of a subject-level evaluation of block-scaled 4-bit formats for EEG sleep staging. This paper provides one. Its contributions are as follows:
\begin{enumerate}
\item \textbf{A controlled benchmark.} We evaluate 13 quantization configurations (Table~\ref{tab:formats}) on two architectures: an epoch-wise CNN and a temporal-context network. A third model, used as an ablation, has the same encoder as the temporal-context network but no temporal layers. We train each model with three seeds in 10-fold subject-wise cross-validation on all \val{data.subjects} subjects of Sleep-EDF-78. Every format is emulated value-exactly. Calibration uses only training subjects, so no information leaks from the test subjects. We compare the formats with per-subject paired tests, Holm correction, and non-inferiority bounds.

\item \textbf{Temporal context explains little of the 4-bit robustness.} With 4-bit weights and activations (W4A4), PTQ costs the epoch-wise CNN \val{w4a4.cnn.range} percentage points (pp) of Macro-F1, the unweighted mean of the per-stage F1 scores. The temporal-context network loses at most \val{w4a4.ctx.max}\pp{}. On Sleep-EDF-78, removing the temporal layers, while keeping the encoder and its training regime, preserves most of this robustness. At equal block size, the scaling recipe of NVFP4 (E4M3 block scales plus an FP32 tensor scale) outperforms the power-of-two block scales of MXFP4. Block size has smaller effects, and their sign varies.

\item \textbf{A training-free recipe.} The first layer, which quantizes the raw EEG, is the most sensitive layer in both networks. For the CNN, it accounts for most of the loss. On Sleep-EDF-78, keeping only the first and last layers in FP8 brings Macro-F1 within \val{mix.maxabs}\pp{} of FP32 and adds at most \val{mix.ovhmax}\% weight storage. In a direct paired comparison, this recipe is equivalent to quantization-aware training (QAT) within $\pm$1\pp{}. QAT beats PTQ applied after the same fine-tuning (significantly so for the CNN). However, QAT does not beat an FP32 model that received the same fine-tuning.

\item \textbf{A subject-level analysis.} We compare per-subject losses with a retraining-noise reference. The 4-bit losses fall on the same subjects across architectures. Among the five stages, N3 loses the most recall. We also repeat the study on the nested Sleep-EDF-20 subset to see which findings hold in a smaller cohort.

\item \textbf{Native measurements.} We report latency, throughput, and dynamic energy on an NVIDIA RTX PRO 6000 Blackwell GPU for the natively supported 16- and 32-bit formats. We add micro-benchmarks of FP8 general matrix multiplications (GEMMs).
\end{enumerate}

\section{Background and Related Work}
\subsection{Automatic sleep staging}
DeepSleepNet~\cite{supratak2017deepsleepnet} established sleep staging from raw single-channel EEG. It combines a CNN front end with recurrent sequence modeling. TinySleepNet~\cite{supratak2020tinysleepnet} kept this design and reduced the parameter count. Sequence-to-sequence models such as SeqSleepNet~\cite{phan2019seqsleepnet}, U-Time~\cite{perslev2019utime}, and SleepTransformer~\cite{phan2022sleeptransformer} also use dependencies across neighboring epochs, much as human scorers do. AttnSleep~\cite{eldele2021attnsleep}, in contrast, applies attention within individual epochs. Sleep-EDF~\cite{kemp2000sleepedf,goldberger2000physionet} is the most widely used public benchmark. Its 20-subject and 78-subject Sleep Cassette subsets (SC-20 and SC-78; here, Sleep-EDF-20 and Sleep-EDF-78) are standard evaluation cohorts, alongside datasets such as MASS-SS3~\cite{oreilly2014mass}. All of these works report FP32 accuracy.

\subsection{Efficient and quantized sleep staging}
DistillSleep~\cite{park2025distillsleep} distills a large teacher into a single-channel student with 109k parameters. The student runs in under 10\,ms per epoch on three embedded platforms. Hardware-oriented studies instead use integer or fixed-point arithmetic. SleepViT~\cite{robitaille2025sleepvit} is a vision-transformer accelerator synthesized in 65-nm CMOS. Its model has 31.6k weights and uses layer-dependent fixed-point quantization. Chen et al.~\cite{chen2026mixed} present a CNN-LSTM accelerator fabricated in 65-nm CMOS. It assigns INT2--INT16 precision layer by layer and keeps 5-class accuracy within 0.15\% of the floating-point baseline on MASS-SS3. A related system-on-chip paper with overlapping authors~\cite{xu2026inear} reports a 5.01\% accuracy drop for static 8-bit quantization of its CNN-LSTM and uses runtime dynamic quantization instead. These studies report cohort-level accuracy for a single deployment configuration of their own design. None of them evaluates block-scaled floating-point formats such as MXFP4 and NVFP4. None compares formats under a common protocol or analyzes how quantization affects individual subjects. This paper addresses these gaps.

\subsection{Low-precision formats}
Integer quantization maps real values to $b$-bit integer codes. A scale is shared across a tensor, a channel, or a group of values. Asymmetric quantization adds a zero point, the integer code that represents zero~\cite{jacob2018quantization,krishnamoorthi2018quantizing,nagel2021white}.

We write E$x$M$y$ for a signed floating-point format with one sign bit, $x$ exponent bits, and $y$ fraction (mantissa) bits. FP32, FP16, and BF16 (bfloat16) are E8M23, E5M10, and E8M7, respectively. The two common FP8 formats are E4M3 and E5M2. Their largest finite values are 448 and 57{,}344. E4M3 offers more precision, and E5M2 offers a wider range~\cite{micikevicius2022fp8}.

\begin{figure}[htbp]
\centering
\includegraphics[width=\figcol]{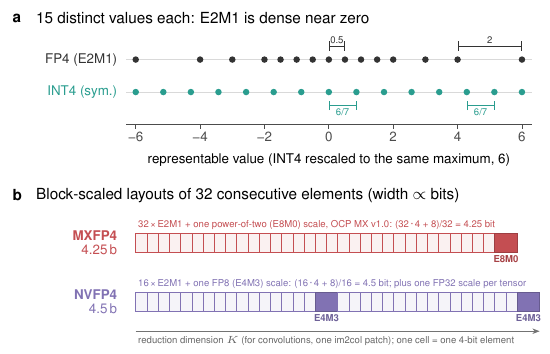}
\caption{4-bit encodings. (a) The 15 E2M1 values are dense near zero and sparse near the maximum, unlike the uniform grid of symmetric INT4 (rescaled here to the same maximum, 6). (b) MXFP4 shares one power-of-two (E8M0) scale among 32 elements. NVFP4 uses one FP8 (E4M3) scale per 16 elements plus an FP32 tensor scale, so it spends 0.25 bit more per element on scales. Cell widths are proportional to bits.}
\label{fig:formats}
\end{figure}

Microscaling formats give each block of elements a shared scale~\cite{ocp2023mx,rouhani2023microscaling}. MXFP4 uses blocks of 32 four-bit E2M1 elements and one 8-bit E8M0 scale per block. E8M0 has no sign or fraction bits, so its finite values are positive powers of two. Including the shared scale, a full block needs 4.25 bits per element. NVFP4 uses blocks of 16 E2M1 elements and one FP8 E4M3 scale per block. It adds one FP32 scale for the whole tensor~\cite{nvidia2025nvfp4}. A full block needs 4.5 bits per element, not counting the tensor scale and any padding (Fig.~\ref{fig:formats}). The OCP rule rounds the scale exponent down, so scaled values can exceed the element range and saturate. Mishra et al.~\cite{mishra2025recipes} report that this rule harms MXFP8 training, and they round the scale up instead. Chhugani et al.~\cite{chhugani2026mxfp4} propose overflow-aware scaling and macro-block scaling to reduce the quantization error of MXFP4.

Low-bit convolutional neural networks (CNNs) often keep the first and last layers at higher precision~\cite{choi2018pact,esser2020lsq}. More generally, the quantization sensitivity of each layer can guide its precision~\cite{dong2019hawq}. Quantization-aware training (QAT) simulates quantization during training. It usually passes gradients through the rounding step with a straight-through estimator, which treats the derivative of rounding as one~\cite{bengio2013estimating,nagel2021white}.

\subsection{Beyond average accuracy}
Compressed networks can keep their overall accuracy while performing much worse on a subset of inputs~\cite{hooker2019forget}. In compressed language models, many answers switch between correct and incorrect while overall accuracy barely changes~\cite{dutta2024accuracy}. Such switches are called \emph{flips}.

For sleep staging, we care about how quantization affects individual subjects. We therefore use the subject as the statistical unit. We also compare quantization-induced changes with a retraining reference. This reference is the variation between FP32 models of the same architecture, trained with different random seeds on the same data splits.

\section{Methods}
\begin{figure}
\centering
\includegraphics[width=\figcol]{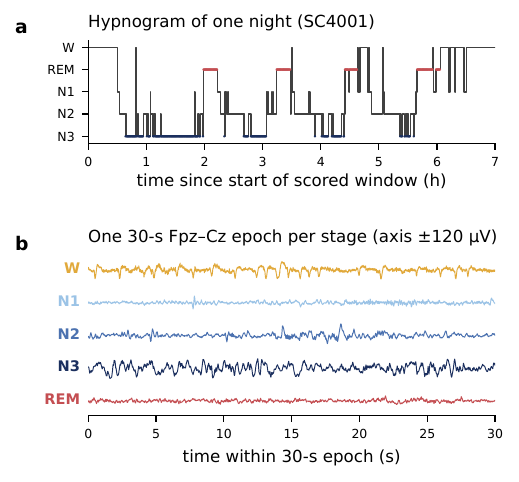}
\caption{Input data. (a) Expert hypnogram of one Sleep-EDF night (SC4001), restricted to the scored window (30\,minutes of wake before the first and after the last sleep epoch). N3 is drawn in dark blue and REM in red. (b) One representative 30-second Fpz--Cz epoch per stage (100\,Hz), drawn on a common $\pm$120\,$\mu$V axis. The models receive the raw epoch. W4A4 formats quantize it to 4 bits in the first layer.}
\label{fig:data}
\end{figure}

\subsection{Data}
We use the Sleep Cassette (SC) subset of the Sleep-EDF Expanded database~\cite{kemp2000sleepedf,goldberger2000physionet}. It contains \val{data.nights} overnight recordings from \val{data.subjects} healthy subjects aged 25--101 years. (Subject numbers 39, 68, 69, 78, and 79 do not exist in the SC numbering.) Figure~\ref{fig:data} shows an example night.

We use the Fpz--Cz EEG channel. It measures the voltage between the midline of the forehead (Fpz) and the vertex, the top of the head (Cz). Both positions follow the international 10--20 system~\cite{jasper1958tentwenty} (Appendix~\ref{app:background}). The signal is sampled at 100\,Hz, and we apply no further filtering. We cut each recording into non-overlapping 30-second epochs of $3{,}000$ samples and align them with the expert labels.

The labels follow the R\&K rules~\cite{rechtschaffen1968manual}. These rules split deep sleep into S3 and S4 by the share of the epoch that contains high-amplitude slow waves. For five-class staging, we merge S3 and S4 into N3, following the AASM grouping~\cite{iber2007aasm}. This step relabels the original scores; it does not rescore the recordings under AASM rules. We remove epochs labeled as movement time or unscored. We keep each night from 30\,minutes before the first sleep epoch to 30\,minutes after the last one, a common convention for Sleep-EDF~\cite{supratak2017deepsleepnet,phan2022sleeptransformer}. This limits the long stretches of wake at the start and end of each recording. Appendix~\ref{app:background} summarizes the EEG features of each stage and shows their spectra in our data.

The resulting dataset has \val{data.total} epochs: W \val{data.w} (\val{data.wpct}\%), N1 \val{data.nI} (\val{data.nIpct}\%), N2 \val{data.nII} (\val{data.nIIpct}\%), N3 \val{data.nIII} (\val{data.nIIIpct}\%), and REM \val{data.rem} (\val{data.rempct}\%). Our N1, N2, N3, and REM counts match those that Yang et al.~\cite{yang2023lwsleepnet} report for Sleep-EDF-78. Our W count is 10 epochs (about 0.02\%) lower than their $65{,}951$.

Removing movement and unscored epochs leaves gaps in the timeline of some nights. Such gaps occur in \val{gaps.withgap} of the \val{gaps.nights} recordings, with \val{gaps.n} gaps in total. The median gap is \val{gaps.medlen} epoch long, and the longest is \val{gaps.maxlen} epochs. For each night, we place the retained epochs in one sequence, in their original order, without placeholders. The last epoch before a gap is therefore followed directly by the first epoch after it. The temporal model predicts each epoch from a window of 31 retained epochs (Section~\ref{sec:models}). For \val{gaps.near} target epochs (\val{gaps.nearpct}\%), this window crosses at least one gap, so the model treats epochs separated by missing data as neighbors. The epoch-wise model sees only the target epoch and is not affected. Section~\ref{sec:controls} tests a gap-aware variant.

\subsection{Evaluation protocol}\label{sec:protocol}
We split the data by subject. Within each fold, all nights of a subject stay in the same subset. Recordings of one person therefore never appear in both the training data and the held-out data. We divide the \val{data.subjects} subjects into 10 disjoint test folds of 7--8 subjects. For each fold, we randomly draw seven validation subjects from the 70 or 71 remaining subjects. The validation subjects are used for early stopping and model selection. The other 63 or 64 subjects form the fitting set, which we call FIT. Only FIT subjects update the network weights.

All data-dependent preprocessing uses FIT only. For each fold, we compute one EEG mean and standard deviation (SD) from FIT. We use them to z-score the FIT, validation, and test signals. We then clip the normalized signal to $[-20,20]$, that is, to 20 FIT SDs around the FIT mean. Clipping limits the effect of extreme amplitudes, such as movement or electrode artifacts, on the quantization ranges. The cross-entropy class weights are $w_c=N/(5\,n_c)$, where $N$ is the number of FIT epochs and $n_c$ is the number of FIT epochs in class $c$. The PTQ calibration set contains \val{data.calib} FIT epochs, \val{data.calibper} per stage. As a result, no information from validation or test subjects enters normalization, class weighting, or calibration. Test subjects are used only for evaluation.

SleepCNN is calibrated on the \val{data.calib} selected epochs directly. For ContextSleepNet, each selected epoch is the center of a 31-epoch window from the same night. (Near the start or end of a night, the window is shifted so that it stays inside the night.) Per fold, the windows contain \val{cal.positions} epoch positions. Because windows can overlap, these positions cover \val{cal.distmin}--\val{cal.distmax} distinct FIT epochs. The center epochs are balanced across stages, but the full windows are not. For example, N2 makes up \val{cal.nII}\% of the window positions and N1 only \val{cal.nI}\%. Activation statistics are collected from the full windows. Section~\ref{sec:formats} describes how the search for the clipping threshold subsamples them.

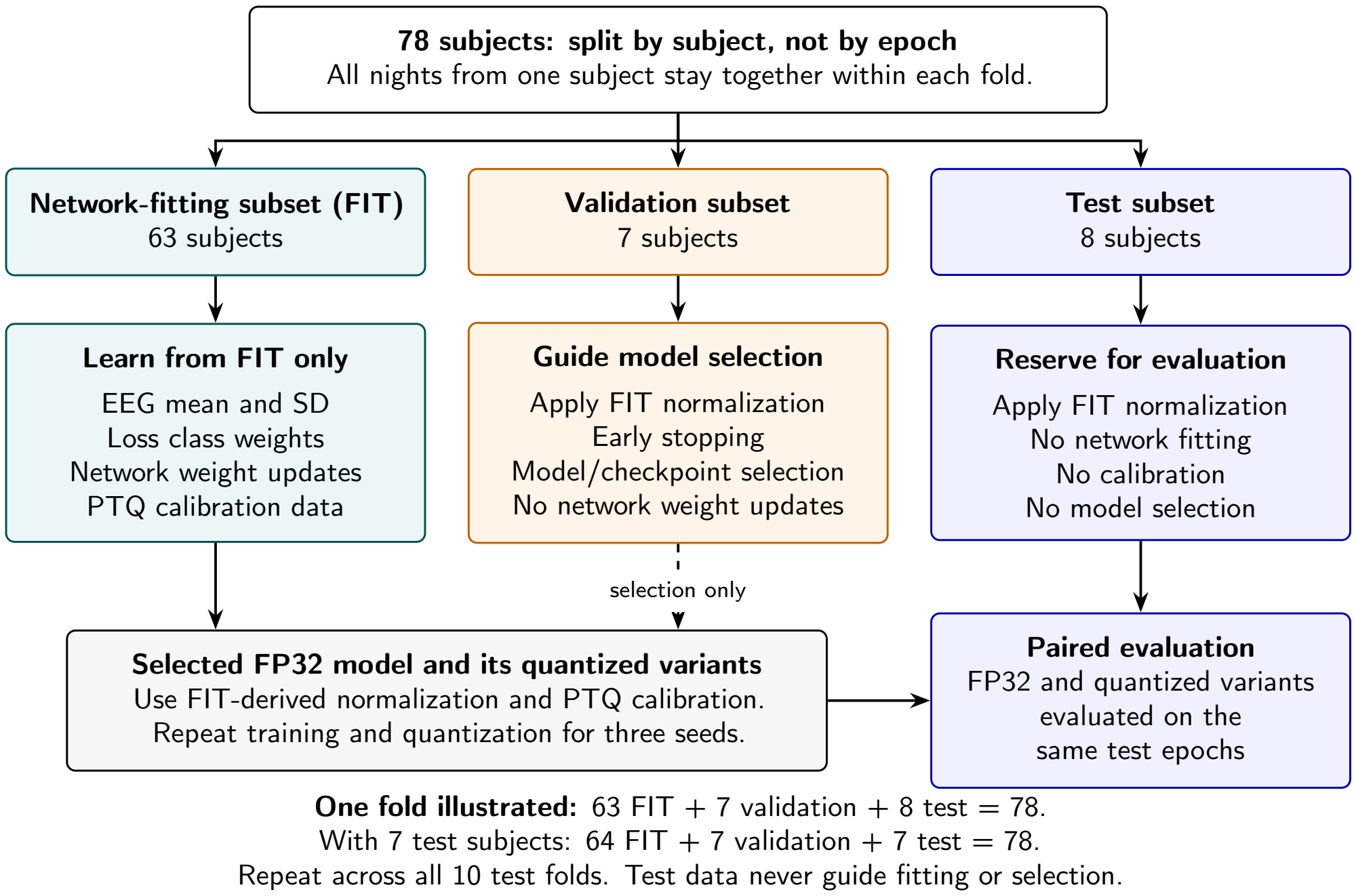
\begin{figure}[htbp]
  \centering
  \resizebox{0.8\textwidth}{!}{
\begin{tikzpicture}[
  font=\sffamily\small,
  box/.style={draw, rounded corners=3pt, align=center,
    text width=4.6cm, inner sep=8pt, line width=0.7pt},
  flow/.style={-{Stealth[length=2.5mm]}, line width=0.9pt},
  fitbox/.style={box, draw=teal!65!black, fill=teal!7},
  valbox/.style={box, draw=orange!75!black, fill=orange!9},
  testbox/.style={box, draw=blue!65!black, fill=blue!6}
]
\node[box, text width=10cm] (all) at (5.7,0)
  {\textbf{78 subjects: split by subject, not by epoch}\\
   All nights from one subject stay together within each fold.};

\node[fitbox] (fit) at (0,-2)
  {\textbf{Network-fitting subset (FIT)}\\63 subjects};
\node[valbox] (val) at (5.7,-2)
  {\textbf{Validation subset}\\7 subjects};
\node[testbox] (test) at (11.4,-2)
  {\textbf{Test subset}\\8 subjects};
\draw[flow] (all.south) -- (5.7,-1) -| (fit.north);
\draw[flow] (all.south) -- (val.north);
\draw[flow] (all.south) -- (5.7,-1) -| (test.north);

\node[fitbox, minimum height=2.6cm] (fituse) at (0,-4.6)
  {\textbf{Learn from FIT only}\\[4pt]
   EEG mean and SD\\
   Loss class weights\\
   Network weight updates\\
   PTQ calibration data};
\node[valbox, minimum height=2.6cm] (valuse) at (5.7,-4.6)
  {\textbf{Guide model selection}\\[4pt]
   Apply FIT normalization\\
   Early stopping\\
   Model/checkpoint selection\\
   No network weight updates};
\node[testbox, minimum height=2.6cm] (testuse) at (11.4,-4.6)
  {\textbf{Reserve for evaluation}\\[4pt]
   Apply FIT normalization\\
   No network fitting\\
   No calibration\\
   No model selection};
\draw[flow] (fit) -- (fituse);
\draw[flow] (val) -- (valuse);
\draw[flow] (test) -- (testuse);

\node[box, text width=8.8cm, minimum height=1.6cm,
      fill=black!3] (models) at (2.85,-7.9)
  {\textbf{Selected FP32 model and its quantized variants}\\
   Use FIT-derived normalization and PTQ calibration.\\
   Repeat training and quantization for three seeds.};
\node[testbox, minimum height=1.6cm] (eval) at (11.4,-7.9)
  {\textbf{Paired evaluation}\\
   FP32 and quantized variants\\
   evaluated on the same test epochs};
\draw[flow] (fituse.south) -- (fituse.south |- models.north);
\draw[flow, dashed] (valuse.south) -- (valuse.south |- models.north);
\node[font=\sffamily\scriptsize, fill=white] at (5.7,-6.55)
  {selection only};
\draw[flow] (models.east) -- (eval.west);
\draw[flow] (testuse.south) -- (eval.north);

\node[align=center, text width=16cm, font=\sffamily\small]
  at (5.7,-9.65)
  {\textbf{One fold illustrated:} 63 FIT + 7 validation + 8 test = 78.\\
   With 7 test subjects: 64 FIT + 7 validation + 7 test = 78.\\
   Repeat across all 10 test folds. Test data never guide fitting or selection.};
\end{tikzpicture}}
  \caption{Subject-wise separation for one cross-validation fold. The fitting subset (FIT) updates the network weights and also supplies the normalization statistics, the class weights, and the post-training quantization (PTQ) calibration data. Validation subjects guide early stopping and model selection. Test subjects are used only for evaluation. The illustrated fold has eight test subjects; folds with seven test subjects have 64 FIT subjects.}
  \label{fig:subject-split}
\end{figure}

We train each model with three random seeds per fold. For each fold and seed, all quantized variants derive from the same trained FP32 model. That FP32 model and its quantized variants are evaluated on the same test epochs. Figure~\ref{fig:subject-split} summarizes the roles of the three subsets.

\subsection{Models} \label{sec:models}
\textbf{SleepCNN} classifies each 30-second epoch on its own. It has \val{cnn.params} parameters and needs \val{cnn.macs} million multiply--accumulate operations (MACs) per epoch. The first convolution has 16 filters with a kernel of 50 samples and a stride of 5. The second has 32 filters with a kernel of 25 samples and a stride of 3. Each convolution is followed by a rectified linear unit (ReLU) and max-pooling by a factor of 4. The resulting 320 features pass through two fully connected layers ($320\to128\to5$), which give one output per stage.

\textbf{ContextSleepNet} also uses neighboring epochs. It has \val{ctx.params} parameters and needs \val{ctx.macs} million MACs per epoch. A per-epoch encoder, inspired by TinySleepNet~\cite{supratak2020tinysleepnet}, first processes each epoch. Its first convolution has 64 filters with a kernel of 50 samples and a stride of 6. Two further convolutions have 64 filters and a kernel of 7 samples. The encoder also uses batch normalization (BN), max-pooling by factors of 8 and 4, and dropout. A linear layer maps the resulting 960 features to a 128-dimensional embedding.

Four residual temporal convolution layers then process the sequence of embeddings. Each layer has 128 channels and a kernel size of 3. Their dilations are 1, 2, 4, and 8, which set the spacing between the epochs that each convolution combines. The receptive field therefore grows from 3 to 7, 15, and finally 31 epochs. A linear classifier outputs five stage scores per epoch. The final receptive field covers the target epoch and up to 15 retained epochs on each side. For contiguous epochs, this is 7.5\,min of context on each side. Across a gap, the window spans more time.

The model is fully convolutional along the epoch axis, so it handles sequences of any length. We train it on 32-epoch windows and evaluate each night in one forward pass. Because it uses future epochs, the model is non-causal. It is meant for offline scoring of a whole night, not for online staging.

Figure~\ref{fig:arch} in Appendix~\ref{app:background} shows SleepCNN, ContextSleepNet, and the ablation model EncoderNet (Section~\ref{sec:controls}). It gives their tensor shapes and shows how the receptive field grows ($3\to7\to15\to31$ epochs).

\subsection{Training}\label{sec:training}
Both models use Adam~\cite{kingma2015adam} with an initial learning rate of $10^{-3}$, a weight decay of $10^{-4}$, and a cosine learning-rate schedule. After each training epoch, we measure validation Macro-F1 and keep the checkpoint with the highest value. Training stops after eight training epochs without improvement or at the budget of 30 training epochs (SleepCNN) or 40 (ContextSleepNet).

SleepCNN uses mini-batches of 64 EEG epochs. ContextSleepNet uses dropout with a probability of 0.5 and mini-batches of 16 windows, each of 32 consecutive retained EEG epochs. The windows are drawn at random within nights. Each training epoch draws enough windows to cover the FIT epochs once on average.

Across runs, the median training epoch of the selected checkpoint was \val{cnn.epmed} for SleepCNN and \val{ctx.epmed} for ContextSleepNet. The full budget, without early stopping, was used by \val{cnn.fullbudget} of the \val{cnn.nruns} SleepCNN runs and by \val{ctx.fullbudget} of the \val{ctx.nruns} ContextSleepNet runs.

We disable TensorFloat-32 (TF32) matrix arithmetic, so the FP32 reference is true FP32. The main study covered both models, 10 test folds, three seeds per fold, and all evaluated conditions. All experiments ran on one NVIDIA RTX PRO 6000 Blackwell Server Edition GPU with PyTorch 2.11 and CUDA 13.0.

\begin{table}[t]
\centering
\caption{Evaluated formats. Scales are given for weights and for activations. Bits/w: bits per weight, including block or channel scales ($K$: reduction length of the layer); the single FP32 tensor scale of NVFP4 is not counted. The INT4-g32 emulation keeps FP32 group scales. A: activations. $^{\dagger}$Non-standard block size, used as an ablation.}
\label{tab:formats}
\setlength{\tabcolsep}{2.5pt}
\footnotesize
\fitcol{%
\begin{tabular}{l l l r}
\toprule
Format & Element & Scale (weights; activations) & Bits/w \\
\midrule
FP16, BF16 & E5M10, E8M7 & none & 16 \\
FP8-E4M3 & E4M3 & FP32/channel; static/tensor & $8+32/K$ \\
FP8-E5M2 & E5M2 & FP32/channel; static/tensor & $8+32/K$ \\
INT8 & INT8 sym. & FP32/channel; static/tensor & $8+32/K$ \\
INT8 (W only) & INT8 sym. & FP32/channel; A in FP32 & $8+32/K$ \\
INT4-g32 (W only) & INT4 asym. & FP32 + 4-bit zero/32; A in BF16 & 5.125 \\
MXFP4-32 & E2M1 & E8M0/32; dynamic E8M0/32 & 4.25 \\
MXFP4-16$^{\dagger}$ & E2M1 & E8M0/16; dynamic E8M0/16 & 4.5 \\
MXFP4-32 (W only) & E2M1 & E8M0/32; A in FP32 & 4.25 \\
NVFP4-16 & E2M1 & E4M3/16 + FP32/tensor; dyn.\ E4M3/16 + static FP32 & 4.5 \\
NVFP4-32$^{\dagger}$ & E2M1 & E4M3/32 + FP32/tensor; dyn.\ E4M3/32 + static FP32 & 4.25 \\
NVFP4-16 (W only) & E2M1 & E4M3/16 + FP32/tensor; A in FP32 & 4.5 \\
\bottomrule
\end{tabular}}
\end{table}

\subsection{Number formats and quantized inference}\label{sec:formats}
Table~\ref{tab:formats} lists the evaluated formats. We quantize the weights and the input activations of every convolutional and fully connected layer. This includes the first layer, whose input is the normalized raw EEG. Before quantization, we fold BN into the preceding convolution. Biases stay in FP32. ReLU, pooling, and the residual additions also run in FP32. The skip path of each residual temporal layer therefore carries unquantized values. Only the copy that enters a convolution or the classifier is quantized.

We quantize along the reduction dimension $K$, the set of values that are summed to produce one output. For a one-dimensional convolution, $K$ is the number of input channels times the kernel width. It is thus the length of one flattened filter and of its input patch. For a fully connected layer, $K$ is the number of input features.

\textbf{INT8 and FP8.} Weights use one scale per output channel:
\begin{equation}
s_o = \frac{\max_k |w_{o,k}|}{q_{\max}},
\end{equation}
where $o$ indexes the output channel, $k$ indexes $K$, and $q_{\max}$ is the largest value of the format (127 for INT8, 448 for E4M3, and 57{,}344 for E5M2). Activations use one fixed clipping threshold per tensor, chosen on calibration data. Let $a_{\max}$ be the largest activation magnitude seen in all calibration batches. We pick the threshold $c$ that minimizes the mean squared quantization error over
\begin{equation}
c \in \{0.30,0.32,\ldots,1.00\}\,a_{\max}.
\end{equation}
The error is computed on a buffer of at most $2\times10^{6}$ activation values per tensor, not on every calibration activation.
Each calibration batch adds up to $2\times10^{5}$ randomly sampled values, and the oldest batches are dropped when the buffer is full. For SleepCNN, the buffer covers all batches. For EncoderNet and ContextSleepNet, it covers only the last 10 to 15 of the 32 batches. The windows are processed in subject order, so these batches come from the FIT subjects with the highest subject numbers. Values beyond the threshold are clipped. Scaled values beyond the finite range of the format saturate at its endpoints. FP8 conversion rounds to the nearest value, with ties to even.

\textbf{INT4-g32.} This weight-only format uses asymmetric quantization in groups of 32 weights. Each group stores an FP32 scale and a 4-bit zero point. A full group therefore costs \val{int4.bits} bits per weight, not counting padding. The activations of this condition stay in BF16.

\textbf{MXFP4.} We follow the power-of-two scaling rule of the OCP MX v1.0 specification~\cite{ocp2023mx}. Each block $\mathcal{B}$ holds 32 consecutive values along $K$. Its maximum magnitude is
\begin{equation}
a_{\mathcal{B}} = \max_{i\in\mathcal{B}} |x_i|.
\end{equation}
For $a_{\mathcal{B}}>0$ and a scale exponent within the E8M0 range, the shared scale and the reconstructed values are
\begin{align}
s_{\mathcal{B}}
&= 2^{\lfloor \log_2 a_{\mathcal{B}} \rfloor - 2},\\
\hat{x}_i
&= s_{\mathcal{B}}\,
Q_{\mathrm{E2M1}}\!\left(\frac{x_i}{s_{\mathcal{B}}}\right).
\end{align}
Here, $Q_{\mathrm{E2M1}}$ rounds to the nearest value in
\[
\{0,\pm0.5,\pm1,\pm1.5,\pm2,\pm3,\pm4,\pm6\}.
\]
Ties go to the even value, and values beyond the range saturate at $\pm6$. Activation scales are computed for each block at run time.

\textbf{NVFP4.} Each block holds 16 consecutive values along $K$. Quantization uses an FP32 tensor scale and an E4M3 block scale~\cite{nvidia2025nvfp4}. Let $A_t>0$ be the amplitude reference of the tensor. For weights, $A_t$ is the largest weight magnitude. For activations, $A_t$ is the largest magnitude seen during calibration, and it stays fixed afterward. The scales are
\begin{align}
s_t &= \frac{A_t}{448\times 6},\\
s_{\mathcal{B}}
&= Q_{\mathrm{E4M3}}\!\left(
\operatorname{clip}\!\left(
\frac{a_{\mathcal{B}}}{6s_t},
2^{-6},
448
\right)\right).
\end{align}
The clip operation limits its argument to the given interval. The lower bound $2^{-6}$ is an implementation choice that keeps block scales at positive normal E4M3 values. The reconstructed values are
\begin{equation}
\hat{x}_i =
s_t s_{\mathcal{B}}\,
Q_{\mathrm{E2M1}}\!\left(
\frac{x_i}{s_t s_{\mathcal{B}}}
\right).
\end{equation}
We evaluate the argument of $Q_{\mathrm{E2M1}}$ as $x_i\cdot\bigl((1/s_t)/s_{\mathcal{B}}\bigr)$, which is the operation order of the TorchAO reference. For activations, $s_t$ is fixed, and the block scales are computed at run time.

Activation blocks are formed along $K$ within each input patch. A block therefore never mixes values from different output positions. Neighboring patches of a convolution overlap, so an input value that belongs to several patches is quantized separately in each one, with the block scale of that patch. In the first layer, each EEG sample enters up to 10 patches in SleepCNN (kernel 50, stride 5) and up to 9 in the larger encoder (stride 6). The emulation thus models the quantization of an unfolded matrix-multiplication operand. It does not model an activation tensor that is stored once in 4 bits. On the tested inputs, both 4-bit quantizers match the TorchAO reference quantizers~\cite{or2025torchao} bit for bit (Section~\ref{sec:controls}).

We emulate all formats, including FP16 and BF16, by fake quantization. Weights and activations are rounded to the values of the format and then stored in FP32. All later arithmetic, including accumulation, is FP32. Our accuracy results therefore capture the effect of the number format alone. They do not model the accumulator precision, operation order, intermediate rounding, or operator fusion of a specific low-precision kernel. For FP16 and BF16, we also check the emulation against native GPU execution (Section~\ref{sec:controls}). We measure speed only for formats with native kernels.

\subsection{Recovering 4-bit accuracy}
We test two mixed-precision variants that need no retraining. Both start from W4A4. The first runs the weights and input activations of the first layer in FP8-E4M3. The second does the same for both the first and the last layer. These layers thus use W8A8, and all other layers stay at W4A4. We call the second variant \emph{FP8 first/last}.

For quantization-aware training (QAT), we start from the trained FP32 model with BN folded into the convolutions. We fine-tune it for five training epochs at a learning rate of $10^{-4}$. Fine-tuning uses the same training loop as above (Adam, cosine schedule, and the best-validation checkpoint). The forward pass simulates 4-bit quantization, and a straight-through estimator passes gradients through the quantizer~\cite{bengio2013estimating}. After fine-tuning, we convert the model with the same procedure as PTQ.

During NVFP4 fine-tuning, the activation tensor scale is computed on the fly from the largest magnitude of each activation tensor. After conversion, the model uses a fixed activation tensor scale calibrated on FIT subjects, as in PTQ. The tensor-scale estimate thus differs between fine-tuning and evaluation.

QAT also adds training. To separate the two effects, we include an FP32 fine-tuning control, \emph{FP32+FT}. It receives the same five-epoch fine-tuning without simulated quantization. We also apply the same PTQ to the FP32+FT model, which gives the condition \emph{FT$\to$PTQ}. Comparing QAT with FT$\to$PTQ separates the benefit of simulating quantization during fine-tuning from the benefit of extra training. We run QAT, FP32+FT, and FT$\to$PTQ for all three seeds.

\subsection{Control experiments}
\label{sec:controls}
\textbf{Same-encoder ablation.} \emph{EncoderNet} is ContextSleepNet without its four temporal convolution layers. It has \val{enc.params} parameters. It keeps the same encoder, classifier, BN, and dropout. It also keeps the training windows, optimizer settings, training budget, early-stopping rule, and calibration windows. At inference, it classifies each epoch from that epoch's 30-second signal alone. We train it with the same folds and seeds and evaluate it under the same PTQ conditions.

For each model, $\Delta$MF1 is the change in Macro-F1 relative to that model's own FP32 baseline. We compute it per subject and compare it between models on the same subjects. The \emph{temporal-layer effect} is ContextSleepNet's $\Delta$MF1 minus EncoderNet's $\Delta$MF1. It captures everything that the temporal layers add: their use of neighboring epochs and their effect on capacity and optimization. The \emph{non-temporal effect} is EncoderNet's $\Delta$MF1 minus SleepCNN's $\Delta$MF1. It combines the differences between the two encoders and between the two training procedures. (EncoderNet uses mini-batches of 16 windows of 32 epochs and a budget of 40 training epochs. SleepCNN uses mini-batches of 64 epochs and a budget of 30; see Section~\ref{sec:training}.) This ablation cannot separate the encoder from its training procedure. We Holm-correct the eight W4A4 contrasts (two effects for each of four formats).

\textbf{First-layer probes.} We quantize only the first layer, in one of three ways: weights only, input activations only, or both. All other layers stay in FP32. These probes separate the effect of quantizing the first-layer weights from that of quantizing the raw EEG input.

\textbf{Recording gaps.} We split each night at every internal gap. We then retrain and evaluate ContextSleepNet on the resulting segments. Training windows stay within a segment, and predictions never use epochs from across a gap. We call this variant the \emph{gap-aware} model. The change applies only to training and evaluation. PTQ calibration reuses the windows of the main model, which are drawn from unsplit nights and may cross gaps.

\textbf{Emulation checks.} We run the BN-folded models natively in FP16 and BF16 on the GPU and compare them with the emulated formats. We also compare our MXFP4 and NVFP4 quantizers with the TorchAO~\cite{or2025torchao} reference implementations. The inputs are heavy-tailed random tensors and every trained weight matrix of fold 0, seed 0 of both models. TorchAO is pinned to version \val{tao.version} and runs inside the notebook. The check fails, rather than being skipped, if any pair cannot be computed or differs. Finally, reloaded checkpoints, the rerun FP32+FT control, and the rerun single-layer probes must reproduce the stored predictions exactly.

\textbf{Sleep-EDF-20.} We train and evaluate all three models with the same pipeline on subjects 0--19 (the SC-20 subset). We use 10 folds of two test subjects and three seeds.

\subsection{Statistics}
Our primary metric is Macro-F1 (MF1), the unweighted mean of the five per-stage F1 scores. It gives each stage equal weight. This matters here because W and N2 make up most epochs. We also report accuracy, per-stage F1, and Cohen's kappa~\cite{cohen1960kappa}:
\begin{equation}
\kappa = \frac{p_o-p_e}{1-p_e}.
\end{equation}
Here, $p_o$ is the observed agreement between predictions and expert labels. The quantity $p_e$ is the agreement expected by chance, computed from the stage frequencies of the predictions and the labels.

\emph{Pooled} metrics combine all test epochs of the cohort for each seed. We then average them over the three seeds.

For statistical inference, the subject is the unit of analysis.
Per-subject MF1 always averages five stages. If a stage does not occur in a subject's expert labels, its F1 is set to zero. For each subject and seed, we subtract the FP32 MF1 from the MF1 under the tested condition. We average these paired differences over the three seeds, which gives one $\Delta$MF1 value per subject. Negative values mean a loss. We report the mean over the \val{data.subjects} subjects with a 95\% percentile bootstrap confidence interval (CI). The CI uses $5{,}000$ resamples of subjects drawn with replacement~\cite{efron1993bootstrap}.

We also test the per-subject differences with a Wilcoxon signed-rank test~\cite{wilcoxon1945}. We adjust its $p$-values with Holm's method across the 20 conditions of each model~\cite{holm1979simple}. The two analyses answer different questions. The bootstrap CI describes the uncertainty of the mean. The Wilcoxon test uses the signs and ranks of the differences, and its usual reading as a location shift assumes symmetric differences. Its associated Hodges--Lehmann pseudomedian is a location measure, not the mean. The CI and the test can therefore disagree. We report both and read the rank test with caution when the differences are skewed.

The 95\% bootstrap CIs are pointwise: each one describes a single condition without a multiplicity adjustment. We handle multiplicity separately, through the Holm-adjusted tests and the adjusted bounds for non-inferiority and equivalence.

A nonsignificant test does not show that no meaningful loss exists. We therefore also test \emph{non-inferiority} with a margin of \val{ni.margin}\pp{}. Let $m=\val{ni.m}$ be the size of the Bonferroni family for each model. From $20{,}000$ bootstrap resamples of subjects, we take the $0.05/m$ quantile of the resampled means as a one-sided lower bound. A condition is non-inferior if this bound exceeds $-\val{ni.margin}$\pp{}. This procedure targets a familywise error rate of 0.05, to the extent that the bootstrap approximation holds. We chose the margin after seeing the main results, so the non-inferiority analysis is exploratory. The margin is of the same order as the SD of the FP32 pooled MF1 across training seeds. This SD ranges from \val{fp32.sd.min} to \val{fp32.sd.max}\pp{} across the three models.

We also test \emph{equivalence} between two conditions from their paired per-subject MF1 differences. Two conditions are equivalent if the adjusted lower bound exceeds $-\val{ni.margin}$\pp{} and the adjusted upper bound is below $+\val{ni.margin}$\pp{}. This test uses the same exploratory margin. For the comparison of FP8 first/last with QAT, the Bonferroni family contains the four model--format pairs.

We also check how sensitive the non-inferiority decisions are to the interval method. For this check, we recompute the lower bounds with a bias-corrected and accelerated (BCa) bootstrap and with a $t$ interval. Both use the same multiplicity adjustment.

The \emph{flip rate} is the share of test epochs whose predicted stage differs from the FP32 prediction. It also counts changes from one wrong stage to another. We compute it over all test epochs for each seed and then average it over seeds.

To put per-subject changes in context, we build a \emph{retraining-noise} reference from FP32 models trained with different seeds on the same folds. Three seeds give three seed pairs. We include each pair in both directions, so the reference is symmetric around zero by construction. This reference shows how much results vary between training runs. Reversing a pair does not create an independent observation.

The subject-level correlation analyses in Section~\ref{sec:subject} were not planned in advance and are exploratory.

\subsection{Cost model and hardware measurements}
We compute packed weight storage, including block and channel scales, block padding, and FP32 biases (Table~\ref{tab:formats}). The single FP32 tensor scale of NVFP4 is negligible and is not counted. We also report a theoretical bit-operation count (BOPs):
\begin{equation}
\mathrm{BOPs}
=
\sum_{\ell}
\mathrm{MACs}_{\ell}\,
b_{w,\ell}\,
b_{a,\ell},
\end{equation}
where $\mathrm{MACs}_{\ell}$ is the number of multiply--accumulate operations in layer $\ell$, and $b_{w,\ell}$ and $b_{a,\ell}$ are its weight and activation bit widths. BOPs is a proxy for compute cost, not a measurement of run time or energy.

We benchmark native inference in FP32, FP16, and BF16. Before benchmarking, we fold BN into the convolutions and cast each model to the target data type. We enable cuDNN autotuning and measure with and without CUDA graphs. Each configuration gets 50 warm-up calls and 200 timed calls. CUDA events time each call on the GPU, and we report the median. For SleepCNN, a call processes a batch of independent epochs. For ContextSleepNet, a call processes one sequence of consecutive epochs.

We estimate dynamic GPU energy with the cumulative energy counter of the NVIDIA Management Library (NVML). At the largest call size, we run inference with CUDA graphs continuously for 5\,s. We take the difference of the counter readings over this interval. We then subtract the energy of an idle interval of the same length.

We also benchmark FP8 general matrix multiplications (GEMMs) with \texttt{torch.\_scaled\_mm} and per-tensor scales. We compare them with BF16 and FP32 GEMMs of the same shapes. The shapes are the im2col matrices of the three layers with the most MACs, at 256 epochs per call. $K$ and $N$ are padded to multiples of 16. We report the mean wall-clock time of 200 calls after 20 warm-up calls. These measurements cover selected matrix multiplications, not complete FP8 inference.

\section{Results}
\begin{table}
\centering
\caption{FP32 baselines on Sleep-EDF-78 (single-channel EEG) next to published results. Published values are from Table~I of~\cite{phan2022sleeptransformer}; the SleepTransformer row is the model trained from scratch (without SHHS pretraining). Protocols differ (data split, loss weighting, augmentation), so the comparison is only indicative. Ours: pooled over the 78 test subjects, mean $\pm$ SD over three seeds.}
\label{tab:baselines}
\setlength{\tabcolsep}{3pt}
\footnotesize
\begin{tabular}{l r r r r}
\toprule
Model & Params & Acc (\%) & MF1 (\%) & $\kappa$ \\
\midrule
SeqSleepNet~\cite{phan2019seqsleepnet} & -- & 82.6 & 76.4 & 0.760 \\
AttnSleep~\cite{eldele2021attnsleep} & -- & 81.3 & 75.1 & 0.74 \\
SleepTransformer~\cite{phan2022sleeptransformer} & 3.7\,M & 81.4 & 74.3 & 0.743 \\
\midrule
SleepCNN (ours) & 55\,k & \val{cnn.fp32.acc} & \val{cnn.fp32.mf1}\,$\pm$\,\val{cnn.fp32.mf1sd} & \val{cnn.fp32.kappa} \\
ContextSleepNet (ours) & 382\,k & \val{ctx.fp32.acc} & \val{ctx.fp32.mf1}\,$\pm$\,\val{ctx.fp32.mf1sd} & \val{ctx.fp32.kappa} \\
\bottomrule
\end{tabular}
\end{table}

\subsection{FP32 baselines}
Table~\ref{tab:baselines} compares our FP32 models with published single-channel results on Sleep-EDF-78. ContextSleepNet reaches \val{ctx.fp32.mf1}\% Macro-F1 and $\kappa=\val{ctx.fp32.kappa}$. These values are close to those reported for AttnSleep and for SleepTransformer trained from scratch, and below those of SeqSleepNet, the strongest baseline in the table. Evaluation protocols differ, however, so the comparison only provides context.

The accuracy of ContextSleepNet is \val{ctx.fp32.acc}\%, which is 1--3\pp{} below the published models in Table~\ref{tab:baselines}. The model predicts N3 for \val{ctx.fp32.nIIIsharepred}\% of epochs, although N3 makes up only \val{data.nIIIpct}\% of the expert labels. Class weighting may contribute to this over-prediction and to the lower accuracy. We did not train an unweighted control, so we cannot measure this contribution.

SleepCNN is a compact model without temporal context. It reaches \val{cnn.fp32.mf1}\% Macro-F1 and $\kappa=\val{cnn.fp32.kappa}$. Across the three training seeds, the SD of pooled Macro-F1 is \val{cnn.fp32.mf1sd}\pp{} for SleepCNN and \val{ctx.fp32.mf1sd}\pp{} for ContextSleepNet.

\begin{table*}
\centering
\caption{Post-training quantization on Sleep-EDF-78 (Fpz--Cz, 10-fold subject-wise CV, 3 training seeds). MF1: pooled Macro-F1 (\%), mean over seeds. $\Delta$MF1: per-subject paired difference to FP32 in percentage points (pp), averaged over seeds, mean and bootstrap 95\% CI over $n=78$ subjects; $^{*}$/$^{**}$: Holm-corrected Wilcoxon signed-rank $p<0.05/0.01$ (family of 20 conditions per model). Flip: \% of 30-s epochs whose predicted stage differs from FP32 (pooled over test epochs per seed, mean over seeds). KiB: packed weights including scale metadata and FP32 biases. W/A: weight/activation bits. $^{\dagger}$Non-standard block size (ablation). All formats are emulated: values are rounded exactly to each format, and the arithmetic is FP32 (native FP16/BF16 is checked in Section~\ref{sec:controls}).}
\label{tab:main}
\setlength{\tabcolsep}{3.2pt}
\small
\fitwide{%
\begin{tabular}{lc rrlr rrlr}
\toprule
 & & \multicolumn{4}{c}{SleepCNN (55\,k parameters)} & \multicolumn{4}{c}{ContextSleepNet (382\,k parameters)} \\
\cmidrule(lr){3-6}\cmidrule(lr){7-10}
Format & W/A & KiB & MF1 & $\Delta$MF1 [95\% CI] & Flip & KiB & MF1 & $\Delta$MF1 [95\% CI] & Flip \\
\midrule
FP32 & 32/32 & \val{cnn.fp32.kib} & \val{cnn.fp32.mf1} & -- & -- & \val{ctx.fp32.kib} & \val{ctx.fp32.mf1} & -- & -- \\
\midrule
FP16 & 16/16 & \val{cnn.fp16.kib} & \val{cnn.fp16.mf1} & $\val{cnn.fp16.d}$ {\scriptsize[$\val{cnn.fp16.dlo}$, $\val{cnn.fp16.dhi}$]} & \val{cnn.fp16.flip} & \val{ctx.fp16.kib} & \val{ctx.fp16.mf1} & $\val{ctx.fp16.d}$ {\scriptsize[$\val{ctx.fp16.dlo}$, $\val{ctx.fp16.dhi}$]} & \val{ctx.fp16.flip} \\
BF16 & 16/16 & \val{cnn.bf16.kib} & \val{cnn.bf16.mf1} & $\val{cnn.bf16.d}$ {\scriptsize[$\val{cnn.bf16.dlo}$, $\val{cnn.bf16.dhi}$]} & \val{cnn.bf16.flip} & \val{ctx.bf16.kib} & \val{ctx.bf16.mf1} & $\val{ctx.bf16.d}$ {\scriptsize[$\val{ctx.bf16.dlo}$, $\val{ctx.bf16.dhi}$]} & \val{ctx.bf16.flip} \\
\midrule
FP8-E4M3 & 8/8 & \val{cnn.fp8e4m3w8a8.kib} & \val{cnn.fp8e4m3w8a8.mf1} & $\val{cnn.fp8e4m3w8a8.d}$ {\scriptsize[$\val{cnn.fp8e4m3w8a8.dlo}$, $\val{cnn.fp8e4m3w8a8.dhi}$]}$^{*}$ & \val{cnn.fp8e4m3w8a8.flip} & \val{ctx.fp8e4m3w8a8.kib} & \val{ctx.fp8e4m3w8a8.mf1} & $\val{ctx.fp8e4m3w8a8.d}$ {\scriptsize[$\val{ctx.fp8e4m3w8a8.dlo}$, $\val{ctx.fp8e4m3w8a8.dhi}$]} & \val{ctx.fp8e4m3w8a8.flip} \\
FP8-E5M2 & 8/8 & \val{cnn.fp8e5m2w8a8.kib} & \val{cnn.fp8e5m2w8a8.mf1} & $\val{cnn.fp8e5m2w8a8.d}$ {\scriptsize[$\val{cnn.fp8e5m2w8a8.dlo}$, $\val{cnn.fp8e5m2w8a8.dhi}$]}$^{**}$ & \val{cnn.fp8e5m2w8a8.flip} & \val{ctx.fp8e5m2w8a8.kib} & \val{ctx.fp8e5m2w8a8.mf1} & $\val{ctx.fp8e5m2w8a8.d}$ {\scriptsize[$\val{ctx.fp8e5m2w8a8.dlo}$, $\val{ctx.fp8e5m2w8a8.dhi}$]} & \val{ctx.fp8e5m2w8a8.flip} \\
INT8 & 8/8 & \val{cnn.int8w8a8.kib} & \val{cnn.int8w8a8.mf1} & $\val{cnn.int8w8a8.d}$ {\scriptsize[$\val{cnn.int8w8a8.dlo}$, $\val{cnn.int8w8a8.dhi}$]} & \val{cnn.int8w8a8.flip} & \val{ctx.int8w8a8.kib} & \val{ctx.int8w8a8.mf1} & $\val{ctx.int8w8a8.d}$ {\scriptsize[$\val{ctx.int8w8a8.dlo}$, $\val{ctx.int8w8a8.dhi}$]} & \val{ctx.int8w8a8.flip} \\
INT8 (W only) & 8/32 & \val{cnn.int8w8.kib} & \val{cnn.int8w8.mf1} & $\val{cnn.int8w8.d}$ {\scriptsize[$\val{cnn.int8w8.dlo}$, $\val{cnn.int8w8.dhi}$]} & \val{cnn.int8w8.flip} & \val{ctx.int8w8.kib} & \val{ctx.int8w8.mf1} & $\val{ctx.int8w8.d}$ {\scriptsize[$\val{ctx.int8w8.dlo}$, $\val{ctx.int8w8.dhi}$]} & \val{ctx.int8w8.flip} \\
\midrule
INT4-g32 (W only) & 4/16 & \val{cnn.int4g32w4.kib} & \val{cnn.int4g32w4.mf1} & $\val{cnn.int4g32w4.d}$ {\scriptsize[$\val{cnn.int4g32w4.dlo}$, $\val{cnn.int4g32w4.dhi}$]}$^{**}$ & \val{cnn.int4g32w4.flip} & \val{ctx.int4g32w4.kib} & \val{ctx.int4g32w4.mf1} & $\val{ctx.int4g32w4.d}$ {\scriptsize[$\val{ctx.int4g32w4.dlo}$, $\val{ctx.int4g32w4.dhi}$]} & \val{ctx.int4g32w4.flip} \\
\midrule
MXFP4-32 & 4/4 & \val{cnn.mxfp432w4a4.kib} & \val{cnn.mxfp432w4a4.mf1} & $\val{cnn.mxfp432w4a4.d}$ {\scriptsize[$\val{cnn.mxfp432w4a4.dlo}$, $\val{cnn.mxfp432w4a4.dhi}$]}$^{**}$ & \val{cnn.mxfp432w4a4.flip} & \val{ctx.mxfp432w4a4.kib} & \val{ctx.mxfp432w4a4.mf1} & $\val{ctx.mxfp432w4a4.d}$ {\scriptsize[$\val{ctx.mxfp432w4a4.dlo}$, $\val{ctx.mxfp432w4a4.dhi}$]} & \val{ctx.mxfp432w4a4.flip} \\
MXFP4-16$^{\dagger}$ & 4/4 & \val{cnn.mxfp416w4a4.kib} & \val{cnn.mxfp416w4a4.mf1} & $\val{cnn.mxfp416w4a4.d}$ {\scriptsize[$\val{cnn.mxfp416w4a4.dlo}$, $\val{cnn.mxfp416w4a4.dhi}$]}$^{**}$ & \val{cnn.mxfp416w4a4.flip} & \val{ctx.mxfp416w4a4.kib} & \val{ctx.mxfp416w4a4.mf1} & $\val{ctx.mxfp416w4a4.d}$ {\scriptsize[$\val{ctx.mxfp416w4a4.dlo}$, $\val{ctx.mxfp416w4a4.dhi}$]} & \val{ctx.mxfp416w4a4.flip} \\
MXFP4-32 (W only) & 4/32 & \val{cnn.mxfp432w4.kib} & \val{cnn.mxfp432w4.mf1} & $\val{cnn.mxfp432w4.d}$ {\scriptsize[$\val{cnn.mxfp432w4.dlo}$, $\val{cnn.mxfp432w4.dhi}$]}$^{**}$ & \val{cnn.mxfp432w4.flip} & \val{ctx.mxfp432w4.kib} & \val{ctx.mxfp432w4.mf1} & $\val{ctx.mxfp432w4.d}$ {\scriptsize[$\val{ctx.mxfp432w4.dlo}$, $\val{ctx.mxfp432w4.dhi}$]} & \val{ctx.mxfp432w4.flip} \\
NVFP4-16 & 4/4 & \val{cnn.nvfp416w4a4.kib} & \val{cnn.nvfp416w4a4.mf1} & $\val{cnn.nvfp416w4a4.d}$ {\scriptsize[$\val{cnn.nvfp416w4a4.dlo}$, $\val{cnn.nvfp416w4a4.dhi}$]}$^{**}$ & \val{cnn.nvfp416w4a4.flip} & \val{ctx.nvfp416w4a4.kib} & \val{ctx.nvfp416w4a4.mf1} & $\val{ctx.nvfp416w4a4.d}$ {\scriptsize[$\val{ctx.nvfp416w4a4.dlo}$, $\val{ctx.nvfp416w4a4.dhi}$]} & \val{ctx.nvfp416w4a4.flip} \\
NVFP4-32$^{\dagger}$ & 4/4 & \val{cnn.nvfp432w4a4.kib} & \val{cnn.nvfp432w4a4.mf1} & $\val{cnn.nvfp432w4a4.d}$ {\scriptsize[$\val{cnn.nvfp432w4a4.dlo}$, $\val{cnn.nvfp432w4a4.dhi}$]}$^{**}$ & \val{cnn.nvfp432w4a4.flip} & \val{ctx.nvfp432w4a4.kib} & \val{ctx.nvfp432w4a4.mf1} & $\val{ctx.nvfp432w4a4.d}$ {\scriptsize[$\val{ctx.nvfp432w4a4.dlo}$, $\val{ctx.nvfp432w4a4.dhi}$]} & \val{ctx.nvfp432w4a4.flip} \\
NVFP4-16 (W only) & 4/32 & \val{cnn.nvfp416w4.kib} & \val{cnn.nvfp416w4.mf1} & $\val{cnn.nvfp416w4.d}$ {\scriptsize[$\val{cnn.nvfp416w4.dlo}$, $\val{cnn.nvfp416w4.dhi}$]}$^{**}$ & \val{cnn.nvfp416w4.flip} & \val{ctx.nvfp416w4.kib} & \val{ctx.nvfp416w4.mf1} & $\val{ctx.nvfp416w4.d}$ {\scriptsize[$\val{ctx.nvfp416w4.dlo}$, $\val{ctx.nvfp416w4.dhi}$]} & \val{ctx.nvfp416w4.flip} \\
\bottomrule
\end{tabular}}
\end{table*}

\begin{figure*}
\centering
\includegraphics[width=\textwidth]{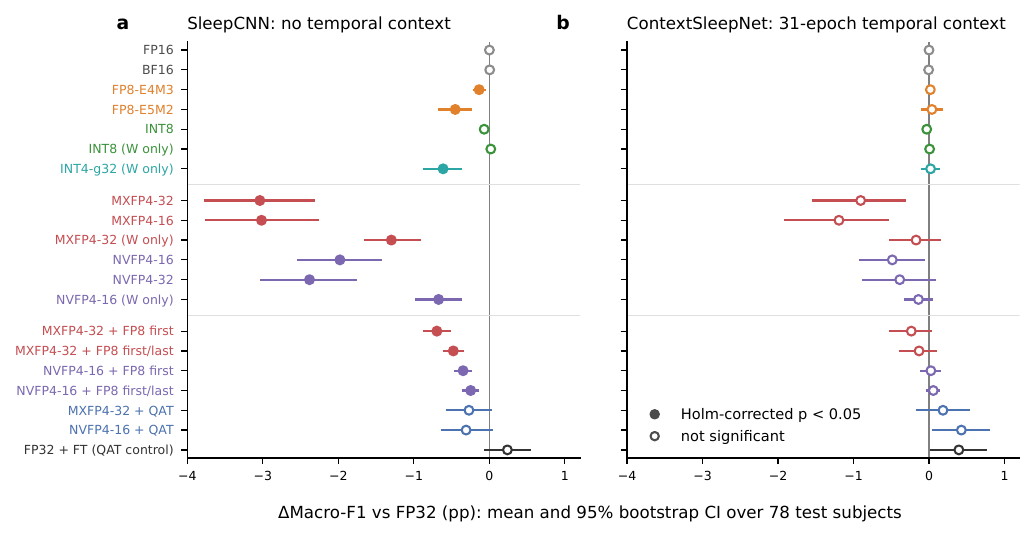}
\caption{Change in Macro-F1 relative to FP32 for every condition on Sleep-EDF-78: mean and 95\% bootstrap CI over the 78 test subjects of the per-subject paired change, averaged over three seeds. Filled markers: Holm-corrected Wilcoxon $p<0.05$ (20 conditions per model). Open markers: not significant. Label colors mark format families. Rows are grouped into 16- and 8-bit formats plus weight-only INT4 (top), block-scaled 4-bit formats (middle), and recovery strategies with the QAT control (bottom). W only: weights quantized, activations in FP32 (BF16 for INT4-g32). (a) SleepCNN loses \val{w4a4.cnn.range}\pp{} under every W4A4 format. (b) ContextSleepNet loses at most \val{w4a4.ctx.max}\pp{}, and no condition is significant after correction.}
\label{fig:forest}
\end{figure*}

\subsection{Format benchmark}
Table~\ref{tab:main} and Fig.~\ref{fig:forest} give the full benchmark.

\textbf{8- and 16-bit formats stay close to FP32.}
FP16 and BF16 change Macro-F1 by at most 0.01\pp{}. Every tested 8-bit format stays within \val{eightbit.maxabs}\pp{} of FP32 for both models. For SleepCNN, FP8-E4M3 and FP8-E5M2 cause small losses that remain significant after Holm correction: $\val{cnn.fp8e4m3w8a8.d}$\pp{} and $\val{cnn.fp8e5m2w8a8.d}$\pp{}, respectively. E5M2 has a wider range but one fewer mantissa bit than E4M3, and it loses more. INT8 W8A8 changes the Macro-F1 of SleepCNN by $\val{cnn.int8w8a8.d}$\pp{}, which is not significant after correction. For ContextSleepNet, no 8- or 16-bit format differs significantly from FP32. Weight-only INT4-g32 changes Macro-F1 by $\val{cnn.int4g32w4.d}$\pp{} for SleepCNN, a significant loss, and by $\val{ctx.int4g32w4.d}$\pp{} for ContextSleepNet.

All 8- and 16-bit formats meet the exploratory non-inferiority criterion (margin \val{ni.margin}\pp{}) for both models. The smallest simultaneous lower bound is $\val{ni.eightbit.minlo}$\pp{}. INT4-g32 also meets the criterion for ContextSleepNet. For SleepCNN, INT4-g32 is borderline, and the decision depends on the interval method. Its percentile-bootstrap lower bound is $\val{nirob.int4.percentile}$\pp{}. Across 50 bootstrap seeds, this bound ranges from $\val{nirob.int4.seedsmin}$ to $\val{nirob.int4.seedsmax}$\pp{}. Its BCa bound is $\val{nirob.int4.bca}$\pp{}. The other \val{nirob.same} of the \val{nirob.n} non-inferiority decisions agree across all three interval methods (percentile bootstrap, BCa bootstrap, and $t$ interval).

\textbf{W4A4 costs SleepCNN much more than ContextSleepNet.}
With 4-bit weights and activations, SleepCNN loses \val{w4a4.cnn.range}\pp{} of Macro-F1. All four Holm-adjusted $p$-values are below 0.001. Under MXFP4-32, \val{cnn.mxfp432w4a4.flip}\% of its epoch predictions differ from FP32.

ContextSleepNet changes by $\val{ctx.mxfp432w4a4.d}$\pp{} under MXFP4-32 and $\val{ctx.mxfp416w4a4.d}$\pp{} under MXFP4-16. Under NVFP4-16 and NVFP4-32, it changes by $\val{ctx.nvfp416w4a4.d}$\pp{} and $\val{ctx.nvfp432w4a4.d}$\pp{}, respectively. The pointwise bootstrap CIs of the mean exclude zero for MXFP4-32, MXFP4-16, and NVFP4-16. The Wilcoxon tests, however, are not significant after Holm correction. Their unadjusted $p$-values are $\val{ctx.mxfp432w4a4.p}$ for MXFP4-32 and $\val{ctx.mxfp416w4a4.p}$ for MXFP4-16. The two analyses differ in their target and in their control of multiplicity. The CIs concern the mean and are pointwise. The tests use signed ranks and are Holm-adjusted. Under MXFP4-32, the per-subject median change is only $\val{abl.ctx.mxfp432w4a4.med}$\pp{}, compared with a mean of $\val{ctx.mxfp432w4a4.d}$\pp{}. A minority of subjects thus carries much of the mean loss (Section~\ref{sec:subject}).

A nonsignificant test does not show that a loss is absent or small. No W4A4 PTQ configuration meets the non-inferiority criterion for either model. For ContextSleepNet, the simultaneous lower bounds range from $\val{ni.ctx.wfour.lomin}$ to $\val{ni.ctx.wfour.lomax}$\pp{}. Failing to show non-inferiority does not show inferiority either. For ContextSleepNet with NVFP4-16, for example, the simultaneous bounds are [$\val{ni.ctx.nvfp416w4a4.lo}$, $\val{ni.ctx.nvfp416w4a4.hi}$]\pp{}. This interval includes both no change and losses larger than the margin.

Relative to FP32, MXFP4-32 shrinks packed weight storage \val{cnn.mxfp432w4a4.compr}-fold for SleepCNN and \val{ctx.mxfp432w4a4.compr}-fold for ContextSleepNet. Under this format, ContextSleepNet loses about one-third as much Macro-F1 as SleepCNN. The ablation in Section~\ref{sec:ablation} shows that temporal context explains only a small part of this difference.

\textbf{NVFP4 outperforms MXFP4 at matched block sizes.}
We compare the two scaling recipes at block sizes of 16 and 32. MXFP4-16 and NVFP4-32 use non-standard block sizes and serve as an ablation. All comparisons use paired per-subject differences. We apply Holm correction across the four recipe comparisons and the four block-size comparisons described below. We write $p_{\mathrm{H}}$ for the adjusted $p$-values.

For SleepCNN, NVFP4 improves Macro-F1 over MXFP4 by $\val{subj.con.cnn.scale16.d}$\pp{} at a block size of 16 ($p_{\mathrm{H}}\val{subj.con.cnn.scale16.ph}$). At a block size of 32, the improvement is $\val{subj.con.cnn.scale32.d}$\pp{} ($p_{\mathrm{H}}=\val{subj.con.cnn.scale32.ph}$). For ContextSleepNet, the corresponding improvements are $\val{subj.con.ctx.scale16.d}$\pp{} ($p_{\mathrm{H}}=\val{subj.con.ctx.scale16.ph}$) and $\val{subj.con.ctx.scale32.d}$\pp{} ($p_{\mathrm{H}}=\val{subj.con.ctx.scale32.ph}$). Three of these four improvements remain significant after correction.

Halving the block size from 32 to 16 has small effects, and their sign varies. For MXFP4, Macro-F1 changes by $\val{subj.con.cnn.blockmx.d}$\pp{} for SleepCNN ($p_{\mathrm{H}}=\val{subj.con.cnn.blockmx.ph}$) and by $\val{subj.con.ctx.blockmx.d}$\pp{} for ContextSleepNet ($p_{\mathrm{H}}=\val{subj.con.ctx.blockmx.ph}$). For NVFP4, it rises by $\val{subj.con.cnn.blocknv.d}$\pp{} for SleepCNN ($p_{\mathrm{H}}=\val{subj.con.cnn.blocknv.ph}$) and changes by $\val{subj.con.ctx.blocknv.d}$\pp{} for ContextSleepNet ($p_{\mathrm{H}}=\val{subj.con.ctx.blocknv.ph}$).

The two scaling recipes differ in two ways. MXFP4 uses an E8M0 block scale. NVFP4 uses an E4M3 block scale and an additional FP32 tensor scale. Our comparison cannot separate these two differences. We hypothesize that the finer resolution of the E4M3 block scale contributes to the advantage of NVFP4. Unless the block scale is clipped, it places the scaled block maximum near 6, the largest E2M1 value. Under the MXFP4 rule used here, the scaled maximum of a nonzero block lies in $[4,8)$ before rounding. Near 4, the top E2M1 level can stay unused. Above 6, the maximum is clipped to 6. This saturation under the round-down rule is known from MXFP8 training~\cite{mishra2025recipes}, and overflow-aware scaling has been proposed for MXFP4~\cite{chhugani2026mxfp4}. This difference may explain our results, but our experiments do not establish it as the cause.

\begin{figure}[!tb]
\centering
\includegraphics[width=\figcol]{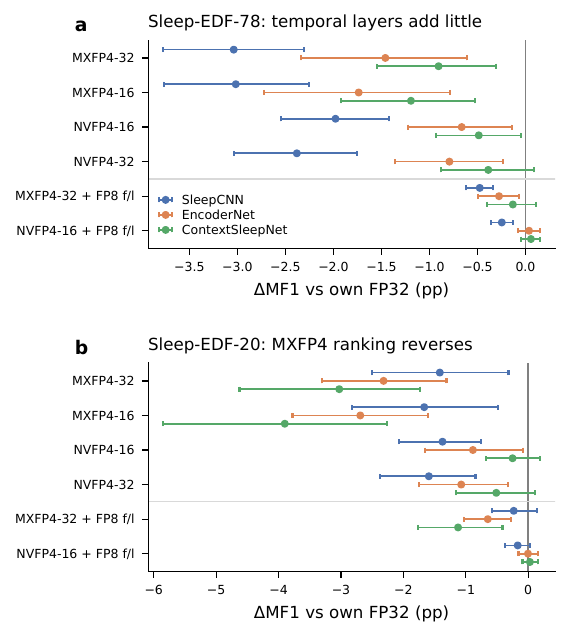}
\caption{Same-encoder ablation. $\Delta$MF1 of each model relative to its own FP32 model (mean and 95\% bootstrap CI over subjects; per-subject paired, three seeds). EncoderNet is ContextSleepNet without its temporal layers. FP8 f/l: FP8 first and last layers. The legend in (a) applies to both panels. (a) Sleep-EDF-78 ($n=78$): EncoderNet loses much less than SleepCNN and only slightly more than ContextSleepNet. (b) Sleep-EDF-20 ($n=20$): under MXFP4, the two models with the larger encoder lose more than SleepCNN. With NVFP4-16 and FP8 first/last layers, all three models stay close to FP32.}
\label{fig:ablation}
\end{figure}

\subsection{Temporal context or other model differences?}
\label{sec:ablation}
Figure~\ref{fig:ablation}a and Table~\ref{tab:ablation} (Appendix~\ref{app:extra}) report the same-encoder ablation. EncoderNet shares ContextSleepNet's encoder and training setup but has no temporal layers. It therefore classifies each epoch on its own. Its FP32 Macro-F1 is \val{abl.enc.fp32.mf1}\%, slightly below the \val{abl.cnn.fp32.mf1}\% of SleepCNN. Yet it loses less than SleepCNN under W4A4. Relative to its own FP32 baseline, EncoderNet changes by $\val{abl.enc.mxfp432w4a4.d}$\pp{} under MXFP4-32 and $\val{abl.enc.mxfp416w4a4.d}$\pp{} under MXFP4-16. Under NVFP4-16 and NVFP4-32, it changes by $\val{abl.enc.nvfp416w4a4.d}$\pp{} and $\val{abl.enc.nvfp432w4a4.d}$\pp{}, respectively.

Across the four W4A4 formats, the non-temporal effect (EncoderNet minus SleepCNN) ranges from $\val{abl.enc.rmin}$ to $\val{abl.enc.rmax}$\pp{}. All four contrasts are significant after Holm correction ($p_{\mathrm{H}}<0.001$). The temporal-layer effect (ContextSleepNet minus EncoderNet) ranges from $\val{abl.tcn.rmin}$ to $\val{abl.tcn.rmax}$\pp{}. None of these contrasts is significant after correction ($p_{\mathrm{H}}\ge\val{abl.tcn.phmin}$). The pointwise bootstrap CIs of the mean, however, exclude zero for MXFP4-32 and NVFP4-32. As above, the CIs and the corrected rank tests differ in their target and in their control of multiplicity.

The point estimates thus suggest that most of ContextSleepNet's lower sensitivity to W4A4 on this cohort comes from non-temporal differences. These include the encoder architecture and the training regime, which this ablation cannot separate. Adding the temporal layers has a comparatively small effect. EncoderNet's per-subject losses also correlate with those of SleepCNN and ContextSleepNet (Spearman $\rho=\val{abl.rho.min}$--$\val{abl.rho.max}$).

\textbf{Sleep-EDF-20 subset.} On the 20-subject subset, all 8- and 16-bit formats stay within \val{sc20.eightbit.maxabs}\pp{} of FP32 for all three models (Appendix~\ref{app:sc20}, Table~\ref{tab:sc20}, and Fig.~\ref{fig:ablation}b). NVFP4-16 with FP8 first/last layers also stays close to FP32. It changes Macro-F1 by $\val{sc20.cnn.nvfp416w4a4fp8firstlast.d}$\pp{} for SleepCNN, $\val{sc20.enc.nvfp416w4a4fp8firstlast.d}$\pp{} for EncoderNet, and $\val{sc20.ctx.nvfp416w4a4fp8firstlast.d}$\pp{} for ContextSleepNet.

Under MXFP4, however, the ranking of the models reverses. Under MXFP4-32, ContextSleepNet changes by $\val{sc20.ctx.mxfp432w4a4.d}$\pp{} and EncoderNet by $\val{sc20.enc.mxfp432w4a4.d}$\pp{}. Both lose more than SleepCNN, which changes by $\val{sc20.cnn.mxfp432w4a4.d}$\pp{}. FP8 first/last layers do not fully restore these two models under MXFP4-32. A change of $\val{sc20.enc.mxfp432w4a4fp8firstlast.d}$\pp{} remains for EncoderNet ($p_{\mathrm{H}}=\val{sc20.enc.mxfp432w4a4fp8firstlast.ph}$) and $\val{sc20.ctx.mxfp432w4a4fp8firstlast.d}$\pp{} for ContextSleepNet ($p_{\mathrm{H}}=\val{sc20.ctx.mxfp432w4a4fp8firstlast.ph}$). SleepCNN's change of $\val{sc20.cnn.mxfp432w4a4fp8firstlast.d}$\pp{} is not significant. For EncoderNet and ContextSleepNet, NVFP4-16 loses much less than MXFP4-32. Their changes under NVFP4-16 are $\val{sc20.enc.nvfp416w4a4.d}$\pp{} and $\val{sc20.ctx.nvfp416w4a4.d}$\pp{}, respectively. This comparison changes both the scaling recipe and the block size, so it does not isolate either factor.

The temporal-layer effect also behaves differently on this subset. Under NVFP4-16, ContextSleepNet loses less than EncoderNet. The effect is $\val{sc20.tcn.nvfp416w4a4.d}$\pp{}, with a 95\% CI of [$\val{sc20.tcn.nvfp416w4a4.lo}$, $\val{sc20.tcn.nvfp416w4a4.hi}$]\pp{}, and it is significant after Holm correction ($p_{\mathrm{H}}=\val{sc20.tcn.nvfp416w4a4.ph}$).
Under NVFP4-32, it is $\val{sc20.tcn.nvfp432w4a4.d}$\pp{} ($p_{\mathrm{H}}=\val{sc20.tcn.nvfp432w4a4.ph}$). Under MXFP4, the sign reverses. The effect is $\val{sc20.tcn.mxfp432w4a4.d}$\pp{} for MXFP4-32 and $\val{sc20.tcn.mxfp416w4a4.d}$\pp{} for MXFP4-16. Neither is significant ($p_{\mathrm{H}}=\val{sc20.tcn.mxfp432w4a4.ph}$ and $\val{sc20.tcn.mxfp416w4a4.ph}$). On this subset, the temporal layers thus help under one scaling recipe and do not help under the other.

The subset has only 20 subjects, two per test fold, so its estimates are less precise. It is also nested within Sleep-EDF-78, so it is not an independent replication. Its models are trained on far less data as well. Each fold leaves 18 subjects for fitting and validation, compared with 70 or 71 in Sleep-EDF-78, and all subjects are 25--34 years old. Differences between the two cohorts can therefore reflect the amount of training data, the population, or both.

Three findings hold in both cohorts. First, 8- and 16-bit formats cause only small changes. Second, NVFP4-16 with FP8 first/last layers stays near FP32. Third, in the point estimates, EncoderNet and ContextSleepNet lose less under NVFP4-16 than under MXFP4-32. We did not test this last contrast directly. Two findings do not carry over. The ranking of the architectures by their sensitivity to MXFP4 reverses, and the temporal-layer effect changes sign with the scaling recipe.

\begin{figure*}
\centering
\includegraphics[width=\textwidth]{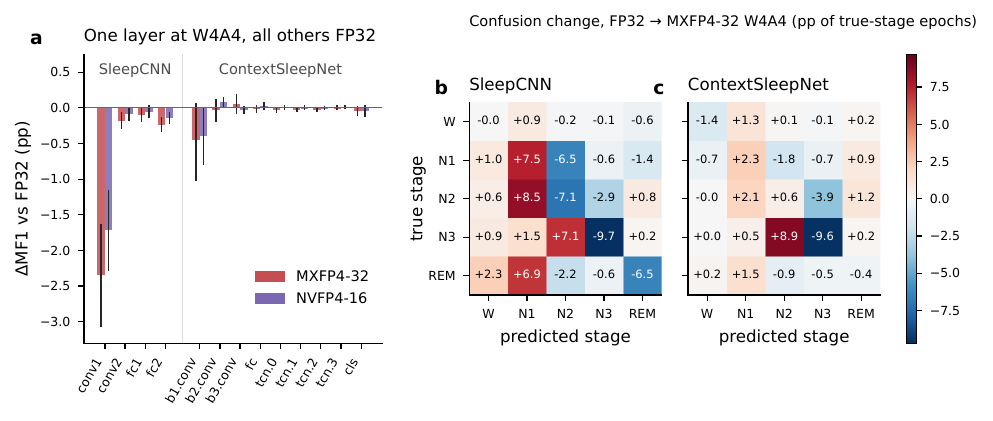}
\caption{Where the 4-bit error arises. (a) $\Delta$MF1 when a single layer is quantized to W4A4 and all others stay in FP32 (mean and 95\% CI). For SleepCNN, most of the loss comes from \texttt{conv1}, the layer that quantizes the raw EEG. (b, c) Change of the row-normalized confusion matrix from FP32 to MXFP4-32 W4A4 (pooled test epochs, mean over seeds). Entries are percentage points of each true stage, so each row sums to zero up to rounding. Both models move true N3 epochs into N2. SleepCNN also moves N2 and REM epochs into N1.}
\label{fig:mechanism}
\end{figure*}

\subsection{Where the error arises}
\label{sec:mechanism}
Quantizing one layer at a time shows that the first convolution is the most sensitive layer (Fig.~\ref{fig:mechanism}a and Table~\ref{tab:sens}). For SleepCNN, quantizing only \texttt{conv1} changes Macro-F1 by $\val{cnn.ls.mxfp432w4a4.conv1.d}$\pp{} under MXFP4-32 and $\val{cnn.ls.nvfp416w4a4.conv1.d}$\pp{} under NVFP4-16. These losses make up most of the corresponding full W4A4 losses. No other single layer costs more than \val{ls.cnn.othermax}\pp{}. In ContextSleepNet, the first convolution also causes the largest loss: $\val{ctx.ls.mxfp432w4a4.b1conv.d}$\pp{} under MXFP4-32 and $\val{ctx.ls.nvfp416w4a4.b1conv.d}$\pp{} under NVFP4-16. Neither change is significant. The temporal layers show smaller effects.

Table~\ref{tab:lone} separates the weights of the first layer from its input. For SleepCNN under MXFP4-32, quantizing only the weights changes Macro-F1 by $\val{lone.cnn.mxfp432w4a4.w.d}$\pp{}. Quantizing only the input changes it by $\val{lone.cnn.mxfp432w4a4.a.d}$\pp{}. Both changes are significant after Holm correction ($p_{\mathrm{H}}<0.01$). Under NVFP4-16, the corresponding changes are $\val{lone.cnn.nvfp416w4a4.w.d}$\pp{} and $\val{lone.cnn.nvfp416w4a4.a.d}$\pp{}, which are also significant. Quantizing both under MXFP4-32 changes Macro-F1 by $\val{lone.cnn.mxfp432w4a4.wa.d}$\pp{}. This loss exceeds the sum of the two separate losses, and the same holds under NVFP4-16. For EncoderNet and ContextSleepNet, quantizing either the weights or the input of the first layer alone changes Macro-F1 by at most 0.24\pp{}. None of these changes is significant.

The first layer is the only layer that receives the normalized EEG directly. Each of its activation blocks groups up to 16 or 32 consecutive EEG samples of an input patch under one shared scale. This scale follows the largest sample in the block. A high-amplitude slow wave or artifact can therefore coarsen the quantization grid for the other samples in the block. This could blur low-amplitude waveform details that matter for staging. Our experiments, however, do not show which physiological features change, or whether this mechanism explains the classification errors.

The confusion matrices show how the errors change (Fig.~\ref{fig:mechanism}b,c). Under MXFP4-32, the share of true N3 epochs classified as N2 rises by \val{cm.cnn.mx.nIIInII}\pp{} for SleepCNN and \val{cm.ctx.mx.nIIInII}\pp{} for ContextSleepNet. N3 recall falls by \val{cm.cnn.mx.nIIIrecall}\pp{} and \val{cm.ctx.mx.nIIIrecall}\pp{}, respectively. This is the largest recall drop among the five stages for both models. N3 also loses the most recall under NVFP4-16.

SleepCNN also shifts predictions toward N1. The share of true N2 epochs classified as N1 rises by \val{cm.cnn.mx.nIInI}\pp{}. The share of true REM epochs classified as N1 rises by \val{cm.cnn.mx.remnI}\pp{}. These shifts are smaller for ContextSleepNet.

When the confusions are normalized by the size of each true stage, the largest increase is N3$\to$N2 for ContextSleepNet. For SleepCNN, it is N2$\to$N1: \val{cm.cnn.mx.nIInI}\pp{}, compared with \val{cm.cnn.mx.nIIInII}\pp{} for N3$\to$N2. In absolute counts, however, N2$\to$N1 is the largest increase for both models. It adds \val{cm.cnn.mx.abs.nIInI} epochs per seed for SleepCNN and \val{cm.ctx.mx.abs.nIInI} for ContextSleepNet. N3$\to$N2 adds \val{cm.cnn.mx.abs.nIIInII} and \val{cm.ctx.mx.abs.nIIInII} epochs, respectively. Absolute counts reflect both the confusion rate and the stage frequency, and N2 is much more common than N3.

Human scorers also find it hard to separate N2 from N3~\cite{rosenberg2013aasm}. Table~\ref{tab:perclass} reports the per-stage F1 scores. The FP32 models over-predict N3, and quantization reduces the number of N3 predictions in these conditions. N3 precision therefore rises enough to increase N3 F1, despite the lower N3 recall. Measured by F1, the stages that lose the most for SleepCNN under MXFP4-32 are REM (from \val{cnn.fp32.f1rem}\% to \val{cnn.mxfp432w4a4.f1rem}\%) and N2 (from \val{cnn.fp32.f1n2}\% to \val{cnn.mxfp432w4a4.f1n2}\%).

\begin{figure}[!tb]
\centering
\includegraphics[width=\figcol]{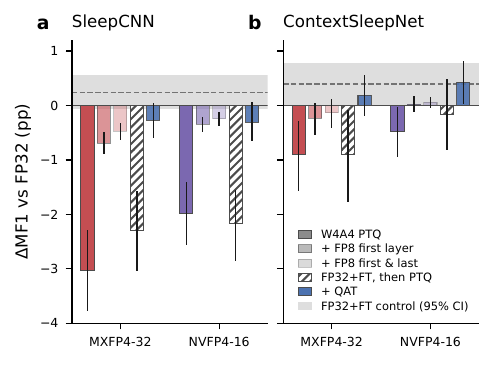}
\caption{Recovering W4A4 accuracy on Sleep-EDF-78 ($\Delta$MF1 vs FP32; mean and 95\% CI). Bar color gives the format (red: MXFP4-32; purple: NVFP4-16). Lighter shades: FP8 first layer, and FP8 first and last layers; both need no retraining. Hatched: FT$\to$PTQ, the FP32+FT model quantized with the same PTQ. Blue: QAT. Gray band and dashed line: 95\% CI and mean of the matched control (FP32+FT, fine-tuned for the same five training epochs as QAT). QAT should be judged against this control and against FT$\to$PTQ, not against FP32.}
\label{fig:recovery}
\end{figure}

\subsection{Recovering 4-bit accuracy}
\label{sec:recovery}
Keeping the first layer in FP8 removes most of SleepCNN's loss. Under MXFP4-32, the change relative to FP32 improves from $\val{cnn.mxfp432w4a4.d}$\pp{} to $\val{cnn.mxfp432w4a4fp8first.d}$\pp{}. Under NVFP4-16, it improves from $\val{cnn.nvfp416w4a4.d}$\pp{} to $\val{cnn.nvfp416w4a4fp8first.d}$\pp{}. Keeping both the first and the last layer in FP8 brings both tested 4-bit formats within \val{mix.maxabs}\pp{} of FP32 for both models (Table~\ref{tab:recovery} and Fig.~\ref{fig:recovery}).

This adds \val{mix.ovhmin}--\val{mix.ovhmax}\% weight storage relative to the corresponding W4A4 configuration. The number of bit operations (BOPs) rises more, because the first layer performs \val{cnn.firstmacpct}\% of all MACs in SleepCNN and \val{ctx.firstmacpct}\% in ContextSleepNet. With FP8 first/last layers, the BOP count is \val{cnn.mxfp432w4a4.bopsx} times that of pure MXFP4-32 for SleepCNN and \val{ctx.mxfp432w4a4.bopsx} times for ContextSleepNet. It is still \val{cnn.mxfp432w4a4.bopsvsfp} and \val{ctx.mxfp432w4a4.bopsvsfp} times lower than FP32, respectively. BOPs is a theoretical proxy for arithmetic cost. It does not establish equal savings in latency or energy.

\begin{table*}
\centering
\caption{Recovering W4A4 accuracy. FP8 first/last: first (raw-EEG input) and last layer in FP8-E4M3 W8A8, all other layers 4-bit. QAT: quantization-aware fine-tuning for five training epochs with the straight-through estimator. FP32+FT: the same fine-tuning without quantization (matched control). FT$\to$PTQ: the FP32+FT model followed by the same PTQ. $\Delta$MF1 vs FP32 as in Table~\ref{tab:main}. ``vs FP32+FT'': QAT (or FT$\to$PTQ) minus the control, per-subject paired. Stars mark Holm-corrected Wilcoxon $p<0.05/0.01$: in the $\Delta$MF1 column within the 20-condition family of each model, and in the ``vs FP32+FT'' column within the two QAT comparisons of each model. FT$\to$PTQ rows lie outside both families and carry no stars. GBOPs: $\sum_\ell$MACs$_\ell\,b_w b_a$ per 30-s epoch, in $10^9$.}
\label{tab:recovery}
\setlength{\tabcolsep}{3pt}
\footnotesize
\fitwide{%
\begin{tabular}{l rrll rrll}
\toprule
 & \multicolumn{4}{c}{SleepCNN} & \multicolumn{4}{c}{ContextSleepNet} \\
\cmidrule(lr){2-5}\cmidrule(lr){6-9}
Condition & KiB & GBOPs & $\Delta$MF1 vs FP32 & vs FP32+FT & KiB & GBOPs & $\Delta$MF1 vs FP32 & vs FP32+FT \\
\midrule
MXFP4-32 PTQ & \val{cnn.mxfp432w4a4.kib} & \val{cnn.mxfp432w4a4.gbops} & $\val{cnn.mxfp432w4a4.d}$ {\scriptsize[$\val{cnn.mxfp432w4a4.dlo}$, $\val{cnn.mxfp432w4a4.dhi}$]}$^{**}$ &  & \val{ctx.mxfp432w4a4.kib} & \val{ctx.mxfp432w4a4.gbops} & $\val{ctx.mxfp432w4a4.d}$ {\scriptsize[$\val{ctx.mxfp432w4a4.dlo}$, $\val{ctx.mxfp432w4a4.dhi}$]} &  \\
MXFP4-32 + FP8 first & \val{cnn.mxfp432w4a4fp8first.kib} & \val{cnn.mxfp432w4a4fp8first.gbops} & $\val{cnn.mxfp432w4a4fp8first.d}$ {\scriptsize[$\val{cnn.mxfp432w4a4fp8first.dlo}$, $\val{cnn.mxfp432w4a4fp8first.dhi}$]}$^{**}$ &  & \val{ctx.mxfp432w4a4fp8first.kib} & \val{ctx.mxfp432w4a4fp8first.gbops} & $\val{ctx.mxfp432w4a4fp8first.d}$ {\scriptsize[$\val{ctx.mxfp432w4a4fp8first.dlo}$, $\val{ctx.mxfp432w4a4fp8first.dhi}$]} &  \\
MXFP4-32 + FP8 first/last & \val{cnn.mxfp432w4a4fp8firstlast.kib} & \val{cnn.mxfp432w4a4fp8firstlast.gbops} & $\val{cnn.mxfp432w4a4fp8firstlast.d}$ {\scriptsize[$\val{cnn.mxfp432w4a4fp8firstlast.dlo}$, $\val{cnn.mxfp432w4a4fp8firstlast.dhi}$]}$^{**}$ &  & \val{ctx.mxfp432w4a4fp8firstlast.kib} & \val{ctx.mxfp432w4a4fp8firstlast.gbops} & $\val{ctx.mxfp432w4a4fp8firstlast.d}$ {\scriptsize[$\val{ctx.mxfp432w4a4fp8firstlast.dlo}$, $\val{ctx.mxfp432w4a4fp8firstlast.dhi}$]} &  \\
MXFP4-32 FT$\to$PTQ & \val{cnn.mxfp432w4a4.kib} & \val{cnn.mxfp432w4a4.gbops} & $\val{ftptq.cnn.mxfp432w4a4.ftptq.d}$ {\scriptsize[$\val{ftptq.cnn.mxfp432w4a4.ftptq.lo}$, $\val{ftptq.cnn.mxfp432w4a4.ftptq.hi}$]} & $\val{ftptq.cnn.mxfp432w4a4.vsft.d}$ {\scriptsize[$\val{ftptq.cnn.mxfp432w4a4.vsft.lo}$, $\val{ftptq.cnn.mxfp432w4a4.vsft.hi}$]} & \val{ctx.mxfp432w4a4.kib} & \val{ctx.mxfp432w4a4.gbops} & $\val{ftptq.ctx.mxfp432w4a4.ftptq.d}$ {\scriptsize[$\val{ftptq.ctx.mxfp432w4a4.ftptq.lo}$, $\val{ftptq.ctx.mxfp432w4a4.ftptq.hi}$]} & $\val{ftptq.ctx.mxfp432w4a4.vsft.d}$ {\scriptsize[$\val{ftptq.ctx.mxfp432w4a4.vsft.lo}$, $\val{ftptq.ctx.mxfp432w4a4.vsft.hi}$]} \\
MXFP4-32 + QAT & \val{cnn.mxfp432w4a4qat.kib} & \val{cnn.mxfp432w4a4qat.gbops} & $\val{cnn.mxfp432w4a4qat.d}$ {\scriptsize[$\val{cnn.mxfp432w4a4qat.dlo}$, $\val{cnn.mxfp432w4a4qat.dhi}$]} & $\val{cnn.mxfp432w4a4qat.ctl}$ {\scriptsize[$\val{cnn.mxfp432w4a4qat.ctllo}$, $\val{cnn.mxfp432w4a4qat.ctlhi}$]}$^{**}$ & \val{ctx.mxfp432w4a4qat.kib} & \val{ctx.mxfp432w4a4qat.gbops} & $\val{ctx.mxfp432w4a4qat.d}$ {\scriptsize[$\val{ctx.mxfp432w4a4qat.dlo}$, $\val{ctx.mxfp432w4a4qat.dhi}$]} & $\val{ctx.mxfp432w4a4qat.ctl}$ {\scriptsize[$\val{ctx.mxfp432w4a4qat.ctllo}$, $\val{ctx.mxfp432w4a4qat.ctlhi}$]}$^{*}$ \\
\midrule
NVFP4-16 PTQ & \val{cnn.nvfp416w4a4.kib} & \val{cnn.nvfp416w4a4.gbops} & $\val{cnn.nvfp416w4a4.d}$ {\scriptsize[$\val{cnn.nvfp416w4a4.dlo}$, $\val{cnn.nvfp416w4a4.dhi}$]}$^{**}$ &  & \val{ctx.nvfp416w4a4.kib} & \val{ctx.nvfp416w4a4.gbops} & $\val{ctx.nvfp416w4a4.d}$ {\scriptsize[$\val{ctx.nvfp416w4a4.dlo}$, $\val{ctx.nvfp416w4a4.dhi}$]} &  \\
NVFP4-16 + FP8 first & \val{cnn.nvfp416w4a4fp8first.kib} & \val{cnn.nvfp416w4a4fp8first.gbops} & $\val{cnn.nvfp416w4a4fp8first.d}$ {\scriptsize[$\val{cnn.nvfp416w4a4fp8first.dlo}$, $\val{cnn.nvfp416w4a4fp8first.dhi}$]}$^{**}$ &  & \val{ctx.nvfp416w4a4fp8first.kib} & \val{ctx.nvfp416w4a4fp8first.gbops} & $\val{ctx.nvfp416w4a4fp8first.d}$ {\scriptsize[$\val{ctx.nvfp416w4a4fp8first.dlo}$, $\val{ctx.nvfp416w4a4fp8first.dhi}$]} &  \\
NVFP4-16 + FP8 first/last & \val{cnn.nvfp416w4a4fp8firstlast.kib} & \val{cnn.nvfp416w4a4fp8firstlast.gbops} & $\val{cnn.nvfp416w4a4fp8firstlast.d}$ {\scriptsize[$\val{cnn.nvfp416w4a4fp8firstlast.dlo}$, $\val{cnn.nvfp416w4a4fp8firstlast.dhi}$]}$^{**}$ &  & \val{ctx.nvfp416w4a4fp8firstlast.kib} & \val{ctx.nvfp416w4a4fp8firstlast.gbops} & $\val{ctx.nvfp416w4a4fp8firstlast.d}$ {\scriptsize[$\val{ctx.nvfp416w4a4fp8firstlast.dlo}$, $\val{ctx.nvfp416w4a4fp8firstlast.dhi}$]} &  \\
NVFP4-16 FT$\to$PTQ & \val{cnn.nvfp416w4a4.kib} & \val{cnn.nvfp416w4a4.gbops} & $\val{ftptq.cnn.nvfp416w4a4.ftptq.d}$ {\scriptsize[$\val{ftptq.cnn.nvfp416w4a4.ftptq.lo}$, $\val{ftptq.cnn.nvfp416w4a4.ftptq.hi}$]} & $\val{ftptq.cnn.nvfp416w4a4.vsft.d}$ {\scriptsize[$\val{ftptq.cnn.nvfp416w4a4.vsft.lo}$, $\val{ftptq.cnn.nvfp416w4a4.vsft.hi}$]} & \val{ctx.nvfp416w4a4.kib} & \val{ctx.nvfp416w4a4.gbops} & $\val{ftptq.ctx.nvfp416w4a4.ftptq.d}$ {\scriptsize[$\val{ftptq.ctx.nvfp416w4a4.ftptq.lo}$, $\val{ftptq.ctx.nvfp416w4a4.ftptq.hi}$]} & $\val{ftptq.ctx.nvfp416w4a4.vsft.d}$ {\scriptsize[$\val{ftptq.ctx.nvfp416w4a4.vsft.lo}$, $\val{ftptq.ctx.nvfp416w4a4.vsft.hi}$]} \\
NVFP4-16 + QAT & \val{cnn.nvfp416w4a4qat.kib} & \val{cnn.nvfp416w4a4qat.gbops} & $\val{cnn.nvfp416w4a4qat.d}$ {\scriptsize[$\val{cnn.nvfp416w4a4qat.dlo}$, $\val{cnn.nvfp416w4a4qat.dhi}$]} & $\val{cnn.nvfp416w4a4qat.ctl}$ {\scriptsize[$\val{cnn.nvfp416w4a4qat.ctllo}$, $\val{cnn.nvfp416w4a4qat.ctlhi}$]}$^{**}$ & \val{ctx.nvfp416w4a4qat.kib} & \val{ctx.nvfp416w4a4qat.gbops} & $\val{ctx.nvfp416w4a4qat.d}$ {\scriptsize[$\val{ctx.nvfp416w4a4qat.dlo}$, $\val{ctx.nvfp416w4a4qat.dhi}$]} & $\val{ctx.nvfp416w4a4qat.ctl}$ {\scriptsize[$\val{ctx.nvfp416w4a4qat.ctllo}$, $\val{ctx.nvfp416w4a4qat.ctlhi}$]} \\
\midrule
FP32+FT (control) & \val{cnn.fp32.kib} & \val{cnn.fp32.gbops} & $\val{cnn.fp32ft.d}$ {\scriptsize[$\val{cnn.fp32ft.dlo}$, $\val{cnn.fp32ft.dhi}$]} &  & \val{ctx.fp32.kib} & \val{ctx.fp32.gbops} & $\val{ctx.fp32ft.d}$ {\scriptsize[$\val{ctx.fp32ft.dlo}$, $\val{ctx.fp32ft.dhi}$]} &  \\
\bottomrule
\end{tabular}}
\end{table*}

QAT also restores Macro-F1 to near the FP32 level for MXFP4-32 and NVFP4-16. For SleepCNN, the mean per-subject changes relative to FP32 are $\val{cnn.mxfp432w4a4qat.d}$\pp{} and $\val{cnn.nvfp416w4a4qat.d}$\pp{}, respectively. For ContextSleepNet, they are $\val{ctx.mxfp432w4a4qat.d}$\pp{} and $\val{ctx.nvfp416w4a4qat.d}$\pp{}. None of these changes is significant after Holm correction. All four QAT conditions meet the non-inferiority criterion, with simultaneous lower bounds of at least $\val{ni.qat.lomin}$\pp{}. The FP8 first/last conditions also meet it, with lower bounds of at least $\val{ni.fl.lomin}$\pp{}.

The matched FP32+FT control raises the mean per-subject Macro-F1 by $\val{cnn.fp32ft.d}$\pp{} for SleepCNN and $\val{ctx.fp32ft.d}$\pp{} for ContextSleepNet. Neither gain is significant after Holm correction. For ContextSleepNet, however, the pointwise 95\% CI of [$\val{ctx.fp32ft.dlo}$, $\val{ctx.fp32ft.dhi}$]\pp{} excludes zero, so extra fine-tuning may still help this model slightly. Relative to this control, SleepCNN's QAT models differ by $\val{cnn.mxfp432w4a4qat.ctl}$\pp{} under MXFP4-32 and $\val{cnn.nvfp416w4a4qat.ctl}$\pp{} under NVFP4-16. Both are significant losses after Holm correction. For ContextSleepNet, the corresponding differences are $\val{ctx.mxfp432w4a4qat.ctl}$\pp{} and $\val{ctx.nvfp416w4a4qat.ctl}$\pp{}. Only the MXFP4-32 difference is significant.

ContextSleepNet with NVFP4-16 QAT shows why this control matters. Relative to the original FP32 model, it changes by $\val{ctx.nvfp416w4a4qat.d}$\pp{}, with a pointwise 95\% CI of [$\val{ctx.nvfp416w4a4qat.dlo}$, $\val{ctx.nvfp416w4a4qat.dhi}$]\pp{}. This CI excludes zero, but the Holm-adjusted Wilcoxon test is not significant. The difference from FP32+FT is not significant either. These results show no advantage of QAT over extra FP32 fine-tuning.

We also compare QAT with PTQ applied after the same FP32 fine-tuning (FT$\to$PTQ).
We Holm-correct the four resulting comparisons (two models and two formats) together. For SleepCNN, FT$\to$PTQ changes Macro-F1 relative to the original FP32 model by $\val{ftptq.cnn.mxfp432w4a4.ftptq.d}$\pp{} under MXFP4-32 and $\val{ftptq.cnn.nvfp416w4a4.ftptq.d}$\pp{} under NVFP4-16. QAT beats these results by $\val{ftptq.cnn.mxfp432w4a4.qatminus.d}$\pp{} and $\val{ftptq.cnn.nvfp416w4a4.qatminus.d}$\pp{}, respectively. Both improvements are significant after Holm correction ($p_{\mathrm{H}}<0.001$).

For ContextSleepNet, QAT beats FT$\to$PTQ by $\val{ftptq.ctx.mxfp432w4a4.qatminus.d}$\pp{} under MXFP4-32 and $\val{ftptq.ctx.nvfp416w4a4.qatminus.d}$\pp{} under NVFP4-16. Neither difference is significant after Holm correction ($p_{\mathrm{H}}=\val{ftptq.ctx.mxfp432w4a4.qatminus.ph}$ and $\val{ftptq.ctx.nvfp416w4a4.qatminus.ph}$, respectively). Fine-tuning before PTQ thus does not reliably restore W4A4 accuracy (Table~\ref{tab:recovery}). These controls support a real benefit of quantization-aware fine-tuning for SleepCNN. For ContextSleepNet, the point estimates also favor QAT, but the corrected tests do not confirm this advantage.

Finally, we compare FP8 first/last layers with QAT directly, using paired per-subject differences averaged over seeds. Across the four model--format pairs, the mean difference (FP8 first/last minus QAT) ranges from $\val{flqat.min}$ to $\val{flqat.max}$\pp{}. None is significant after Holm correction ($p_{\mathrm{H}}\ge\val{flqat.phmin}$). All Bonferroni-adjusted one-sided bounds lie within [$\val{flqat.lomin}$, $\val{flqat.himax}$]\pp{}, inside the exploratory equivalence margin of $\pm\val{ni.margin}$\pp{}. All four comparisons therefore meet the equivalence criterion for the mean per-subject Macro-F1. QAT has the higher point estimate in \val{flqat.nqat} of the four comparisons. For these models and formats on Sleep-EDF-78, FP8 first/last layers thus match QAT within the margin, without any extra training.

\begin{figure*}
\centering
\includegraphics[width=\textwidth]{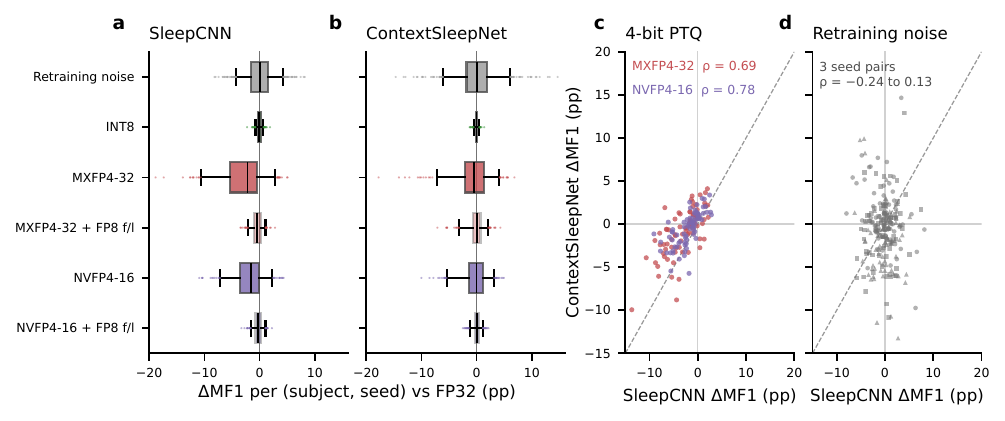}
\caption{Subject-level view. (a, b) Distribution of $\Delta$MF1 per (subject, seed) pair ($n=234$) for selected conditions and for retraining noise (FP32 versus FP32 trained with another seed; both orders of each seed pair, $n=468$). Boxes: interquartile range and median; whiskers: 5th--95th percentiles; dots: values beyond the whiskers. FP8 f/l: FP8 first and last layers. (c) Per-subject $\Delta$MF1 under 4-bit PTQ (seed-averaged) of SleepCNN against ContextSleepNet. $\rho$: Spearman correlation; dashed line: identity. The same subjects lose accuracy in both architectures. (d) The same plot for retraining noise (three seed pairs, one marker shape each): the signed changes of the two architectures do not agree. Their sign is arbitrary, so this panel does not rule out that the same subjects are unstable in both models (Section~\ref{sec:subject}).}
\label{fig:subject}
\end{figure*}

\subsection{Subject-level effects}
\label{sec:subject}
\begin{table}
\centering
\caption{Subject-level view: distribution of per-(subject, seed) $\Delta$MF1 (pp; $n=234$) and epoch flip rate (defined as in Table~\ref{tab:main}). \emph{Retraining noise}: FP32 versus FP32 trained with another seed on the same folds, both orders of each of the three seed pairs ($n=468$, symmetric around zero); it is the reference for both the tail and the flip rate. P5: 5th percentile; $<\!-5$: share of subject--seed pairs losing more than 5\,pp.}
\label{tab:tail}
\setlength{\tabcolsep}{2.4pt}
\footnotesize
\fitcol{%
\begin{tabular}{l rrrr rrrr}
\toprule
 & \multicolumn{4}{c}{SleepCNN} & \multicolumn{4}{c}{ContextSleepNet} \\
\cmidrule(lr){2-5}\cmidrule(lr){6-9}
Condition & Med. & P5 & $<\!-5$ & Flip & Med. & P5 & $<\!-5$ & Flip \\
\midrule
\emph{Retraining noise} & $\val{cnn.noise.med}$ & $\val{cnn.noise.pfive}$ & \val{cnn.noise.ltfive} & \val{cnn.noise.tflip} & $\val{ctx.noise.med}$ & $\val{ctx.noise.pfive}$ & \val{ctx.noise.ltfive} & \val{ctx.noise.tflip} \\
\midrule
INT8 & $\val{cnn.int8w8a8.med}$ & $\val{cnn.int8w8a8.pfive}$ & \val{cnn.int8w8a8.ltfive} & \val{cnn.int8w8a8.tflip} & $\val{ctx.int8w8a8.med}$ & $\val{ctx.int8w8a8.pfive}$ & \val{ctx.int8w8a8.ltfive} & \val{ctx.int8w8a8.tflip} \\
FP8-E4M3 & $\val{cnn.fp8e4m3w8a8.med}$ & $\val{cnn.fp8e4m3w8a8.pfive}$ & \val{cnn.fp8e4m3w8a8.ltfive} & \val{cnn.fp8e4m3w8a8.tflip} & $\val{ctx.fp8e4m3w8a8.med}$ & $\val{ctx.fp8e4m3w8a8.pfive}$ & \val{ctx.fp8e4m3w8a8.ltfive} & \val{ctx.fp8e4m3w8a8.tflip} \\
MXFP4-32 & $\val{cnn.mxfp432w4a4.med}$ & $\val{cnn.mxfp432w4a4.pfive}$ & \val{cnn.mxfp432w4a4.ltfive} & \val{cnn.mxfp432w4a4.tflip} & $\val{ctx.mxfp432w4a4.med}$ & $\val{ctx.mxfp432w4a4.pfive}$ & \val{ctx.mxfp432w4a4.ltfive} & \val{ctx.mxfp432w4a4.tflip} \\
NVFP4-16 & $\val{cnn.nvfp416w4a4.med}$ & $\val{cnn.nvfp416w4a4.pfive}$ & \val{cnn.nvfp416w4a4.ltfive} & \val{cnn.nvfp416w4a4.tflip} & $\val{ctx.nvfp416w4a4.med}$ & $\val{ctx.nvfp416w4a4.pfive}$ & \val{ctx.nvfp416w4a4.ltfive} & \val{ctx.nvfp416w4a4.tflip} \\
MXFP4-32 + FP8 f/l & $\val{cnn.mxfp432w4a4fp8firstlast.med}$ & $\val{cnn.mxfp432w4a4fp8firstlast.pfive}$ & \val{cnn.mxfp432w4a4fp8firstlast.ltfive} & \val{cnn.mxfp432w4a4fp8firstlast.tflip} & $\val{ctx.mxfp432w4a4fp8firstlast.med}$ & $\val{ctx.mxfp432w4a4fp8firstlast.pfive}$ & \val{ctx.mxfp432w4a4fp8firstlast.ltfive} & \val{ctx.mxfp432w4a4fp8firstlast.tflip} \\
NVFP4-16 + FP8 f/l & $\val{cnn.nvfp416w4a4fp8firstlast.med}$ & $\val{cnn.nvfp416w4a4fp8firstlast.pfive}$ & \val{cnn.nvfp416w4a4fp8firstlast.ltfive} & \val{cnn.nvfp416w4a4fp8firstlast.tflip} & $\val{ctx.nvfp416w4a4fp8firstlast.med}$ & $\val{ctx.nvfp416w4a4fp8firstlast.pfive}$ & \val{ctx.nvfp416w4a4fp8firstlast.ltfive} & \val{ctx.nvfp416w4a4fp8firstlast.tflip} \\
NVFP4-16 + QAT & $\val{cnn.nvfp416w4a4qat.med}$ & $\val{cnn.nvfp416w4a4qat.pfive}$ & \val{cnn.nvfp416w4a4qat.ltfive} & \val{cnn.nvfp416w4a4qat.tflip} & $\val{ctx.nvfp416w4a4qat.med}$ & $\val{ctx.nvfp416w4a4qat.pfive}$ & \val{ctx.nvfp416w4a4qat.ltfive} & \val{ctx.nvfp416w4a4qat.tflip} \\
\bottomrule
\end{tabular}}
\end{table}

Table~\ref{tab:tail} and Fig.~\ref{fig:subject}a,b show the distribution of per-subject changes. Changing only the training seed already changes predictions. FP32 models trained with different seeds disagree on \val{cnn.noise.tflip}\% of epochs for SleepCNN and \val{ctx.noise.tflip}\% for ContextSleepNet. In this retraining reference, each seed pair is counted in both directions. Macro-F1 drops by more than 5\pp{} in \val{cnn.noise.ltfive}\% of the ordered subject--seed-pair comparisons for SleepCNN and \val{ctx.noise.ltfive}\% for ContextSleepNet. These repeated comparisons are descriptive; they do not count independent subjects.

Under MXFP4-32, SleepCNN loses more than in this reference. Its median change over the subject--seed pairs is $\val{cnn.mxfp432w4a4.med}$\pp{}. In \val{cnn.mxfp432w4a4.ltfive}\% of these pairs, the quantized model loses more than 5\pp{} relative to its matched FP32 model.

For ContextSleepNet, \val{ctx.mxfp432w4a4.ltfive}\% of the subject--seed pairs lose more than 5\pp{} under MXFP4-32, and \val{ctx.nvfp416w4a4.ltfive}\% do so under NVFP4-16. These shares are close to or below the \val{ctx.noise.ltfive}\% of its retraining reference. Its MXFP4-32 flip rate of \val{ctx.mxfp432w4a4.tflip}\% is also below its retraining flip rate of \val{ctx.noise.tflip}\%. The flip rate counts disagreements but does not show whether they help or hurt. We therefore read it together with the accuracy changes and the retraining reference.

FP8 first/last layers remove SleepCNN's large per-subject losses. Under either tested format, no subject--seed pair then loses more than 5\pp{} (Table~\ref{tab:tail}).

Under 4-bit PTQ, the same subjects lose accuracy in both architectures (Fig.~\ref{fig:subject}c). The Spearman correlation between the per-subject changes of SleepCNN and ContextSleepNet is \val{subj.mx.xmodel} under MXFP4-32 and \val{subj.nv.xmodel} under NVFP4-16 (both $p<0.001$). Computed separately for each training seed, these cross-architecture correlations range from \val{subj.xseed.min} to \val{subj.xseed.max}. Within each model, the correlations between seeds range from \val{subj.seed.min} to \val{subj.seed.max}. For retraining noise, the cross-architecture correlations of the signed changes range only from \val{subj.noise.xmodel.min} to \val{subj.noise.xmodel.max} (Fig.~\ref{fig:subject}d). This reference has a limit, however. The sign of a retraining difference is arbitrary for each subject and unrelated between the two architectures. Its correlation would therefore be near zero even if the same subjects had unstable predictions in both models. A quantization loss, in contrast, has a consistent sign. The retraining reference thus cannot show that the shared pattern is specific to quantization. EncoderNet's per-subject changes also correlate with those of both other models ($\rho=\val{abl.rho.min}$--$\val{abl.rho.max}$).

Subjects with larger 4-bit losses tend to have a lower FP32 Macro-F1. The correlations between FP32 Macro-F1 and the signed change range from $\rho=\val{subj.rbase.min}$ to $\val{subj.rbase.max}$ (all $p<0.05$). This association needs care, because the FP32 score also enters the change. Larger losses also go with a smaller share of N3 epochs ($\rho=\val{subj.rnthree.min}$--$\val{subj.rnthree.max}$, all $p<0.05$). This association alone does not show that N3$\to$N2 errors explain why some subjects are more vulnerable. Per-subject Macro-F1 is also less stable for subjects with few or no N3 epochs, which may contribute to this association.

Together, these exploratory results show that larger 4-bit losses recur in the same subjects across the tested architectures and training seeds. They suggest a shared, subject-level susceptibility in this cohort, but they do not identify its cause. In particular, they do not separate a susceptibility that is specific to quantization from a general instability of these subjects' predictions. The lower FP32 Macro-F1 of the affected subjects is compatible with both. A cohort mean alone cannot reveal which subjects are repeatedly affected.

\subsection{Checks of the protocol}
The gap-aware ContextSleepNet reaches \val{gapaw.fp32.mf1}\% pooled FP32 Macro-F1. Its mean paired per-subject difference from the main model is $\val{gapaw.fp32.d}$\pp{}, with a 95\% CI of [$\val{gapaw.fp32.lo}$, $\val{gapaw.fp32.hi}$]\pp{}.

We also compare the quantization-induced changes of the two versions under eight conditions. These are the four W4A4 formats, both FP8 first/last variants, INT8 W8A8, and FP8-E4M3 W8A8. Each version's change is measured against its own FP32 model. The mean differences between the two versions are at most \val{gapaw.diff.maxabs}\pp{} in absolute value, and every pointwise 95\% CI includes zero. These estimates agree with the main findings, although they do not establish equivalence. In this check, calibration still uses the original windows from unsplit nights.

For both models, native FP16 and BF16 execution predicts the same stage as the emulated formats for at least \val{nat.agree.min}\% of epochs. The mean per-subject Macro-F1 differs by at most \val{nat.diff.max}\pp{}.

Our MXFP4-32 and NVFP4-16 quantizers match the reference quantizers of TorchAO \val{tao.version} (pinned; PyTorch \val{tao.torch}). All \val{tao.exact} tested tensor--format pairs give identical dequantized FP32 values. The tested tensors are three random tensors and all 13 weight matrices of the BN-folded SleepCNN and ContextSleepNet models for fold 0, seed 0. Each tensor is tested in both formats. These checks establish agreement on the tested inputs.

Finally, the reproducibility checks report 100\% agreement with the stored predictions for both models. They cover reloaded FP32 checkpoints, rerun FP32 fine-tuning controls, and rerun first-layer W4A4 probes under MXFP4-32 and NVFP4-16.

\begin{table}
\centering
\caption{Native GPU inference on an NVIDIA RTX PRO 6000 Blackwell Server Edition (CUDA graphs, cuDNN autotuning, 50 warm-up and 200 timed calls, median). Latency at the smallest call size (1 epoch for SleepCNN; 31 epochs, one receptive field, for ContextSleepNet); throughput and dynamic energy (NVML board energy minus idle) at the largest call size (256 / 1{,}024 epochs).}
\label{tab:hw}
\setlength{\tabcolsep}{3pt}
\footnotesize
\begin{tabular}{l l r r r}
\toprule
Model & Dtype & Latency (ms) & M epochs/s & $\mu$J/epoch \\
\midrule
SleepCNN & FP32 & \val{hw.cnn.float32.lat} & \val{hw.cnn.float32.tput} & \val{hw.cnn.float32.uj} \\
 & FP16 & \val{hw.cnn.float16.lat} & \val{hw.cnn.float16.tput} & \val{hw.cnn.float16.uj} \\
 & BF16 & \val{hw.cnn.bfloat16.lat} & \val{hw.cnn.bfloat16.tput} & \val{hw.cnn.bfloat16.uj} \\
\midrule
ContextSleepNet & FP32 & \val{hw.ctx.float32.lat} & \val{hw.ctx.float32.tput} & \val{hw.ctx.float32.uj} \\
 & FP16 & \val{hw.ctx.float16.lat} & \val{hw.ctx.float16.tput} & \val{hw.ctx.float16.uj} \\
 & BF16 & \val{hw.ctx.bfloat16.lat} & \val{hw.ctx.bfloat16.tput} & \val{hw.ctx.bfloat16.uj} \\
\bottomrule
\end{tabular}
\end{table}

\subsection{Native GPU inference}
Table~\ref{tab:hw} reports native FP32, FP16, and BF16 inference. Both models are small, so kernel-launch overhead dominates single calls. Even with CUDA graphs, one SleepCNN epoch takes \val{hw.cnn.float32.lat}\,ms in FP32 and \val{hw.cnn.float16.lat}\,ms in FP16. CUDA graphs still cut this latency by a factor of \val{hw.cnn.graphx}. At large call sizes, FP16 raises the throughput of ContextSleepNet \val{hw.ctx.tputx}-fold. It also cuts its dynamic energy per epoch \val{hw.ctx.ujx}-fold, from \val{hw.ctx.float32.uj} to \val{hw.ctx.float16.uj}\,$\mu$J. For SleepCNN, FP16 improves throughput by a factor of \val{hw.cnn.tputx} and energy by a factor of \val{hw.cnn.ujx}. Across \val{hw.rep.n} measurement sessions on the same GPU model, throughput varied by at most \val{hw.rep.perf}\%. Dynamic energy per epoch, however, varied by up to \val{hw.rep.energy}\%, so the energy figures are only indicative.

At 256 epochs per call, FP8 GEMMs beat BF16 only on the first-layer im2col products, which have by far the most rows. For SleepCNN \texttt{conv1}, FP8 takes \val{gemm.sleepcnn.conv1.fp8}\,$\mu$s and BF16 takes \val{gemm.sleepcnn.conv1.bf16}\,$\mu$s. For ContextSleepNet \texttt{b1.conv}, the times are \val{gemm.contextsleepnet.b1conv.fp8} and \val{gemm.contextsleepnet.b1conv.bf16}\,$\mu$s. On the other tested layers, FP8 is slower than BF16 (e.g., SleepCNN \texttt{conv2}: \val{gemm.sleepcnn.conv2.fp8} vs.\ \val{gemm.sleepcnn.conv2.bf16}\,$\mu$s). The layer that is most sensitive to 4-bit quantization is thus also the layer where FP8 pays off on this GPU. We do not report speed for the 4-bit formats, because we only emulate them.

\section{Discussion}
\subsection{Implications for edge accelerators}
Four findings bear directly on hardware--software co-design for sleep-monitoring devices.

\emph{First, 8 bits were enough for these models.} FP16, BF16, INT8, and FP8-E4M3 change Macro-F1 by at most \val{eightbit.noefivemax}\pp{} for either model. For an EEG front end, we prefer INT8 or E4M3 to E5M2. Per-channel weight scales and calibrated activation ranges already cover the dynamic range, so the wider range of E5M2 is not needed. Its coarser mantissa, in contrast, costs SleepCNN a significant \val{cnn.fp8e5m2w8a8.dabs}\pp{}. This finding is limited to our two convolutional models and our quantizer. Xu et al.~\cite{xu2026inear} report a 5.01\% accuracy drop for static 8-bit quantization of a CNN-LSTM, and we did not test recurrent models.

\emph{Second, the scaling recipe matters at 4 bits.} NVFP4 (E4M3 block scales plus an FP32 tensor scale) beats MXFP4 (power-of-two block scales) in all four matched comparisons, three of them significantly. Halving the block size, in contrast, does not help MXFP4 and costs 0.25 extra bits per element.

\emph{Third, a 4-bit datapath should come with an 8-bit path for the first and last layers} (FP8-E4M3 in our tests). The first layer performs \val{ctx.firstmacpct}--\val{cnn.firstmacpct}\% of all MACs, so its precision limits the achievable savings. Relative to pure MXFP4-32 W4A4, FP8 first/last layers raise BOPs by a factor of \val{ctx.mxfp432w4a4.bopsx}--\val{cnn.mxfp432w4a4.bopsx}, but they add at most \val{mix.ovhmax}\% weight storage. Among the tested layers on our GPU, the first layer is also the only one where native FP8 GEMMs ran faster than BF16. The 8-bit path was sufficient in both cohorts only with NVFP4. With MXFP4-32, it left a significant loss for two of the three models on Sleep-EDF-20 (Section~\ref{sec:ablation}).

\emph{Fourth, the architecture can matter more than the format.} MXFP4-32 ContextSleepNet needs \val{ctx.mxfp432w4a4.kib}\,KiB of weights, less than FP32 SleepCNN (\val{cnn.fp32.kib}\,KiB). It needs \val{ctx.mxfp432w4a4.gbops}\,GBOPs per epoch instead of \val{cnn.fp32.gbops}\,GBOPs. Yet it reaches a higher pooled Macro-F1 (\val{ctx.mxfp432w4a4.mf1}\% vs.\ \val{cnn.fp32.mf1}\%). In this comparison, a fixed weight-storage budget was better spent on a larger model than on higher precision. The comparison counts weights only, not activation memory. The two models also differ in temporal context, and ContextSleepNet needs future epochs (Section~\ref{sec:models}). Our ablation suggests that the larger encoder or its training regime lowers the sensitivity to 4-bit error. On Sleep-EDF-20, however, this advantage held only for NVFP4.

For closed-loop applications, per-stage performance matters as much as Macro-F1. Slow-oscillation stimulation, for example, targets slow-wave (N3) sleep~\cite{ngo2013auditory}. Under MXFP4-32 W4A4, N3 recall drops by \val{cm.cnn.mx.nIIIrecall}\pp{} for SleepCNN and \val{cm.ctx.mx.nIIIrecall}\pp{} for ContextSleepNet. Devices built for closed-loop use should therefore be validated on N3 recall, not only on Macro-F1.

\subsection{What makes a model robust to 4-bit error?}
On Sleep-EDF-78, the same-encoder ablation shows that the temporal layers contribute little to the robustness of ContextSleepNet. Most of this robustness comes from non-temporal differences. One of them is the training regime. ContextSleepNet trains on mini-batches of 32-epoch windows with a budget of 40 training epochs. SleepCNN trains on mini-batches of 64 independent epochs with a budget of 30. Several architectural differences could also matter. The larger encoder has 64 first-layer filters instead of 16, it is deeper, and it uses BN, which is folded into the weights before quantization.

The first-layer probes point to a difference in the first layer. In SleepCNN, rounding either the weights or the input of \texttt{conv1} costs 0.5--1\pp{}, and rounding both costs more than the sum. In the larger encoder, neither operand alone has a detectable effect. One possible explanation is that 64 filters represent the raw EEG redundantly. Later layers could then average out the rounding errors of individual filters. We have not tested this idea, which would require varying only the width of the first layer. Moreover, on Sleep-EDF-20, the larger encoder was more sensitive to MXFP4 than SleepCNN. Each model was also trained on about one-quarter as many subjects there, so we cannot tell whether the cohort or the amount of training data causes this reversal. Either way, quantization robustness is not a fixed property of an architecture, and claims about it must be tested on independent cohorts.

\subsection{Reporting quantized clinical models}
The mean $\Delta$MF1 hides two patterns that our per-subject analysis reveals. First, the 4-bit loss concentrates in a subset of subjects. Second, the same subjects are affected in all tested architectures. A small mean loss therefore does not show that every patient is unaffected. We recommend that quantization studies on clinical signals report four items:
\begin{enumerate}
\renewcommand{\labelenumi}{(\roman{enumi})}
\item per-subject paired statistics with multiplicity correction;
\item the subject-level tail, next to a retraining-noise reference (for example, the median, the lower percentiles, and the share of subjects with a large loss);
\item a matched fine-tuning control and PTQ applied after the same fine-tuning, whenever QAT is evaluated;
\item ablations that change one architectural factor at a time, before robustness is attributed to that factor.
\end{enumerate}
Without (iii), ContextSleepNet's NVFP4 QAT result could have been read as an improvement over FP32. Without (iv), we would have credited its robustness to temporal context.

\subsection{An unexpected block-size effect}
For ContextSleepNet, MXFP4 with 16-element blocks performs slightly worse than with 32-element blocks ($\val{subj.con.ctx.blockmx.d}$\pp{}, $p_\mathrm{H}=\val{subj.con.ctx.blockmx.ph}$). Smaller blocks usually reduce quantization error, so this direction is unexpected. The round-down scale rule of MXFP4 offers one possible explanation, in line with the saturation reported for MX formats~\cite{mishra2025recipes}. After scaling, any block maximum that falls in $(6,8)$ saturates at the E2M1 limit of 6. Halving the block size doubles the number of blocks, and thus roughly doubles the number of block maxima that can saturate. The E4M3 block scales of NVFP4 place the block maximum near 6 and so largely avoid this failure mode. The evidence for this effect is thin, however. It rests on one contrast: for SleepCNN, the same change has no detectable effect ($\val{subj.con.cnn.blockmx.d}$\pp{}, $p_\mathrm{H}=\val{subj.con.cnn.blockmx.ph}$). We also did not test the explanation, for example by rounding the scale exponent up.

\subsection{Limitations}
(i) We emulate all formats with FP32 arithmetic. The value grids are exact, but the emulation does not model accumulator precision or kernel-level rounding. Latency and energy for MXFP4, NVFP4, and INT4 must come from native kernels or a hardware model.
(ii) We study one public cohort of healthy subjects scored with the R\&K rules. Clinical populations, such as patients with sleep apnea, and AASM-scored datasets may behave differently. Sleep-EDF-20 is a subset of this cohort, not an independent replication. Future work should test an external dataset and a published sleep-staging backbone.
(iii) We evaluated two architectures, one ablation, three seeds, one calibration-set size, and a QAT schedule of five training epochs. The ablation isolates the temporal layers. It cannot separate the encoder from its training regime or isolate individual encoder factors (width, depth, and BN). Under MXFP4, the architecture ranking also differed between Sleep-EDF-78 and the smaller Sleep-EDF-20 subset. That comparison confounds the cohort with the amount of training data.
(iv) ContextSleepNet is non-causal. A causal variant for real-time staging may be less robust.
(v) Our class-weighted loss favors Macro-F1 over accuracy, and we did not train an unweighted control. Each quantized model, however, is compared with its own FP32 model, which was trained with the same loss.
(vi) The subject-level correlation analyses and the 1-pp non-inferiority and equivalence margin are exploratory.
(vii) NVFP4 QAT uses dynamic activation tensor scales during fine-tuning and static scales after conversion. We did not evaluate a fully static QAT variant.
(viii) Three implementation choices limit how closely the emulation matches a device. Activation blocks are formed on unfolded input patches, so a value shared by overlapping patches is quantized once per patch. The skip path of the temporal layers stays in FP32, which may add to the low sensitivity of these layers. For EncoderNet and ContextSleepNet, the threshold search for INT8 and FP8 activations sees only part of the calibration data.
(ix) All runs share one assignment of subjects to folds; only the training seed varies. Per-subject Macro-F1 assigns zero to a stage that a subject lacks, and it is unstable for subjects with few epochs of a stage.

\section{Conclusion}
For single-channel EEG sleep staging, 8- and 16-bit formats cost at most \val{eightbit.maxabs}\pp{} of Macro-F1. They are also non-inferior to FP32 at an exploratory 1-pp margin. Pure W4A4 PTQ, in contrast, did not establish non-inferiority. Its cost depends on both the format and the model, and on Sleep-EDF-78 it depends little on temporal context. Two training-free choices close most of the gap. NVFP4 scaling loses less than MXFP4 scaling, and keeping the first and last layers in FP8 recovers most of the remaining loss. Combined, NVFP4-16 with FP8 first and last layers keeps every model within 0.25\pp{} of FP32 in both cohorts. QAT also restores Macro-F1 to near the FP32 level, but it does not beat an equally fine-tuned FP32 model. It is also equivalent within 1\pp{} to FP8 first and last layers without retraining. The 4-bit PTQ error is not spread evenly. It recurs in the same subjects across architectures, and it reduces N3 recall the most. These results, obtained in emulation, support pairing 4-bit compute with FP8 for the first and last layers in hardware for wearable sleep monitoring. Finally, evaluations of quantized clinical models should report per-subject effects against a retraining-noise reference and should test architectural explanations with ablations.

\FloatBarrier
\section*{Data and code availability}
Sleep-EDF Expanded is publicly available from PhysioNet~\cite{goldberger2000physionet}. A single notebook (PyTorch~\cite{paszke2019pytorch}, MNE-Python~\cite{gramfort2013mne}) produces all experiments. It downloads the data, runs the cross-validation, and saves every trained checkpoint and every prediction. The notebook also computes every control analysis, the pinned TorchAO comparison, and the per-stage spectra of Appendix~\ref{app:background}. Scripts released with the notebook generate the numbers, tables, and figures of this paper from these saved outputs. The release also includes unit tests of the number formats. Retraining on other GPUs or software versions can yield slightly different models. The released checkpoints and predictions reproduce the reported values exactly.

\section*{Acknowledgment}
The authors used an AI assistant for language and flow editing of this manuscript and consulted it on implementing the experiments in Python. The authors reviewed all code, results, and text and take full responsibility for the content of this paper.

\bibliographystyle{IEEEtran}
\bibliography{references}

@article{kemp2000sleepedf,
  author={Kemp, B. and Zwinderman, A.H. and Tuk, B. and Kamphuisen, H.A.C. and Oberye, J.J.L.},
  title={{Analysis of a sleep-dependent neuronal feedback loop: the slow-wave microcontinuity of the EEG}},
  journal={IEEE Transactions on Biomedical Engineering},
  year={2000},
  volume={47},
  number={9},
  pages={1185--1194},
  doi={10.1109/10.867928}
}

@inproceedings{kingma2015adam,
  title     = {Adam: A Method for Stochastic Optimization},
  author    = {Kingma, Diederik P. and Ba, Jimmy},
  booktitle = {International Conference on Learning Representations},
  year      = {2015},
  url       = {https://arxiv.org/abs/1412.6980}
}

@article{oreilly2014mass,
  author  = {O'Reilly, Christian and Gosselin, Nadia and Carrier, Julie and Nielsen, Tore},
  title   = {Montreal Archive of Sleep Studies: an open-access resource for instrument benchmarking and exploratory research},
  journal = {Journal of Sleep Research},
  volume  = {23},
  number  = {6},
  pages   = {628--635},
  year    = {2014},
  doi     = {10.1111/jsr.12169}
}

@article{goldberger2000physionet,
  author={Goldberger, Ary L. and Amaral, Luis A. N. and Glass, Leon and Hausdorff, Jeffrey M. and Ivanov, Plamen Ch. and Mark, Roger G. and Mietus, Joseph E. and Moody, George B. and Peng, Chung-Kang and Stanley, H. Eugene},
  title={{PhysioBank, PhysioToolkit, and PhysioNet}},
  journal={Circulation},
  year={2000},
  volume={101},
  number={23},
  pages={e215--e220},
  doi={10.1161/01.CIR.101.23.e215}
}

@article{supratak2017deepsleepnet,
  author={Supratak, Akara and Dong, Hao and Wu, Chao and Guo, Yike},
  title={{DeepSleepNet: A Model for Automatic Sleep Stage Scoring Based on Raw Single-Channel EEG}},
  journal={IEEE Transactions on Neural Systems and Rehabilitation Engineering},
  year={2017},
  volume={25},
  number={11},
  pages={1998--2008},
  doi={10.1109/TNSRE.2017.2721116}
}

@inproceedings{supratak2020tinysleepnet,
  author={Supratak, Akara and Guo, Yike},
  title={{TinySleepNet: An Efficient Deep Learning Model for Sleep Stage Scoring based on Raw Single-Channel EEG}},
  booktitle={2020 42nd Annual International Conference of the IEEE Engineering in Medicine \& Biology Society (EMBC)},
  year={2020},
  pages={641--644},
  doi={10.1109/EMBC44109.2020.9176741}
}

@article{phan2019seqsleepnet,
  author={Phan, Huy and Andreotti, Fernando and Cooray, Navin and Chen, Oliver Y. and De Vos, Maarten},
  title={{SeqSleepNet: End-to-End Hierarchical Recurrent Neural Network for Sequence-to-Sequence Automatic Sleep Staging}},
  journal={IEEE Transactions on Neural Systems and Rehabilitation Engineering},
  year={2019},
  volume={27},
  number={3},
  pages={400--410},
  doi={10.1109/TNSRE.2019.2896659}
}

@article{phan2022sleeptransformer,
  author={Phan, Huy and Mikkelsen, Kaare and Chen, Oliver Y. and Koch, Philipp and Mertins, Alfred and De Vos, Maarten},
  title={{SleepTransformer: Automatic Sleep Staging With Interpretability and Uncertainty Quantification}},
  journal={IEEE Transactions on Biomedical Engineering},
  year={2022},
  volume={69},
  number={8},
  pages={2456--2467},
  doi={10.1109/TBME.2022.3147187}
}

@article{eldele2021attnsleep,
  author={Eldele, Emadeldeen and Chen, Zhenghua and Liu, Chengyu and Wu, Min and Kwoh, Chee-Keong and Li, Xiaoli and Guan, Cuntai},
  title={{An Attention-Based Deep Learning Approach for Sleep Stage Classification With Single-Channel EEG}},
  journal={IEEE Transactions on Neural Systems and Rehabilitation Engineering},
  year={2021},
  volume={29},
  pages={809--818},
  doi={10.1109/TNSRE.2021.3076234}
}

@inproceedings{jacob2018quantization,
  author={Jacob, Benoit and Kligys, Skirmantas and Chen, Bo and Zhu, Menglong and Tang, Matthew and Howard, Andrew and Adam, Hartwig and Kalenichenko, Dmitry},
  title={{Quantization and Training of Neural Networks for Efficient Integer-Arithmetic-Only Inference}},
  booktitle={2018 IEEE/CVF Conference on Computer Vision and Pattern Recognition},
  year={2018},
  pages={2704--2713},
  doi={10.1109/CVPR.2018.00286}
}

@article{wilcoxon1945,
  author={Wilcoxon, Frank},
  title={{Individual Comparisons by Ranking Methods}},
  journal={Biometrics Bulletin},
  year={1945},
  volume={1},
  number={6},
  pages={80--83},
  doi={10.2307/3001968}
}

@article{cohen1960kappa,
  author={Cohen, Jacob},
  title={{A Coefficient of Agreement for Nominal Scales}},
  journal={Educational and Psychological Measurement},
  year={1960},
  volume={20},
  number={1},
  pages={37--46},
  doi={10.1177/001316446002000104}
}

@article{dankerhopfe2009interrater,
  author={Danker-Hopfe, Heidi and Anderer, Peter and Zeitlhofer, Josef and Boeck, Marion and Dorn, Hans and Gruber, Georg and Heller, Esther and Loretz, Erna and Moser, Doris and Parapatics, Silvia and Saletu, Bernd and Schmidt, Andrea and Dorffner, Georg},
  title={{Interrater reliability for sleep scoring according to the Rechtschaffen {\&} Kales and the new AASM standard}},
  journal={Journal of Sleep Research},
  year={2009},
  volume={18},
  number={1},
  pages={74--84},
  doi={10.1111/j.1365-2869.2008.00700.x}
}

@article{rosenberg2013aasm,
  author={Rosenberg, Richard S. and Van Hout, Steven},
  title={{The American Academy of Sleep Medicine Inter-scorer Reliability Program: Sleep Stage Scoring}},
  journal={Journal of Clinical Sleep Medicine},
  year={2013},
  volume={9},
  number={1},
  pages={81--87},
  doi={10.5664/jcsm.2350}
}

@article{gramfort2013mne,
  author={Gramfort, Alexandre and others},
  title={{MEG and EEG data analysis with MNE-Python}},
  journal={Frontiers in Neuroscience},
  year={2013},
  volume={7},
  pages={267},
  doi={10.3389/fnins.2013.00267}
}

@misc{rouhani2023microscaling,
  author={Bita Darvish Rouhani and Ritchie Zhao and Ankit More and Mathew Hall and Alireza Khodamoradi and Summer Deng and Dhruv Choudhary and Marius Cornea and Eric Dellinger and Kristof Denolf and Stosic Dusan and Venmugil Elango and others},
  title={{Microscaling Data Formats for Deep Learning}},
  howpublished={arXiv preprint arXiv:2310.10537},
  year={2023}
}

@misc{nvidia2025nvfp4,
  author={{NVIDIA}},
  title={{Pretraining Large Language Models with NVFP4}},
  howpublished={arXiv preprint arXiv:2509.25149},
  year={2025}
}

@misc{micikevicius2022fp8,
  author={Paulius Micikevicius and Dusan Stosic and Neil Burgess and Marius Cornea and Pradeep Dubey and Richard Grisenthwaite and Sangwon Ha and Alexander Heinecke and Patrick Judd and John Kamalu and Naveen Mellempudi and Stuart Oberman and others},
  title={{FP8 Formats for Deep Learning}},
  howpublished={arXiv preprint arXiv:2209.05433},
  year={2022}
}

@misc{nagel2021white,
  author={Markus Nagel and Marios Fournarakis and Rana Ali Amjad and Yelysei Bondarenko and Mart van Baalen and Tijmen Blankevoort},
  title={{A White Paper on Neural Network Quantization}},
  howpublished={arXiv preprint arXiv:2106.08295},
  year={2021}
}

@misc{krishnamoorthi2018quantizing,
  author={Raghuraman Krishnamoorthi},
  title={{Quantizing deep convolutional networks for efficient inference: A whitepaper}},
  howpublished={arXiv preprint arXiv:1806.08342},
  year={2018}
}

@misc{bengio2013estimating,
  author={Yoshua Bengio and Nicholas L{\'e}onard and Aaron Courville},
  title={{Estimating or Propagating Gradients Through Stochastic Neurons for Conditional Computation}},
  howpublished={arXiv preprint arXiv:1308.3432},
  year={2013}
}

@misc{hooker2019forget,
  author={Sara Hooker and Aaron Courville and Gregory Clark and Yann Dauphin and Andrea Frome},
  title={{What Do Compressed Deep Neural Networks Forget?}},
  howpublished={arXiv preprint arXiv:1911.05248},
  year={2019}
}

@inproceedings{dutta2024accuracy,
  author={Abhinav Dutta and Sanjeev Krishnan and Nipun Kwatra and Ramachandran Ramjee},
  title={{Accuracy is Not All You Need}},
  booktitle={Advances in Neural Information Processing Systems (NeurIPS)},
  year={2024},
  note={arXiv:2407.09141}
}

@inproceedings{esser2020lsq,
  author={Steven K. Esser and Jeffrey L. McKinstry and Deepika Bablani and Rathinakumar Appuswamy and Dharmendra S. Modha},
  title={{Learned Step Size Quantization}},
  booktitle={International Conference on Learning Representations (ICLR)},
  year={2020},
  note={arXiv:1902.08153}
}

@inproceedings{dong2019hawq,
  author={Zhen Dong and Zhewei Yao and Amir Gholami and Michael Mahoney and Kurt Keutzer},
  title={{HAWQ: Hessian AWare Quantization of Neural Networks with Mixed-Precision}},
  booktitle={IEEE/CVF International Conference on Computer Vision (ICCV)},
  year={2019},
  note={arXiv:1905.03696}
}

@inproceedings{perslev2019utime,
  author={Mathias Perslev and Michael Hejselbak Jensen and Sune Darkner and Poul J{\o}rgen Jennum and Christian Igel},
  title={{U-Time: A Fully Convolutional Network for Time Series Segmentation Applied to Sleep Staging}},
  booktitle={Advances in Neural Information Processing Systems (NeurIPS)},
  year={2019},
  note={arXiv:1910.11162}
}

@inproceedings{paszke2019pytorch,
  author={Adam Paszke and Sam Gross and Francisco Massa and Adam Lerer and James Bradbury and Gregory Chanan and Trevor Killeen and Zeming Lin and Natalia Gimelshein and Luca Antiga and Alban Desmaison and Andreas K{\"o}pf and others},
  title={{PyTorch: An Imperative Style, High-Performance Deep Learning Library}},
  booktitle={Advances in Neural Information Processing Systems (NeurIPS)},
  year={2019},
  note={arXiv:1912.01703}
}

@inproceedings{or2025torchao,
  author={Andrew Or and Apurva Jain and Daniel Vega-Myhre and Jesse Cai and Charles David Hernandez and Zhenrui Zheng and Driss Guessous and Vasiliy Kuznetsov and Christian Puhrsch and Mark Saroufim and Supriya Rao and Thien Tran and others},
  title={{TorchAO: PyTorch-Native Training-to-Serving Model Optimization}},
  booktitle={ICML 2025 Workshop on Championing Open-source Development in ML (CODEML)},
  year={2025},
  note={arXiv:2507.16099}
}

@misc{choi2018pact,
  author={Jungwook Choi and Zhuo Wang and Swagath Venkataramani and Pierce I-Jen Chuang and Vijayalakshmi Srinivasan and Kailash Gopalakrishnan},
  title={{PACT: Parameterized Clipping Activation for Quantized Neural Networks}},
  howpublished={arXiv preprint arXiv:1805.06085},
  year={2018}
}

@article{holm1979simple,
  author={Holm, Sture},
  title={{A Simple Sequentially Rejective Multiple Test Procedure}},
  journal={Scandinavian Journal of Statistics},
  year={1979},
  volume={6},
  number={2},
  pages={65--70}
}

@book{efron1993bootstrap,
  author={Efron, Bradley and Tibshirani, Robert J.},
  title={{An Introduction to the Bootstrap}},
  publisher={Chapman \& Hall},
  address={New York},
  year={1993}
}

@book{iber2007aasm,
  author={Iber, Conrad and Ancoli-Israel, Sonia and Chesson, Andrew L. and Quan, Stuart F.},
  title={{The AASM Manual for the Scoring of Sleep and Associated Events: Rules, Terminology and Technical Specifications}},
  publisher={American Academy of Sleep Medicine},
  address={Westchester, IL},
  year={2007}
}

@book{rechtschaffen1968manual,
  author={Rechtschaffen, Allan and Kales, Anthony},
  title={{A Manual of Standardized Terminology, Techniques and Scoring System for Sleep Stages of Human Subjects}},
  publisher={U.S. Government Printing Office},
  address={Washington, DC},
  year={1968}
}

@techreport{ocp2023mx,
  author={{Open Compute Project}},
  title={{OCP Microscaling Formats (MX) Specification, Version 1.0}},
  institution={Open Compute Project},
  year={2023}
}

@article{yang2023lwsleepnet,
  author={Yang, Chenguang and Li, Baozhu and Li, Yamei and He, Yixuan and Zhang, Yuan},
  title={{LWSleepNet: A lightweight attention-based deep learning model for sleep staging with singlechannel EEG}},
  journal={Digital Health},
  year={2023},
  volume={9},
  pages={20552076231188206},
  doi={10.1177/20552076231188206}
}

@article{jasper1958tentwenty,
  author  = {Jasper, Herbert H.},
  title   = {Report of the committee on methods of clinical examination in electroencephalography},
  journal = {Electroencephalography and Clinical Neurophysiology},
  volume  = {10},
  number  = {2},
  pages   = {370--375},
  year    = {1958},
  doi     = {10.1016/0013-4694(58)90053-1}
}

@article{yuste2017four,
  author = {Yuste, Rafael and Goering, Sara and {Ag\"{u}era y Arcas}, Blaise and others},
  title = {{Four ethical priorities for neurotechnologies and AI}},
  journal = {Nature},
  year = {2017},
  volume = {551},
  number = {7679},
  pages = {159--163},
  doi = {10.1038/551159a}
}

@article{ienca2017towards,
  author = {Ienca, Marcello and Andorno, Roberto},
  title = {{Towards new human rights in the age of neuroscience and neurotechnology}},
  journal = {Life Sciences, Society and Policy},
  year = {2017},
  volume = {13},
  number = {1},
  pages = {5},
  doi = {10.1186/s40504-017-0050-1}
}

@article{verma2010micropower,
  author = {Verma, Naveen and Shoeb, Ali and Bohorquez, Jose and Dawson, Joel and Guttag, John and Chandrakasan, Anantha P.},
  title = {{A Micro-Power EEG Acquisition SoC With Integrated Feature Extraction Processor for a Chronic Seizure Detection System}},
  journal = {IEEE Journal of Solid-State Circuits},
  year = {2010},
  volume = {45},
  number = {4},
  pages = {804--816},
  doi = {10.1109/JSSC.2010.2042245}
}

@article{ngo2013auditory,
  author = {Ngo, Hong-Viet V. and Martinetz, Thomas and Born, Jan and M{\"o}lle, Matthias},
  title = {{Auditory Closed-Loop Stimulation of the Sleep Slow Oscillation Enhances Memory}},
  journal = {Neuron},
  year = {2013},
  volume = {78},
  number = {3},
  pages = {545--553},
  doi = {10.1016/j.neuron.2013.03.006}
}

@article{arnal2020dreem,
  author = {Arnal, Pierrick J. and Thorey, Valentin and Debellemaniere, Eden and Ballard, Michael E. and Bou Hernandez, Albert and Guillot, Antoine and Jourde, Hugo and Harris, Mason and Guillard, Mathias and Van Beers, Pascal and Chennaoui, Mounir and Sauvet, Fabien},
  title = {{The Dreem Headband compared to polysomnography for electroencephalographic signal acquisition and sleep staging}},
  journal = {Sleep},
  year = {2020},
  volume = {43},
  number = {11},
  pages = {zsaa097},
  doi = {10.1093/sleep/zsaa097}
}

@article{dezambotti2019wearable,
  author = {{de Zambotti}, Massimiliano and Cellini, Nicola and Goldstone, Aim{\'e}e and Colrain, Ian M. and Baker, Fiona C.},
  title = {{Wearable Sleep Technology in Clinical and Research Settings}},
  journal = {Medicine \& Science in Sports \& Exercise},
  year = {2019},
  volume = {51},
  number = {7},
  pages = {1538--1557},
  doi = {10.1249/MSS.0000000000001947}
}

@article{park2025distillsleep,
  author = {Park, Keondo and Hong, Joopyo and Lee, Wooseok and Shin, Hyun-Woo and Kim, Hyung-Sin},
  title = {{DistillSleep}: real-time, on-device, interpretable sleep staging from single-channel electroencephalogram},
  journal = {Sleep},
  year = {2025},
  volume = {48},
  number = {12},
  pages = {zsaf240},
  doi = {10.1093/sleep/zsaf240}
}

@misc{robitaille2025sleepvit,
  author = {Robitaille, Tristan and Liu, Xilin},
  title = {Vision Transformer Accelerator {ASIC} for Real-Time Low-Power Sleep Staging},
  year = {2025},
  howpublished = {arXiv preprint arXiv:2502.16334}
}

@article{chen2026mixed,
  author = {Chen, Xinyu and Xu, Yaqian and Hou, Yuhan and Liu, Bowen and Richardson, Andrew G. and Liu, Xilin},
  title = {A Mixed-Precision {CNN-LSTM} Accelerator Featuring Long-Kernel Memory Reuse and Dynamic Quantization for Sleep Staging},
  journal = {IEEE Solid-State Circuits Letters},
  year = {2026},
  note = {early access; arXiv:2611.00660},
  doi = {10.1109/LSSC.2026.3746976}
}

@misc{iqbal2018plotneuralnet,
  author = {Iqbal, Haris},
  title = {{PlotNeuralNet}: {LaTeX} code for drawing neural networks for reports and presentation},
  year = {2018},
  howpublished = {\url{https://github.com/HarisIqbal88/PlotNeuralNet}},
  note = {{MIT} License}
}

@misc{xu2026inear,
  author = {Xu, Yaqian and Chen, Xinyu and Hou, Yuhan and Xu, Jianxiong and You, Hao and Huang, Yu and Liu, Bowen and Hung, Ashley and Richardson, Andrew G. and Genov, Roman and Liu, Xilin},
  title = {An In-Ear Sleep Modulation {SoC} Featuring a {CNN-LSTM} Accelerator with Long-Kernel Memory Reuse and Runtime Dynamic Quantization},
  year = {2026},
  howpublished = {arXiv preprint arXiv:2611.11103},
  doi = {10.1109/VLSITECHNOLOGYANDCIR65830.2026.11577635}
}

@misc{mishra2025recipes,
  author = {Mishra, Asit and Stosic, Dusan and Layton, Simon and Micikevicius, Paulius},
  title = {Recipes for Pre-training {LLMs} with {MXFP8}},
  year = {2025},
  howpublished = {arXiv preprint arXiv:2506.08027}
}

@misc{chhugani2026mxfp4,
  author = {Chhugani, Jatin and Jeong, Geonhwa and Su, Bor-Yiing and Pan, Yunjie and Yang, Hanmei and Ankit, Aayush and Yu, Jiecao and Deng, Summer and Chen, Yunqing and Satish, Nadathur and Kim, Changkyu},
  title = {Unveiling the Potential of Quantization with {MXFP4}: Strategies for Quantization Error Reduction},
  year = {2026},
  howpublished = {arXiv preprint arXiv:2603.08713}
}

\appendices
\section{Background: EEG, sleep stages, and the models}
\label{app:background}
\appfloatsetup
\textbf{EEG rhythms.} The EEG measures voltage differences between scalp electrodes. These differences reflect the summed activity of cortical neurons. EEG rhythms are grouped into five frequency bands: delta (0.5--4\,Hz), theta (4--8\,Hz), alpha (8--13\,Hz), sigma (about 11--16\,Hz), and beta (13--30\,Hz). The \emph{slow waves} of deep sleep lie at 0.5--2\,Hz, within the delta band. Sleep spindles lie in the sigma band. Faster (gamma) activity is not used in visual sleep scoring.

\begin{figure}
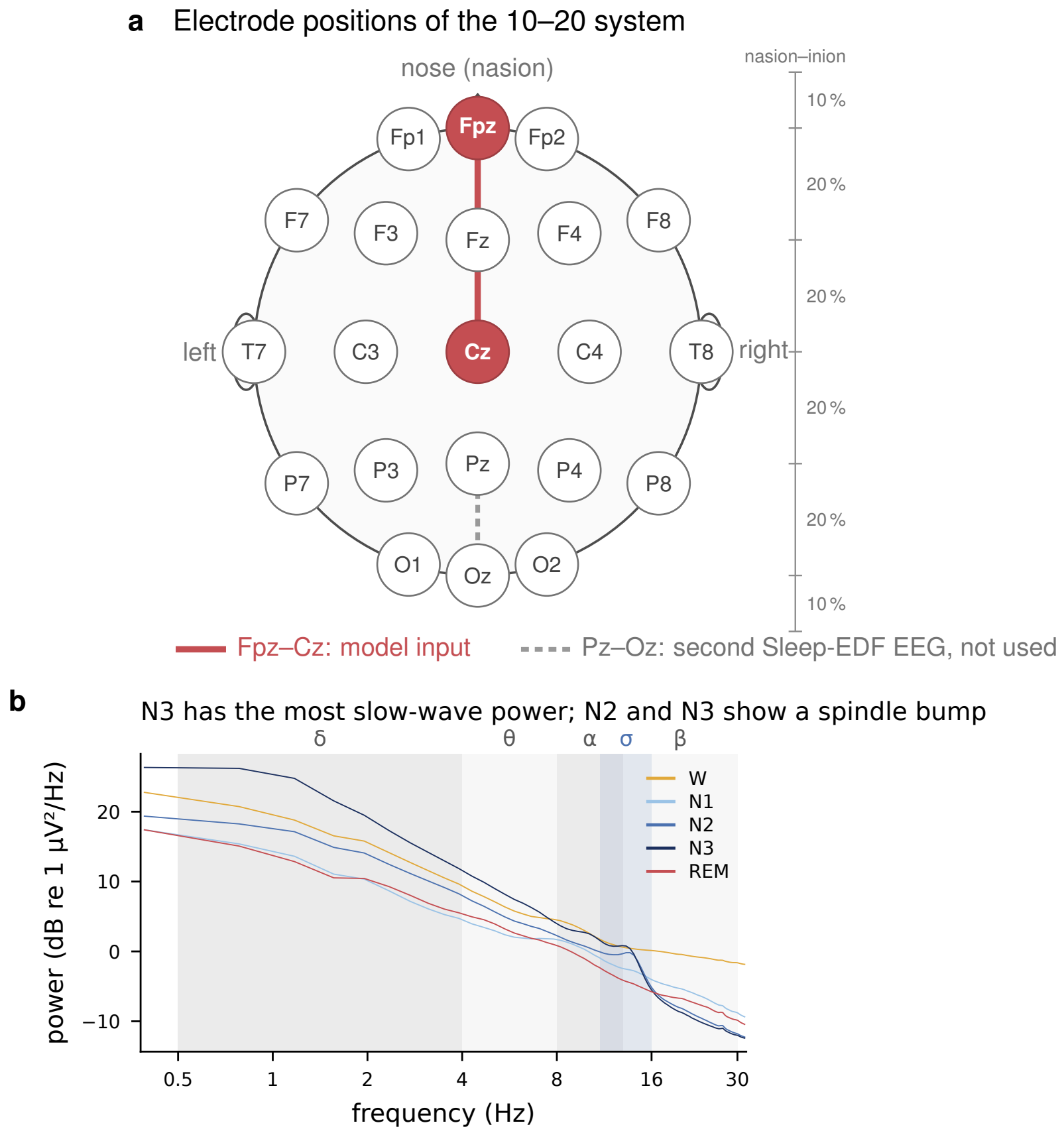

\centering
\signalfig
\caption{Signal background. (a) Electrode positions of the international 10--20 system (top view, nose up). Along the midline, Fpz lies 10\% of the nasion--inion distance from the nasion, and neighboring electrodes are 20\% apart. The model input is the Fpz--Cz derivation (red). The second Sleep-EDF EEG derivation, Pz--Oz (dashed), is not used. (b) Mean log power spectral density of the Fpz--Cz signal per stage over all \val{data.total} epochs, in dB relative to 1\,$\mu$V$^2$/Hz before normalization (Welch's method, 2.56-s Hann windows, 50\% overlap; notebook Cell 21). Shaded: delta, theta, alpha, sigma, and beta bands. The band edges are approximate, because the sigma band overlaps the conventional alpha and beta ranges.}
\label{fig:signal}
\end{figure}

\textbf{Electrode position.} Sleep-EDF records two EEG derivations: Fpz--Cz and Pz--Oz~\cite{kemp2000sleepedf}. We use Fpz--Cz, the voltage between the forehead midline (Fpz) and the vertex (Cz) in the international 10--20 system~\cite{jasper1958tentwenty} (Fig.~\ref{fig:signal}a). Because Fpz--Cz is a frontal derivation, it picks up the posterior alpha rhythm of relaxed wakefulness only weakly. In addition, human scorers use eye-movement and chin-muscle channels to recognize REM sleep, but these channels are not part of our input. Eye movements appear in Fpz--Cz only as artifacts. Both factors make single-channel staging harder than expert scoring with a full montage.

\begin{table}
\centering
\caption{Sleep stages: main EEG criteria (simplified from~\cite{iber2007aasm,rechtschaffen1968manual}) and share of the \val{data.total} epochs used here.}
\label{tab:stages}
\setlength{\tabcolsep}{3pt}
\footnotesize
\fitcol{%
\begin{tabular}{l >{\raggedright\arraybackslash}p{\stagefeatw} >{\raggedright\arraybackslash}p{\stagerkw} r}
\toprule
AASM & Main EEG features & R\&K & Share \\
\midrule
W & Alpha rhythm (8--13\,Hz) with eyes closed, beta activity, eye blinks & W & \val{data.wpct}\% \\
N1 & Alpha replaced by low-amplitude, mixed-frequency (mostly 4--7\,Hz) activity; vertex sharp waves; slow eye movements & S1 & \val{data.nIpct}\% \\
N2 & K-complexes and/or sleep spindles (11--16\,Hz) on a low-amplitude, mixed-frequency background & S2 & \val{data.nIIpct}\% \\
N3 & Slow waves (0.5--2\,Hz, $>$75\,$\mu$V peak to peak) in $\ge$20\% of the epoch & S3 (20--50\%) + S4 ($>$50\%) & \val{data.nIIIpct}\% \\
REM & Low-amplitude, mixed-frequency EEG (sometimes sawtooth waves), rapid eye movements, low chin-muscle tone & REM & \val{data.rempct}\% \\
\bottomrule
\end{tabular}}
\end{table}

\textbf{Sleep stages.} Table~\ref{tab:stages} summarizes the scoring criteria~\cite{iber2007aasm,rechtschaffen1968manual}. The R\&K rules split deep sleep into S3 and S4 by how much of the epoch contains slow waves. The AASM rules merge both stages into N3, so we relabel every S3 and S4 epoch as N3. Figure~\ref{fig:signal}b shows the mean spectrum of each stage in our data. N3 has the most slow-wave power (0.5--2\,Hz). N2 and N3 both show a bump in the spindle range (12--14\,Hz). It rises \val{spec.bump.nII} and \val{spec.bump.nIII}\,dB, respectively, above the neighboring 9--11 and 15--17\,Hz bands. W has by far the most power at 15--30\,Hz, at least \val{spec.wbeta.margin}\,dB more than any other stage. The spectra of N1 and REM, in contrast, differ by at most \val{spec.nIrem.maxdb}\,dB between 0.5 and 30\,Hz. This similarity helps explain why the two stages are often confused when eye-movement and muscle channels are missing.

\begin{figure}[htbp]
\centering
\includegraphics[width=\textwidth]{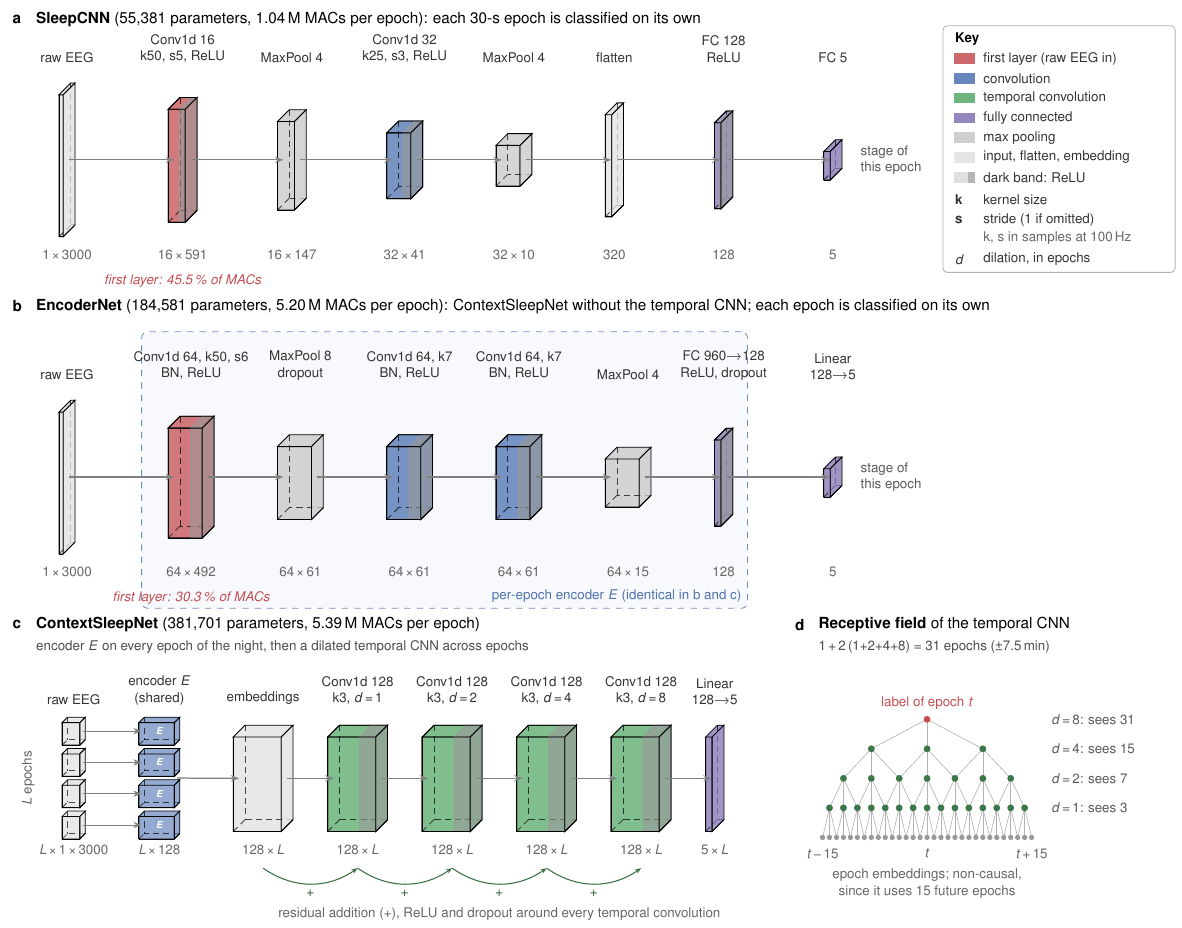}
\caption{The three architectures. In (a) and (b), box width grows with the number of channels and box height with the sequence length (both compressed). Vectors are drawn as thin boxes. (a) SleepCNN classifies each 30-s epoch on its own. (b) EncoderNet, the ablation model (Section~\ref{sec:controls}), applies the per-epoch encoder of ContextSleepNet (dashed frame) and a linear classifier to each epoch. It uses no information from neighboring epochs. (c) ContextSleepNet runs the same encoder on every epoch of the night. It combines the 128-dimensional embeddings with four residual dilated temporal convolutions. (d) With kernel 3 and dilations 1, 2, 4, and 8, the label of epoch $t$ depends on epochs $t-15$ to $t+15$. Block labels: k, kernel size; s, stride (both in samples at 100\,Hz; stride 1 if omitted); $d$, dilation of the temporal convolutions in epochs. Numbers under the blocks are output shapes (channels $\times$ length; $L$: number of epochs). The first layer (red), which receives the raw EEG, is the most sensitive to 4-bit quantization (Section~\ref{sec:mechanism}). The red labels give its share of all MACs in SleepCNN and in EncoderNet (\val{ctx.firstmacpct}\% in ContextSleepNet). Block diagrams use the PlotNeuralNet TikZ styles~\cite{iqbal2018plotneuralnet}.}
\label{fig:arch}
\end{figure}

\textbf{Epochs, windows, and calibration.} Epochs do not overlap, so each 30-s segment has one label. ContextSleepNet trains on random windows of 32 consecutive epochs and labels a whole night in one pass. Each prediction uses a 31-epoch receptive field (Fig.~\ref{fig:arch}d). For PTQ calibration, SleepCNN receives the stratified calibration epochs directly. For ContextSleepNet, these epochs are the centers of 31-epoch windows (Section~\ref{sec:protocol}).

\FloatBarrier
\section{Sleep-EDF-20 subset}
\label{app:sc20}
\begin{table*}
\centering
\caption{Results on the nested Sleep-EDF-20 subset (subjects 0--19; \val{sc20.total} epochs), retrained with the same pipeline, three seeds, and 10-fold subject-wise CV (two test subjects per fold). $\Delta$MF1 vs the FP32 model of the same architecture (pp, mean and 95\% bootstrap CI over $n=20$ subjects). $^{*}$/$^{**}$: Holm-corrected Wilcoxon $p<0.05/0.01$ within each model (17 conditions). Sleep-EDF-20 is a subset of Sleep-EDF-78, so this is not an independent replication.}
\label{tab:sc20}
\setlength{\tabcolsep}{3pt}
\footnotesize
\fitwide{%
\begin{tabular}{l lll}
\toprule
Format & SleepCNN & EncoderNet & ContextSleepNet \\
\midrule
FP32 MF1 (\%) / $\kappa$ & \val{sc20.cnn.fp32.mf1} / \val{sc20.cnn.fp32.kappa} & \val{sc20.enc.fp32.mf1} / \val{sc20.enc.fp32.kappa} & \val{sc20.ctx.fp32.mf1} / \val{sc20.ctx.fp32.kappa} \\
\midrule
FP16 & $\val{sc20.cnn.fp16.d}$ {\scriptsize[$\val{sc20.cnn.fp16.lo}$, $\val{sc20.cnn.fp16.hi}$]} & $\val{sc20.enc.fp16.d}$ {\scriptsize[$\val{sc20.enc.fp16.lo}$, $\val{sc20.enc.fp16.hi}$]} & $\val{sc20.ctx.fp16.d}$ {\scriptsize[$\val{sc20.ctx.fp16.lo}$, $\val{sc20.ctx.fp16.hi}$]} \\
BF16 & $\val{sc20.cnn.bf16.d}$ {\scriptsize[$\val{sc20.cnn.bf16.lo}$, $\val{sc20.cnn.bf16.hi}$]} & $\val{sc20.enc.bf16.d}$ {\scriptsize[$\val{sc20.enc.bf16.lo}$, $\val{sc20.enc.bf16.hi}$]} & $\val{sc20.ctx.bf16.d}$ {\scriptsize[$\val{sc20.ctx.bf16.lo}$, $\val{sc20.ctx.bf16.hi}$]} \\
\midrule
FP8-E4M3 & $\val{sc20.cnn.fp8e4m3w8a8.d}$ {\scriptsize[$\val{sc20.cnn.fp8e4m3w8a8.lo}$, $\val{sc20.cnn.fp8e4m3w8a8.hi}$]} & $\val{sc20.enc.fp8e4m3w8a8.d}$ {\scriptsize[$\val{sc20.enc.fp8e4m3w8a8.lo}$, $\val{sc20.enc.fp8e4m3w8a8.hi}$]} & $\val{sc20.ctx.fp8e4m3w8a8.d}$ {\scriptsize[$\val{sc20.ctx.fp8e4m3w8a8.lo}$, $\val{sc20.ctx.fp8e4m3w8a8.hi}$]} \\
FP8-E5M2 & $\val{sc20.cnn.fp8e5m2w8a8.d}$ {\scriptsize[$\val{sc20.cnn.fp8e5m2w8a8.lo}$, $\val{sc20.cnn.fp8e5m2w8a8.hi}$]} & $\val{sc20.enc.fp8e5m2w8a8.d}$ {\scriptsize[$\val{sc20.enc.fp8e5m2w8a8.lo}$, $\val{sc20.enc.fp8e5m2w8a8.hi}$]} & $\val{sc20.ctx.fp8e5m2w8a8.d}$ {\scriptsize[$\val{sc20.ctx.fp8e5m2w8a8.lo}$, $\val{sc20.ctx.fp8e5m2w8a8.hi}$]} \\
INT8 & $\val{sc20.cnn.int8w8a8.d}$ {\scriptsize[$\val{sc20.cnn.int8w8a8.lo}$, $\val{sc20.cnn.int8w8a8.hi}$]} & $\val{sc20.enc.int8w8a8.d}$ {\scriptsize[$\val{sc20.enc.int8w8a8.lo}$, $\val{sc20.enc.int8w8a8.hi}$]} & $\val{sc20.ctx.int8w8a8.d}$ {\scriptsize[$\val{sc20.ctx.int8w8a8.lo}$, $\val{sc20.ctx.int8w8a8.hi}$]} \\
INT8 (W only) & $\val{sc20.cnn.int8w8.d}$ {\scriptsize[$\val{sc20.cnn.int8w8.lo}$, $\val{sc20.cnn.int8w8.hi}$]} & $\val{sc20.enc.int8w8.d}$ {\scriptsize[$\val{sc20.enc.int8w8.lo}$, $\val{sc20.enc.int8w8.hi}$]} & $\val{sc20.ctx.int8w8.d}$ {\scriptsize[$\val{sc20.ctx.int8w8.lo}$, $\val{sc20.ctx.int8w8.hi}$]} \\
\midrule
INT4-g32 (W only) & $\val{sc20.cnn.int4g32w4.d}$ {\scriptsize[$\val{sc20.cnn.int4g32w4.lo}$, $\val{sc20.cnn.int4g32w4.hi}$]} & $\val{sc20.enc.int4g32w4.d}$ {\scriptsize[$\val{sc20.enc.int4g32w4.lo}$, $\val{sc20.enc.int4g32w4.hi}$]} & $\val{sc20.ctx.int4g32w4.d}$ {\scriptsize[$\val{sc20.ctx.int4g32w4.lo}$, $\val{sc20.ctx.int4g32w4.hi}$]} \\
\midrule
MXFP4-32 & $\val{sc20.cnn.mxfp432w4a4.d}$ {\scriptsize[$\val{sc20.cnn.mxfp432w4a4.lo}$, $\val{sc20.cnn.mxfp432w4a4.hi}$]} & $\val{sc20.enc.mxfp432w4a4.d}$ {\scriptsize[$\val{sc20.enc.mxfp432w4a4.lo}$, $\val{sc20.enc.mxfp432w4a4.hi}$]}$^{**}$ & $\val{sc20.ctx.mxfp432w4a4.d}$ {\scriptsize[$\val{sc20.ctx.mxfp432w4a4.lo}$, $\val{sc20.ctx.mxfp432w4a4.hi}$]}$^{**}$ \\
MXFP4-16$^{\dagger}$ & $\val{sc20.cnn.mxfp416w4a4.d}$ {\scriptsize[$\val{sc20.cnn.mxfp416w4a4.lo}$, $\val{sc20.cnn.mxfp416w4a4.hi}$]} & $\val{sc20.enc.mxfp416w4a4.d}$ {\scriptsize[$\val{sc20.enc.mxfp416w4a4.lo}$, $\val{sc20.enc.mxfp416w4a4.hi}$]}$^{**}$ & $\val{sc20.ctx.mxfp416w4a4.d}$ {\scriptsize[$\val{sc20.ctx.mxfp416w4a4.lo}$, $\val{sc20.ctx.mxfp416w4a4.hi}$]}$^{**}$ \\
MXFP4-32 (W only) & $\val{sc20.cnn.mxfp432w4.d}$ {\scriptsize[$\val{sc20.cnn.mxfp432w4.lo}$, $\val{sc20.cnn.mxfp432w4.hi}$]} & $\val{sc20.enc.mxfp432w4.d}$ {\scriptsize[$\val{sc20.enc.mxfp432w4.lo}$, $\val{sc20.enc.mxfp432w4.hi}$]}$^{*}$ & $\val{sc20.ctx.mxfp432w4.d}$ {\scriptsize[$\val{sc20.ctx.mxfp432w4.lo}$, $\val{sc20.ctx.mxfp432w4.hi}$]} \\
NVFP4-16 & $\val{sc20.cnn.nvfp416w4a4.d}$ {\scriptsize[$\val{sc20.cnn.nvfp416w4a4.lo}$, $\val{sc20.cnn.nvfp416w4a4.hi}$]}$^{**}$ & $\val{sc20.enc.nvfp416w4a4.d}$ {\scriptsize[$\val{sc20.enc.nvfp416w4a4.lo}$, $\val{sc20.enc.nvfp416w4a4.hi}$]} & $\val{sc20.ctx.nvfp416w4a4.d}$ {\scriptsize[$\val{sc20.ctx.nvfp416w4a4.lo}$, $\val{sc20.ctx.nvfp416w4a4.hi}$]} \\
NVFP4-32$^{\dagger}$ & $\val{sc20.cnn.nvfp432w4a4.d}$ {\scriptsize[$\val{sc20.cnn.nvfp432w4a4.lo}$, $\val{sc20.cnn.nvfp432w4a4.hi}$]}$^{*}$ & $\val{sc20.enc.nvfp432w4a4.d}$ {\scriptsize[$\val{sc20.enc.nvfp432w4a4.lo}$, $\val{sc20.enc.nvfp432w4a4.hi}$]} & $\val{sc20.ctx.nvfp432w4a4.d}$ {\scriptsize[$\val{sc20.ctx.nvfp432w4a4.lo}$, $\val{sc20.ctx.nvfp432w4a4.hi}$]} \\
NVFP4-16 (W only) & $\val{sc20.cnn.nvfp416w4.d}$ {\scriptsize[$\val{sc20.cnn.nvfp416w4.lo}$, $\val{sc20.cnn.nvfp416w4.hi}$]}$^{**}$ & $\val{sc20.enc.nvfp416w4.d}$ {\scriptsize[$\val{sc20.enc.nvfp416w4.lo}$, $\val{sc20.enc.nvfp416w4.hi}$]} & $\val{sc20.ctx.nvfp416w4.d}$ {\scriptsize[$\val{sc20.ctx.nvfp416w4.lo}$, $\val{sc20.ctx.nvfp416w4.hi}$]} \\
\midrule
MXFP4-32 + FP8 f/l & $\val{sc20.cnn.mxfp432w4a4fp8firstlast.d}$ {\scriptsize[$\val{sc20.cnn.mxfp432w4a4fp8firstlast.lo}$, $\val{sc20.cnn.mxfp432w4a4fp8firstlast.hi}$]} & $\val{sc20.enc.mxfp432w4a4fp8firstlast.d}$ {\scriptsize[$\val{sc20.enc.mxfp432w4a4fp8firstlast.lo}$, $\val{sc20.enc.mxfp432w4a4fp8firstlast.hi}$]}$^{*}$ & $\val{sc20.ctx.mxfp432w4a4fp8firstlast.d}$ {\scriptsize[$\val{sc20.ctx.mxfp432w4a4fp8firstlast.lo}$, $\val{sc20.ctx.mxfp432w4a4fp8firstlast.hi}$]}$^{*}$ \\
NVFP4-16 + FP8 f/l & $\val{sc20.cnn.nvfp416w4a4fp8firstlast.d}$ {\scriptsize[$\val{sc20.cnn.nvfp416w4a4fp8firstlast.lo}$, $\val{sc20.cnn.nvfp416w4a4fp8firstlast.hi}$]} & $\val{sc20.enc.nvfp416w4a4fp8firstlast.d}$ {\scriptsize[$\val{sc20.enc.nvfp416w4a4fp8firstlast.lo}$, $\val{sc20.enc.nvfp416w4a4fp8firstlast.hi}$]} & $\val{sc20.ctx.nvfp416w4a4fp8firstlast.d}$ {\scriptsize[$\val{sc20.ctx.nvfp416w4a4fp8firstlast.lo}$, $\val{sc20.ctx.nvfp416w4a4fp8firstlast.hi}$]} \\
\bottomrule
\end{tabular}}
\end{table*}

Table~\ref{tab:sc20} lists all formats for the three models trained with the same pipeline on the Sleep-EDF-20 subset. The subset contains subjects 0--19 (aged 25--34), \val{sc20.total} epochs, and 39 nights. The FP32 Macro-F1 is \val{sc20.cnn.fp32.mf1}\% for SleepCNN, \val{sc20.enc.fp32.mf1}\% for EncoderNet, and \val{sc20.ctx.fp32.mf1}\% for ContextSleepNet. Sleep-EDF-20 is a subset of Sleep-EDF-78, so the two cohorts are not independent. Each Sleep-EDF-20 fold has only two test subjects. Section~\ref{sec:ablation} summarizes which findings hold on this subset.

\FloatBarrier
\section{Additional results}
\label{app:extra}
\begin{table*}
\centering
\caption{Same-encoder ablation on Sleep-EDF-78. EncoderNet is ContextSleepNet without the temporal convolutions (same encoder, classifier, training and calibration). $\Delta$MF1 of each model relative to its own FP32 model (per-subject paired, mean over 3 seeds; 95\% bootstrap CI over 78 subjects). Non-temporal effect (encoder and training regime): EncoderNet minus SleepCNN; temporal-layer effect: ContextSleepNet minus EncoderNet (differences of per-subject $\Delta$MF1). $^{*}$/$^{**}$: Holm-corrected Wilcoxon $p<0.05/0.01$ over the eight W4A4 contrasts. FP8 first/last rows are shown for reference; they lie outside this family and carry no stars.}
\label{tab:ablation}
\setlength{\tabcolsep}{3pt}
\footnotesize
\fitwide{%
\begin{tabular}{l lll ll}
\toprule
 & \multicolumn{3}{c}{$\Delta$MF1 vs own FP32 (pp)} & \multicolumn{2}{c}{Contrast (pp)} \\
\cmidrule(lr){2-4}\cmidrule(lr){5-6}
Condition & SleepCNN & EncoderNet & ContextSleepNet & Non-temporal effect & Temporal-layer effect \\
\midrule
FP32 MF1 (\%) & \val{abl.cnn.fp32.mf1} & \val{abl.enc.fp32.mf1} & \val{abl.ctx.fp32.mf1} & & \\
\midrule
MXFP4-32 & $\val{abl.cnn.mxfp432w4a4.d}$ {\scriptsize[$\val{abl.cnn.mxfp432w4a4.lo}$, $\val{abl.cnn.mxfp432w4a4.hi}$]} & $\val{abl.enc.mxfp432w4a4.d}$ {\scriptsize[$\val{abl.enc.mxfp432w4a4.lo}$, $\val{abl.enc.mxfp432w4a4.hi}$]} & $\val{abl.ctx.mxfp432w4a4.d}$ {\scriptsize[$\val{abl.ctx.mxfp432w4a4.lo}$, $\val{abl.ctx.mxfp432w4a4.hi}$]} & $\val{abl.con.enc.mxfp432w4a4.d}$ {\scriptsize[$\val{abl.con.enc.mxfp432w4a4.lo}$, $\val{abl.con.enc.mxfp432w4a4.hi}$]}$^{**}$ & $\val{abl.con.tcn.mxfp432w4a4.d}$ {\scriptsize[$\val{abl.con.tcn.mxfp432w4a4.lo}$, $\val{abl.con.tcn.mxfp432w4a4.hi}$]} \\
MXFP4-16$^{\dagger}$ & $\val{abl.cnn.mxfp416w4a4.d}$ {\scriptsize[$\val{abl.cnn.mxfp416w4a4.lo}$, $\val{abl.cnn.mxfp416w4a4.hi}$]} & $\val{abl.enc.mxfp416w4a4.d}$ {\scriptsize[$\val{abl.enc.mxfp416w4a4.lo}$, $\val{abl.enc.mxfp416w4a4.hi}$]} & $\val{abl.ctx.mxfp416w4a4.d}$ {\scriptsize[$\val{abl.ctx.mxfp416w4a4.lo}$, $\val{abl.ctx.mxfp416w4a4.hi}$]} & $\val{abl.con.enc.mxfp416w4a4.d}$ {\scriptsize[$\val{abl.con.enc.mxfp416w4a4.lo}$, $\val{abl.con.enc.mxfp416w4a4.hi}$]}$^{**}$ & $\val{abl.con.tcn.mxfp416w4a4.d}$ {\scriptsize[$\val{abl.con.tcn.mxfp416w4a4.lo}$, $\val{abl.con.tcn.mxfp416w4a4.hi}$]} \\
NVFP4-16 & $\val{abl.cnn.nvfp416w4a4.d}$ {\scriptsize[$\val{abl.cnn.nvfp416w4a4.lo}$, $\val{abl.cnn.nvfp416w4a4.hi}$]} & $\val{abl.enc.nvfp416w4a4.d}$ {\scriptsize[$\val{abl.enc.nvfp416w4a4.lo}$, $\val{abl.enc.nvfp416w4a4.hi}$]} & $\val{abl.ctx.nvfp416w4a4.d}$ {\scriptsize[$\val{abl.ctx.nvfp416w4a4.lo}$, $\val{abl.ctx.nvfp416w4a4.hi}$]} & $\val{abl.con.enc.nvfp416w4a4.d}$ {\scriptsize[$\val{abl.con.enc.nvfp416w4a4.lo}$, $\val{abl.con.enc.nvfp416w4a4.hi}$]}$^{**}$ & $\val{abl.con.tcn.nvfp416w4a4.d}$ {\scriptsize[$\val{abl.con.tcn.nvfp416w4a4.lo}$, $\val{abl.con.tcn.nvfp416w4a4.hi}$]} \\
NVFP4-32$^{\dagger}$ & $\val{abl.cnn.nvfp432w4a4.d}$ {\scriptsize[$\val{abl.cnn.nvfp432w4a4.lo}$, $\val{abl.cnn.nvfp432w4a4.hi}$]} & $\val{abl.enc.nvfp432w4a4.d}$ {\scriptsize[$\val{abl.enc.nvfp432w4a4.lo}$, $\val{abl.enc.nvfp432w4a4.hi}$]} & $\val{abl.ctx.nvfp432w4a4.d}$ {\scriptsize[$\val{abl.ctx.nvfp432w4a4.lo}$, $\val{abl.ctx.nvfp432w4a4.hi}$]} & $\val{abl.con.enc.nvfp432w4a4.d}$ {\scriptsize[$\val{abl.con.enc.nvfp432w4a4.lo}$, $\val{abl.con.enc.nvfp432w4a4.hi}$]}$^{**}$ & $\val{abl.con.tcn.nvfp432w4a4.d}$ {\scriptsize[$\val{abl.con.tcn.nvfp432w4a4.lo}$, $\val{abl.con.tcn.nvfp432w4a4.hi}$]} \\
\midrule
MXFP4-32 + FP8 f/l & $\val{abl.cnn.mxfp432w4a4fp8firstlast.d}$ {\scriptsize[$\val{abl.cnn.mxfp432w4a4fp8firstlast.lo}$, $\val{abl.cnn.mxfp432w4a4fp8firstlast.hi}$]} & $\val{abl.enc.mxfp432w4a4fp8firstlast.d}$ {\scriptsize[$\val{abl.enc.mxfp432w4a4fp8firstlast.lo}$, $\val{abl.enc.mxfp432w4a4fp8firstlast.hi}$]} & $\val{abl.ctx.mxfp432w4a4fp8firstlast.d}$ {\scriptsize[$\val{abl.ctx.mxfp432w4a4fp8firstlast.lo}$, $\val{abl.ctx.mxfp432w4a4fp8firstlast.hi}$]} & $\val{abl.con.enc.mxfp432w4a4fp8firstlast.d}$ {\scriptsize[$\val{abl.con.enc.mxfp432w4a4fp8firstlast.lo}$, $\val{abl.con.enc.mxfp432w4a4fp8firstlast.hi}$]} & $\val{abl.con.tcn.mxfp432w4a4fp8firstlast.d}$ {\scriptsize[$\val{abl.con.tcn.mxfp432w4a4fp8firstlast.lo}$, $\val{abl.con.tcn.mxfp432w4a4fp8firstlast.hi}$]} \\
NVFP4-16 + FP8 f/l & $\val{abl.cnn.nvfp416w4a4fp8firstlast.d}$ {\scriptsize[$\val{abl.cnn.nvfp416w4a4fp8firstlast.lo}$, $\val{abl.cnn.nvfp416w4a4fp8firstlast.hi}$]} & $\val{abl.enc.nvfp416w4a4fp8firstlast.d}$ {\scriptsize[$\val{abl.enc.nvfp416w4a4fp8firstlast.lo}$, $\val{abl.enc.nvfp416w4a4fp8firstlast.hi}$]} & $\val{abl.ctx.nvfp416w4a4fp8firstlast.d}$ {\scriptsize[$\val{abl.ctx.nvfp416w4a4fp8firstlast.lo}$, $\val{abl.ctx.nvfp416w4a4fp8firstlast.hi}$]} & $\val{abl.con.enc.nvfp416w4a4fp8firstlast.d}$ {\scriptsize[$\val{abl.con.enc.nvfp416w4a4fp8firstlast.lo}$, $\val{abl.con.enc.nvfp416w4a4fp8firstlast.hi}$]} & $\val{abl.con.tcn.nvfp416w4a4fp8firstlast.d}$ {\scriptsize[$\val{abl.con.tcn.nvfp416w4a4fp8firstlast.lo}$, $\val{abl.con.tcn.nvfp416w4a4fp8firstlast.hi}$]} \\
\bottomrule
\end{tabular}}
\end{table*}

\begin{table}
\centering
\caption{Single-layer sensitivity: $\Delta$MF1 (pp, 95\% CI) when only the named layer is quantized to W4A4 and all others stay FP32. Stars: Holm-corrected Wilcoxon $p<0.05/0.01$ within the probes of each model (8 for SleepCNN, 18 for ContextSleepNet).}
\label{tab:sens}
\setlength{\tabcolsep}{2.5pt}
\footnotesize
\fitcol{%
\begin{tabular}{l l l l}
\toprule
Model & Layer & MXFP4-32 & NVFP4-16 \\
\midrule
SleepCNN & \texttt{conv1} & $\val{cnn.ls.mxfp432w4a4.conv1.d}$ {\scriptsize[$\val{cnn.ls.mxfp432w4a4.conv1.dlo}$, $\val{cnn.ls.mxfp432w4a4.conv1.dhi}$]}$^{**}$ & $\val{cnn.ls.nvfp416w4a4.conv1.d}$ {\scriptsize[$\val{cnn.ls.nvfp416w4a4.conv1.dlo}$, $\val{cnn.ls.nvfp416w4a4.conv1.dhi}$]}$^{**}$ \\
 & \texttt{conv2} & $\val{cnn.ls.mxfp432w4a4.conv2.d}$ {\scriptsize[$\val{cnn.ls.mxfp432w4a4.conv2.dlo}$, $\val{cnn.ls.mxfp432w4a4.conv2.dhi}$]}$^{**}$ & $\val{cnn.ls.nvfp416w4a4.conv2.d}$ {\scriptsize[$\val{cnn.ls.nvfp416w4a4.conv2.dlo}$, $\val{cnn.ls.nvfp416w4a4.conv2.dhi}$]} \\
 & \texttt{fc1} & $\val{cnn.ls.mxfp432w4a4.fc1.d}$ {\scriptsize[$\val{cnn.ls.mxfp432w4a4.fc1.dlo}$, $\val{cnn.ls.mxfp432w4a4.fc1.dhi}$]}$^{**}$ & $\val{cnn.ls.nvfp416w4a4.fc1.d}$ {\scriptsize[$\val{cnn.ls.nvfp416w4a4.fc1.dlo}$, $\val{cnn.ls.nvfp416w4a4.fc1.dhi}$]}$^{*}$ \\
 & \texttt{fc2} & $\val{cnn.ls.mxfp432w4a4.fc2.d}$ {\scriptsize[$\val{cnn.ls.mxfp432w4a4.fc2.dlo}$, $\val{cnn.ls.mxfp432w4a4.fc2.dhi}$]}$^{**}$ & $\val{cnn.ls.nvfp416w4a4.fc2.d}$ {\scriptsize[$\val{cnn.ls.nvfp416w4a4.fc2.dlo}$, $\val{cnn.ls.nvfp416w4a4.fc2.dhi}$]}$^{**}$ \\
\midrule
ContextSleepNet & \texttt{b1.conv} & $\val{ctx.ls.mxfp432w4a4.b1conv.d}$ {\scriptsize[$\val{ctx.ls.mxfp432w4a4.b1conv.dlo}$, $\val{ctx.ls.mxfp432w4a4.b1conv.dhi}$]} & $\val{ctx.ls.nvfp416w4a4.b1conv.d}$ {\scriptsize[$\val{ctx.ls.nvfp416w4a4.b1conv.dlo}$, $\val{ctx.ls.nvfp416w4a4.b1conv.dhi}$]} \\
 & \texttt{b2.conv} & $\val{ctx.ls.mxfp432w4a4.b2conv.d}$ {\scriptsize[$\val{ctx.ls.mxfp432w4a4.b2conv.dlo}$, $\val{ctx.ls.mxfp432w4a4.b2conv.dhi}$]} & $\val{ctx.ls.nvfp416w4a4.b2conv.d}$ {\scriptsize[$\val{ctx.ls.nvfp416w4a4.b2conv.dlo}$, $\val{ctx.ls.nvfp416w4a4.b2conv.dhi}$]} \\
 & \texttt{b3.conv} & $\val{ctx.ls.mxfp432w4a4.b3conv.d}$ {\scriptsize[$\val{ctx.ls.mxfp432w4a4.b3conv.dlo}$, $\val{ctx.ls.mxfp432w4a4.b3conv.dhi}$]} & $\val{ctx.ls.nvfp416w4a4.b3conv.d}$ {\scriptsize[$\val{ctx.ls.nvfp416w4a4.b3conv.dlo}$, $\val{ctx.ls.nvfp416w4a4.b3conv.dhi}$]} \\
 & \texttt{fc} & $\val{ctx.ls.mxfp432w4a4.fc.d}$ {\scriptsize[$\val{ctx.ls.mxfp432w4a4.fc.dlo}$, $\val{ctx.ls.mxfp432w4a4.fc.dhi}$]} & $\val{ctx.ls.nvfp416w4a4.fc.d}$ {\scriptsize[$\val{ctx.ls.nvfp416w4a4.fc.dlo}$, $\val{ctx.ls.nvfp416w4a4.fc.dhi}$]} \\
 & \texttt{tcn.0} & $\val{ctx.ls.mxfp432w4a4.tcn0.d}$ {\scriptsize[$\val{ctx.ls.mxfp432w4a4.tcn0.dlo}$, $\val{ctx.ls.mxfp432w4a4.tcn0.dhi}$]} & $\val{ctx.ls.nvfp416w4a4.tcn0.d}$ {\scriptsize[$\val{ctx.ls.nvfp416w4a4.tcn0.dlo}$, $\val{ctx.ls.nvfp416w4a4.tcn0.dhi}$]} \\
 & \texttt{tcn.1} & $\val{ctx.ls.mxfp432w4a4.tcn1.d}$ {\scriptsize[$\val{ctx.ls.mxfp432w4a4.tcn1.dlo}$, $\val{ctx.ls.mxfp432w4a4.tcn1.dhi}$]} & $\val{ctx.ls.nvfp416w4a4.tcn1.d}$ {\scriptsize[$\val{ctx.ls.nvfp416w4a4.tcn1.dlo}$, $\val{ctx.ls.nvfp416w4a4.tcn1.dhi}$]} \\
 & \texttt{tcn.2} & $\val{ctx.ls.mxfp432w4a4.tcn2.d}$ {\scriptsize[$\val{ctx.ls.mxfp432w4a4.tcn2.dlo}$, $\val{ctx.ls.mxfp432w4a4.tcn2.dhi}$]} & $\val{ctx.ls.nvfp416w4a4.tcn2.d}$ {\scriptsize[$\val{ctx.ls.nvfp416w4a4.tcn2.dlo}$, $\val{ctx.ls.nvfp416w4a4.tcn2.dhi}$]} \\
 & \texttt{tcn.3} & $\val{ctx.ls.mxfp432w4a4.tcn3.d}$ {\scriptsize[$\val{ctx.ls.mxfp432w4a4.tcn3.dlo}$, $\val{ctx.ls.mxfp432w4a4.tcn3.dhi}$]} & $\val{ctx.ls.nvfp416w4a4.tcn3.d}$ {\scriptsize[$\val{ctx.ls.nvfp416w4a4.tcn3.dlo}$, $\val{ctx.ls.nvfp416w4a4.tcn3.dhi}$]} \\
 & \texttt{cls} & $\val{ctx.ls.mxfp432w4a4.cls.d}$ {\scriptsize[$\val{ctx.ls.mxfp432w4a4.cls.dlo}$, $\val{ctx.ls.mxfp432w4a4.cls.dhi}$]} & $\val{ctx.ls.nvfp416w4a4.cls.d}$ {\scriptsize[$\val{ctx.ls.nvfp416w4a4.cls.dlo}$, $\val{ctx.ls.nvfp416w4a4.cls.dhi}$]} \\
\bottomrule
\end{tabular}}
\end{table}

\begin{table*}
\centering
\caption{Only the first layer quantized (all other layers FP32): weights only, input activations only, or both. $\Delta$MF1 in pp (95\% CI). $^{*}$/$^{**}$: Holm-corrected Wilcoxon $p<0.05/0.01$ over the twelve W-only and A-only conditions.}
\label{tab:lone}
\setlength{\tabcolsep}{2.5pt}
\footnotesize
\fitwide{%
\begin{tabular}{l l lll}
\toprule
Model & Format & Weights only & Input only & Both \\
\midrule
SleepCNN & MXFP4-32 & $\val{lone.cnn.mxfp432w4a4.w.d}$ {\scriptsize[$\val{lone.cnn.mxfp432w4a4.w.lo}$, $\val{lone.cnn.mxfp432w4a4.w.hi}$]}$^{**}$ & $\val{lone.cnn.mxfp432w4a4.a.d}$ {\scriptsize[$\val{lone.cnn.mxfp432w4a4.a.lo}$, $\val{lone.cnn.mxfp432w4a4.a.hi}$]}$^{**}$ & $\val{lone.cnn.mxfp432w4a4.wa.d}$ {\scriptsize[$\val{lone.cnn.mxfp432w4a4.wa.lo}$, $\val{lone.cnn.mxfp432w4a4.wa.hi}$]} \\
 & NVFP4-16 & $\val{lone.cnn.nvfp416w4a4.w.d}$ {\scriptsize[$\val{lone.cnn.nvfp416w4a4.w.lo}$, $\val{lone.cnn.nvfp416w4a4.w.hi}$]}$^{**}$ & $\val{lone.cnn.nvfp416w4a4.a.d}$ {\scriptsize[$\val{lone.cnn.nvfp416w4a4.a.lo}$, $\val{lone.cnn.nvfp416w4a4.a.hi}$]}$^{**}$ & $\val{lone.cnn.nvfp416w4a4.wa.d}$ {\scriptsize[$\val{lone.cnn.nvfp416w4a4.wa.lo}$, $\val{lone.cnn.nvfp416w4a4.wa.hi}$]} \\
\midrule
EncoderNet & MXFP4-32 & $\val{lone.enc.mxfp432w4a4.w.d}$ {\scriptsize[$\val{lone.enc.mxfp432w4a4.w.lo}$, $\val{lone.enc.mxfp432w4a4.w.hi}$]} & $\val{lone.enc.mxfp432w4a4.a.d}$ {\scriptsize[$\val{lone.enc.mxfp432w4a4.a.lo}$, $\val{lone.enc.mxfp432w4a4.a.hi}$]} & $\val{lone.enc.mxfp432w4a4.wa.d}$ {\scriptsize[$\val{lone.enc.mxfp432w4a4.wa.lo}$, $\val{lone.enc.mxfp432w4a4.wa.hi}$]} \\
 & NVFP4-16 & $\val{lone.enc.nvfp416w4a4.w.d}$ {\scriptsize[$\val{lone.enc.nvfp416w4a4.w.lo}$, $\val{lone.enc.nvfp416w4a4.w.hi}$]} & $\val{lone.enc.nvfp416w4a4.a.d}$ {\scriptsize[$\val{lone.enc.nvfp416w4a4.a.lo}$, $\val{lone.enc.nvfp416w4a4.a.hi}$]} & $\val{lone.enc.nvfp416w4a4.wa.d}$ {\scriptsize[$\val{lone.enc.nvfp416w4a4.wa.lo}$, $\val{lone.enc.nvfp416w4a4.wa.hi}$]} \\
\midrule
ContextSleepNet & MXFP4-32 & $\val{lone.ctx.mxfp432w4a4.w.d}$ {\scriptsize[$\val{lone.ctx.mxfp432w4a4.w.lo}$, $\val{lone.ctx.mxfp432w4a4.w.hi}$]} & $\val{lone.ctx.mxfp432w4a4.a.d}$ {\scriptsize[$\val{lone.ctx.mxfp432w4a4.a.lo}$, $\val{lone.ctx.mxfp432w4a4.a.hi}$]} & $\val{lone.ctx.mxfp432w4a4.wa.d}$ {\scriptsize[$\val{lone.ctx.mxfp432w4a4.wa.lo}$, $\val{lone.ctx.mxfp432w4a4.wa.hi}$]} \\
 & NVFP4-16 & $\val{lone.ctx.nvfp416w4a4.w.d}$ {\scriptsize[$\val{lone.ctx.nvfp416w4a4.w.lo}$, $\val{lone.ctx.nvfp416w4a4.w.hi}$]} & $\val{lone.ctx.nvfp416w4a4.a.d}$ {\scriptsize[$\val{lone.ctx.nvfp416w4a4.a.lo}$, $\val{lone.ctx.nvfp416w4a4.a.hi}$]} & $\val{lone.ctx.nvfp416w4a4.wa.d}$ {\scriptsize[$\val{lone.ctx.nvfp416w4a4.wa.lo}$, $\val{lone.ctx.nvfp416w4a4.wa.hi}$]} \\
\bottomrule
\end{tabular}}
\end{table*}

\begin{table*}
\centering
\caption{Per-stage F1 (\%), pooled over the 78 test subjects, mean over three seeds.}
\label{tab:perclass}
\setlength{\tabcolsep}{3.5pt}
\footnotesize
\begin{tabular}{l rrrrr rrrrr}
\toprule
 & \multicolumn{5}{c}{SleepCNN} & \multicolumn{5}{c}{ContextSleepNet} \\
\cmidrule(lr){2-6}\cmidrule(lr){7-11}
Format & W & N1 & N2 & N3 & REM & W & N1 & N2 & N3 & REM \\
\midrule
FP32 & \val{cnn.fp32.f1w} & \val{cnn.fp32.f1n1} & \val{cnn.fp32.f1n2} & \val{cnn.fp32.f1n3} & \val{cnn.fp32.f1rem} & \val{ctx.fp32.f1w} & \val{ctx.fp32.f1n1} & \val{ctx.fp32.f1n2} & \val{ctx.fp32.f1n3} & \val{ctx.fp32.f1rem} \\
FP16 & \val{cnn.fp16.f1w} & \val{cnn.fp16.f1n1} & \val{cnn.fp16.f1n2} & \val{cnn.fp16.f1n3} & \val{cnn.fp16.f1rem} & \val{ctx.fp16.f1w} & \val{ctx.fp16.f1n1} & \val{ctx.fp16.f1n2} & \val{ctx.fp16.f1n3} & \val{ctx.fp16.f1rem} \\
BF16 & \val{cnn.bf16.f1w} & \val{cnn.bf16.f1n1} & \val{cnn.bf16.f1n2} & \val{cnn.bf16.f1n3} & \val{cnn.bf16.f1rem} & \val{ctx.bf16.f1w} & \val{ctx.bf16.f1n1} & \val{ctx.bf16.f1n2} & \val{ctx.bf16.f1n3} & \val{ctx.bf16.f1rem} \\
FP8-E4M3 & \val{cnn.fp8e4m3w8a8.f1w} & \val{cnn.fp8e4m3w8a8.f1n1} & \val{cnn.fp8e4m3w8a8.f1n2} & \val{cnn.fp8e4m3w8a8.f1n3} & \val{cnn.fp8e4m3w8a8.f1rem} & \val{ctx.fp8e4m3w8a8.f1w} & \val{ctx.fp8e4m3w8a8.f1n1} & \val{ctx.fp8e4m3w8a8.f1n2} & \val{ctx.fp8e4m3w8a8.f1n3} & \val{ctx.fp8e4m3w8a8.f1rem} \\
FP8-E5M2 & \val{cnn.fp8e5m2w8a8.f1w} & \val{cnn.fp8e5m2w8a8.f1n1} & \val{cnn.fp8e5m2w8a8.f1n2} & \val{cnn.fp8e5m2w8a8.f1n3} & \val{cnn.fp8e5m2w8a8.f1rem} & \val{ctx.fp8e5m2w8a8.f1w} & \val{ctx.fp8e5m2w8a8.f1n1} & \val{ctx.fp8e5m2w8a8.f1n2} & \val{ctx.fp8e5m2w8a8.f1n3} & \val{ctx.fp8e5m2w8a8.f1rem} \\
INT8 & \val{cnn.int8w8a8.f1w} & \val{cnn.int8w8a8.f1n1} & \val{cnn.int8w8a8.f1n2} & \val{cnn.int8w8a8.f1n3} & \val{cnn.int8w8a8.f1rem} & \val{ctx.int8w8a8.f1w} & \val{ctx.int8w8a8.f1n1} & \val{ctx.int8w8a8.f1n2} & \val{ctx.int8w8a8.f1n3} & \val{ctx.int8w8a8.f1rem} \\
INT8 (W only) & \val{cnn.int8w8.f1w} & \val{cnn.int8w8.f1n1} & \val{cnn.int8w8.f1n2} & \val{cnn.int8w8.f1n3} & \val{cnn.int8w8.f1rem} & \val{ctx.int8w8.f1w} & \val{ctx.int8w8.f1n1} & \val{ctx.int8w8.f1n2} & \val{ctx.int8w8.f1n3} & \val{ctx.int8w8.f1rem} \\
INT4-g32 (W only) & \val{cnn.int4g32w4.f1w} & \val{cnn.int4g32w4.f1n1} & \val{cnn.int4g32w4.f1n2} & \val{cnn.int4g32w4.f1n3} & \val{cnn.int4g32w4.f1rem} & \val{ctx.int4g32w4.f1w} & \val{ctx.int4g32w4.f1n1} & \val{ctx.int4g32w4.f1n2} & \val{ctx.int4g32w4.f1n3} & \val{ctx.int4g32w4.f1rem} \\
MXFP4-32 & \val{cnn.mxfp432w4a4.f1w} & \val{cnn.mxfp432w4a4.f1n1} & \val{cnn.mxfp432w4a4.f1n2} & \val{cnn.mxfp432w4a4.f1n3} & \val{cnn.mxfp432w4a4.f1rem} & \val{ctx.mxfp432w4a4.f1w} & \val{ctx.mxfp432w4a4.f1n1} & \val{ctx.mxfp432w4a4.f1n2} & \val{ctx.mxfp432w4a4.f1n3} & \val{ctx.mxfp432w4a4.f1rem} \\
MXFP4-16$^{\dagger}$ & \val{cnn.mxfp416w4a4.f1w} & \val{cnn.mxfp416w4a4.f1n1} & \val{cnn.mxfp416w4a4.f1n2} & \val{cnn.mxfp416w4a4.f1n3} & \val{cnn.mxfp416w4a4.f1rem} & \val{ctx.mxfp416w4a4.f1w} & \val{ctx.mxfp416w4a4.f1n1} & \val{ctx.mxfp416w4a4.f1n2} & \val{ctx.mxfp416w4a4.f1n3} & \val{ctx.mxfp416w4a4.f1rem} \\
MXFP4-32 (W only) & \val{cnn.mxfp432w4.f1w} & \val{cnn.mxfp432w4.f1n1} & \val{cnn.mxfp432w4.f1n2} & \val{cnn.mxfp432w4.f1n3} & \val{cnn.mxfp432w4.f1rem} & \val{ctx.mxfp432w4.f1w} & \val{ctx.mxfp432w4.f1n1} & \val{ctx.mxfp432w4.f1n2} & \val{ctx.mxfp432w4.f1n3} & \val{ctx.mxfp432w4.f1rem} \\
NVFP4-16 & \val{cnn.nvfp416w4a4.f1w} & \val{cnn.nvfp416w4a4.f1n1} & \val{cnn.nvfp416w4a4.f1n2} & \val{cnn.nvfp416w4a4.f1n3} & \val{cnn.nvfp416w4a4.f1rem} & \val{ctx.nvfp416w4a4.f1w} & \val{ctx.nvfp416w4a4.f1n1} & \val{ctx.nvfp416w4a4.f1n2} & \val{ctx.nvfp416w4a4.f1n3} & \val{ctx.nvfp416w4a4.f1rem} \\
NVFP4-32$^{\dagger}$ & \val{cnn.nvfp432w4a4.f1w} & \val{cnn.nvfp432w4a4.f1n1} & \val{cnn.nvfp432w4a4.f1n2} & \val{cnn.nvfp432w4a4.f1n3} & \val{cnn.nvfp432w4a4.f1rem} & \val{ctx.nvfp432w4a4.f1w} & \val{ctx.nvfp432w4a4.f1n1} & \val{ctx.nvfp432w4a4.f1n2} & \val{ctx.nvfp432w4a4.f1n3} & \val{ctx.nvfp432w4a4.f1rem} \\
NVFP4-16 (W only) & \val{cnn.nvfp416w4.f1w} & \val{cnn.nvfp416w4.f1n1} & \val{cnn.nvfp416w4.f1n2} & \val{cnn.nvfp416w4.f1n3} & \val{cnn.nvfp416w4.f1rem} & \val{ctx.nvfp416w4.f1w} & \val{ctx.nvfp416w4.f1n1} & \val{ctx.nvfp416w4.f1n2} & \val{ctx.nvfp416w4.f1n3} & \val{ctx.nvfp416w4.f1rem} \\
MXFP4-32 + FP8 f/l & \val{cnn.mxfp432w4a4fp8firstlast.f1w} & \val{cnn.mxfp432w4a4fp8firstlast.f1n1} & \val{cnn.mxfp432w4a4fp8firstlast.f1n2} & \val{cnn.mxfp432w4a4fp8firstlast.f1n3} & \val{cnn.mxfp432w4a4fp8firstlast.f1rem} & \val{ctx.mxfp432w4a4fp8firstlast.f1w} & \val{ctx.mxfp432w4a4fp8firstlast.f1n1} & \val{ctx.mxfp432w4a4fp8firstlast.f1n2} & \val{ctx.mxfp432w4a4fp8firstlast.f1n3} & \val{ctx.mxfp432w4a4fp8firstlast.f1rem} \\
NVFP4-16 + FP8 f/l & \val{cnn.nvfp416w4a4fp8firstlast.f1w} & \val{cnn.nvfp416w4a4fp8firstlast.f1n1} & \val{cnn.nvfp416w4a4fp8firstlast.f1n2} & \val{cnn.nvfp416w4a4fp8firstlast.f1n3} & \val{cnn.nvfp416w4a4fp8firstlast.f1rem} & \val{ctx.nvfp416w4a4fp8firstlast.f1w} & \val{ctx.nvfp416w4a4fp8firstlast.f1n1} & \val{ctx.nvfp416w4a4fp8firstlast.f1n2} & \val{ctx.nvfp416w4a4fp8firstlast.f1n3} & \val{ctx.nvfp416w4a4fp8firstlast.f1rem} \\
\bottomrule
\end{tabular}
\end{table*}

Table~\ref{tab:ablation} gives the full same-encoder ablation shown in Fig.~\ref{fig:ablation}a. Table~\ref{tab:sens} gives the single-layer sensitivities shown in Fig.~\ref{fig:mechanism}a. Table~\ref{tab:lone} gives the first-layer weight and input probes, and Table~\ref{tab:perclass} gives the per-stage F1 scores of all formats.
\FloatBarrier

\end{document}